\documentclass{article}

\usepackage{microtype}
\usepackage{graphicx}
\usepackage{booktabs} 

\usepackage{soul,xcolor}

\usepackage{bm} 

\usepackage{enumitem}  
\usepackage{comment}  
\usepackage{kbordermatrix}  

\renewcommand{\kbldelim}{(}  
\renewcommand{\kbrdelim}{)}

\usepackage{pgf} 
\usepackage[utf8]{inputenc}
\DeclareUnicodeCharacter{2212}{-}
\DeclareUnicodeCharacter{1D464}{$w$}  

\usepackage[textsize=tiny]{todonotes}

\usepackage{pgfplots}
\usepgfplotslibrary{groupplots,dateplot}
\usetikzlibrary{patterns,shapes.arrows}
\pgfplotsset{compat=newest}

\usepackage{subcaption}

\graphicspath{{figures/}}

\usepackage{hyperref}  

\usepackage{colonequals}

\usepackage[accepted]{icml2023}

\usepackage{amsmath}
\usepackage{amssymb}
\usepackage{mathtools}
\usepackage{amsthm}
\usepackage{bbold}  
\usepackage{mathrsfs}  

\usepackage{adjustbox}
\usepackage[capitalize,noabbrev]{cleveref}

\newcommand\numberthis{\addtocounter{equation}{1}\tag{\theequation}} 

\theoremstyle{plain}
\newtheorem{theorem}{Theorem}[section]
\newtheorem{proposition}[theorem]{Proposition}
\newtheorem{lemma}[theorem]{Lemma}
\newtheorem{corollary}[theorem]{Corollary}
\theoremstyle{definition}
\newtheorem{definition}[theorem]{Definition}

\newtheorem*{notation}{Notation}
\theoremstyle{remark}
\newtheorem{remark}[theorem]{Remark}

\newcommand{\cB}{\mathcal{B}}
\newcommand{\cC}{\mathcal{C}}
\newcommand{\cE}{\mathcal{E}}
\newcommand{\cJ}{\mathcal{J}}
\newcommand{\cL}{\mathcal{L}}
\newcommand{\cM}{\mathcal{M}}
\newcommand{\cN}{\mathcal{N}}
\newcommand{\cO}{\mathcal{O}}
\newcommand{\cP}{\mathcal{P}}
\newcommand{\cQ}{\mathcal{Q}}
\newcommand{\cS}{\mathcal{S}}
\newcommand{\cX}{\mathcal{X}}
\newcommand{\cY}{\mathcal{Y}}
\newcommand{\cW}{\mathcal{W}}

\newcommand{\bF}{\mathbb{F}}
\newcommand{\bG}{\mathbb{G}}
\newcommand{\bH}{\mathbb{H}}
\newcommand{\bN}{\mathbb{N}}
\newcommand{\bR}{\mathbb{R}}

\newcommand{\kL}{\mathfrak{L}}

\newcommand{\W}{\overrightarrow{W}}
\newcommand{\ud}{\underline{d}}  

\newcommand{\defeq}{:=}
\newcommand{\eqdef}{=:}

\DeclareMathOperator{\trc}{tr}
\DeclareMathOperator{\vc}{vec}
\DeclareMathOperator{\rank}{rank}

\DeclareMathOperator{\spn}{span}
\DeclareMathOperator{\id}{id}
\DeclareMathOperator{\Crit}{Crit}
\DeclareMathOperator{\Diag}{Diag}
\DeclareMathOperator{\GL}{GL}
\DeclareMathOperator{\supp}{supp}
\newcommand{\rel}{\mathrm{rel}}
\newcommand{\absol}{\mathrm{abs}}

\DeclareMathOperator*{\argmin}{arg\,min}
  
\DeclareMathOperator{\dif}{d\!}  

\newcommand{\od}[2]{\frac{\dif #1}{\dif #2}}  

\let\ang\relax  
\DeclarePairedDelimiter{\ang}{\langle}{\rangle}  
\let\norm\relax  
\DeclarePairedDelimiter\norm{\|}{\|}   
\let\set\relax
\DeclarePairedDelimiter\set{\{}{\}}   
\let\abs\relax  
\DeclarePairedDelimiter\abs{|}{|}   

\icmltitlerunning{Deep Linear Networks Trained with Bures-Wasserstein Loss}

\begin{document}

\setstcolor{blue} 

\twocolumn[
\icmltitle{Critical Points and Convergence Analysis of Generative Deep Linear Networks Trained with Bures-Wasserstein Loss}

\icmlsetsymbol{equal}{*}

\begin{icmlauthorlist}
\icmlauthor{Pierre Bréchet}{mis}
\icmlauthor{Katerina Papagiannouli}{mis}
\icmlauthor{Jing An}{duke}
\icmlauthor{Guido Montúfar}{mis,ucla}
\end{icmlauthorlist}

\icmlaffiliation{mis}{Max Planck Institute for Mathematics in the Sciences, 
Leipzig, Germany}
\icmlaffiliation{duke}{Department of Mathematics, Duke University, Durham, NC, USA} 
\icmlaffiliation{ucla}{Departments of Mathematics and Statistics, UCLA, Los Angeles, CA, USA} 

\icmlcorrespondingauthor{Pierre Bréchet}{pierre.brechet@mis.mpg.de}
\icmlkeywords{deep linear networks, low-rank approximation, Bures-Wasserstein
distance,optimal transport, implicit generative model, critical points} 

\vskip 0.3in
]

\printAffiliationsAndNotice{}  

\begin{abstract}
   We consider a deep matrix factorization model of covariance matrices trained with the Bures-Wasserstein distance. 
   While recent works have made advances in the study of the optimization problem for overparametrized low-rank matrix approximation, much emphasis has been placed on discriminative settings and the square loss. In contrast, our model considers another type of loss and connects with the generative setting. 
   We characterize the critical points and minimizers of the Bures-Wasserstein
   distance over the space of rank-bounded matrices. The Hessian of this loss at
   low-rank matrices can theoretically blow up, which creates challenges to
   analyze convergence of gradient optimization methods. 
   We establish convergence results for gradient flow using a smooth perturbative version of the loss as well as convergence results for finite step size gradient descent under certain assumptions on the initial weights. 
   \end{abstract}

\section{Introduction}

We investigate generative deep linear networks and their optimization using the Bures-Wasserstein distance. More precisely, we consider the problem of approximating a target Gaussian distribution with a deep linear neural network generator of Gaussian distributions by minimizing the Bures-Wasserstein distance. 
This problem is of interest in two ways. First, it pertains to the optimization of deep linear networks for a type of loss that is qualitatively different from the well-studied and very particular squared error loss. 
Second, it can be regarded as a simplified but instructive instance of the
parameter optimization problem in generative networks, specifically Wasserstein
generative adversarial networks, which are currently not as well understood as
discriminative models.

The optimization landscapes and the properties of parameter optimization procedures for neural networks are among the most puzzling and actively studied topics in theoretical deep learning \citep[see, e.g.][]{mei2018mean, liu2022loss}. 
Deep linear networks, i.e.\ neural networks having the identity as activation function, serve as simplified models for such investigations \citep{Baldi1989Neural,Kawaguchi2016Deep,Trager2020Pure,Kohn2021GeometryOL,Bah2021Learning}. The study of linear networks has guided the development of several useful notions and intuitions in the theoretical analysis of neural networks, from the absence of bad local minima to the role of parametrization and overparametrization in gradient optimization \citep{Arora2018Optimization,Arora2019Convergence,Arora2019Implicit}. 
Many previous works have focused on discriminative or autoregressive settings and have emphasized the squared error loss. Although this loss is indeed a popular choice in regression tasks, it interacts in a very special way with the particular geometry of linear networks \citep{Trager2020Pure}. 
The behavior of linear networks optimized with different losses has also been considered in several works \citep{Laurent2018Deep, Lu2017Depth, Trager2020Pure} but is less well understood.

The Bures-Wasserstein distance was introduced by \citet{Bures1969extension} to study Hermitian operators in quantum information, particularly density matrices. It induces a metric on the space of positive semi-definite matrices, 
and corresponds to the 2-Wasserstein distance between
two centered Gaussian distributions \citep{bhatia2019bures}. Wasserstein
distances have several useful properties, e.g.\ they remain well defined between disjointly supported measures and have duality formulations \citep{villani2003topics} that allow for practical implementations. 
This makes them good candidates and indeed popular choices for learning generative models, with a well-known case being the Wasserstein Generative Adversarial Networks (WGANs) \citep{Arjovsky2017Wasserstein}. 
While the 1-Wasserstein distance has been most commonly used in this context, the Bures-Wasserstein distance has also attracted interest, e.g.\ in the works of  \citet{NEURIPS2018_b613e70f,chewi2020gradient,mallasto2022entropy}, and has also appeared in the context of linear quadratic Wasserstein generative adversarial networks \citep{Feizi2020understanding}. 

Notably, \citet{de2021bures} observed experimentally that the Bures-Wasserstein
metric reduces the infamous problem of mode collapse in GANs. In particular, the
authors reported improvements in mode coverage and generation quality by adding
the Bures metric to the objective function of a GAN\@.
Our work casts light on the theoretical properties of Bures-Wasserstein metric as a loss
function to train deep linear generative neural networks, by studying a specific 2-Wasserstein GAN model. 

A 2-Wasserstein GAN is a minimum 2-Wasserstein distance estimator expressed in Kantorovich duality (see details in Appendix~\ref{app:BW}). 
This model can serve as a platform to develop the theory particularly when the inner problem can be solved in closed-form. 
Such a formula is available when comparing pairs of Gaussian distributions, in particular centered Gaussians, which corresponds precisely to the Bures-Wasserstein distance between the corresponding covariance matrices. 
Strikingly, even in this simple case, the properties of the corresponding optimization problem are not well understood; we aim to address this in the present work.

\subsection{Contributions} 

We establish a series of results on the optimization of deep linear networks trained with the Bures-Wasserstein loss: 

\begin{itemize}[leftmargin=*]
    \item We obtain an analogue of the Eckart-Young-Mirsky theorem characterizing the critical points and minimizers of the Bures-Wasserstein distance over matrices of a given rank (\Cref{thm:critical-points-L1}). 
    
    \item To circumvent the non-smooth behaviour of the Bures-Wasserstein loss
        when the matrices drop rank, we introduce a smooth perturbative version
        (Definition~\ref{pertloss} and Lemma~\ref{lem:diff-losses}), and
        characterize its critical points and minimizers over rank-constrained
        matrices (Theorem~\ref{thm:critical-points-L1tau}). Under some
        conditions on the function realization, we connect them to critical
        points on the parameter space (\Cref{prop:crit-param-func}). 
    
    \item For the Bures-Wasserstein loss and its smooth version, in
        Theorem~\ref{thm:convergenceGF} and \Cref{rem:GForig}, we show  exponential convergence of the gradient
        flow assuming balanced initial weights (Definition~\ref{def:balanced-weights}) and a 
        modified margin deficiency condition (Definition~\ref{def:MDM}). 
    
    \item For the Bures-Wasserstein loss and its smooth version, in
        Theorem~\ref{thm:gradientdescent}, we show convergence of gradient descent provided the step size is small enough and the initial weights are balanced. 
\end{itemize}

\subsection{Related works}

\paragraph{Low rank matrix approximation} 
The function space of a linear network corresponds to $n \times m$ matrices
of rank at most $\ud$, the smallest width of the network. Hence optimization in the function space is closely related to the problem of approximating a given data matrix by a low-rank matrix. 
When the approximation error is measured in Frobenius norm, \citet{Eckart1936approximation} characterized the optimal bounded-rank approximation of a given matrix in terms of its singular value decomposition. 
\citet{mirsky1960symmetric} obtained the same characterization for the more general case of unitary invariant matrix norms, which include the 
Euclidean operator norm and the Schatten-$p$ norms. 
There are further generalizations to 
certain weighted norms \citep{
ruben1979lower,dutta2017problem}. 
However, for general norms the problem is known to be difficult \citep{zhao2017lowl1,gillis2018complexity,gillis2017low}. 

\paragraph{Loss landscape of deep linear networks}

For the squared error loss, the optimization landscape of linear networks has been studied in numerous works. 
The pioneering work of \citet{Baldi1989Neural} focused on the two-layer case, and showed that there is a single minimum (up to a trivial parametrization symmetry) and all other critical points are saddle points. 
\citet{Kawaguchi2016Deep} obtained corresponding results for deep linear networks and showed the existence of bad saddles (with no negative Hessian eigenvalues) in parameter space for networks with more than three layers. 
\citet{Lu2017Depth} showed that if the loss is such that any local minimizer in parameter space can be perturbed to an equally good minimizer with full-rank factor matrices, then all local minimizers in parameter space are local minimizers in function space. 
\citet{yun2018global} found sets of parameters in which any critical point is a global minimizer, and any outside critical point is a saddle point. 
We also mention other works that study critical points for different types of neural network architectures, such as deep linear residual networks \citep{hardt2017identity} and deep linear convolutional networks \citep{Kohn2021GeometryOL,kohn2023function}. 

There are also several results for different losses. 
\citet{Laurent2018Deep} showed that for deep linear nets with no bottlenecks all local minima are global for arbitrary convex differentiable losses. 
\citet{Trager2020Pure} found that 
for linear networks with arbitrarily rank-constrained function space, the squared error loss is special in the sense that it ensures the non-existence of non-global local minima. However, for arbitrary convex losses, non-global local minimizers, when they exist, are always pure, meaning that they correspond to local minimizers in function space.

\paragraph{Optimization dynamics of deep linear networks} 

\citet{DBLP:journals/corr/SaxeMG13} studied the learning dynamics of deep linear
networks under different types of initial conditions. 
\citet{Arora2019Implicit} obtained a closed-form expression for the
parametrization along time in a deep linear network for the squared error loss. 
Notably, the authors found that solutions with a lower rank are preferred as the depth of the network increases. 
\citet{Arora2018Optimization} derived several invariances of the flow and compared the dynamics in parameter and function spaces. 
For the squared error loss, \citet{Arora2019Convergence} proved linear convergence of gradient descent for linear networks without bottlenecks, with weights initialized to fulfill two assumptions --- approximate balancedness and so that the end-to-end matrix is close in some sense to the solution. 
We frame our discussion by similar assumptions. Under the balancedness assumption for the initial weights, \citet{Bah2021Learning} showed that for deep linear neural networks,
the gradient flow of the squared error loss can be cast as a Riemannian gradient flow in the function space, and as such converges to a critical point which is a global minimizer on the manifold of fixed rank matrices of a given rank. 
More recently, \citet{nguegnang2021convergence} extended this convergence analysis to the (full-batch) gradient descent algorithm. 

As a last note, a detailed analysis of the dynamics in the case of shallow linear networks with the squared error loss was
conducted by \citet{Tarmoun2021Understanding,Min2021Explicit}. The authors use  symmetric and asymmetric factorization of a shallow linear network to study its convergence dynamics. The role of the ``imbalancedness'' of
the weights was also remarked in those works. 

\paragraph{Bures-Wasserstein distance} 
The Bures-Wasserstein distance has been of particular interests due to its geometrical properties. 
\citet{chewi2020gradient} studied the convergence of gradient descent algorithms for the Bures-Wasserstein barycenter, proving linear rates of convergence. 
In contrast to our work, they considered a Polyak-\L{}ojasiewicz inequality derived from optimal transport theory to circumvent the non geodesic convexity of the barycenter. 
In the same vein, \citet{NEURIPS2018_b613e70f} exploited optimal transport theory to optimize the distance between two elliptical distributions. To avoid rank deficiency, they perturbed the diagonal elements of the covariance matrix by a small parameter. 
We also mention that \citet{Feizi2020understanding} characterized the optimal solution of a 2-Wasserstein GAN with a rank-$k$ linear generator as the $k$-PCA solution. 
We will obtain an analogous result in our particular parametrization, along with detailed descriptions of critical points.

\subsection{Notations} 

We adopt the following notations. For any $n \in \bN$, let $[n] \defeq \{1,2, \ldots, n \}$. 
We equip $\bR^n$ with its usual inner product, and we denote by~$\cO(n)$ the space of real orthogonal matrices of size $n$. Let~$\cS(n)$ be the space of real symmetric matrices of size~$n$. 
We denote~$\cS_+(n)$ (resp.\ $\cS_{++}(n)$) the space of real symmetric positive semi-definite (resp.\ definite) matrices of size~$n$. 
We use $\cM(k; n, m)$ (resp.\ $\cM(\leqslant k; n, m)$) to denote the set of matrices of size $n \times m$ with rank exactly $k$ (resp.\ at most $k$). 
If not specified, the size of the matrix is $n \times m$. 
The scalar product between two matrices~$A, B \in \bR^{n \times m}$ is~$\ang{A, B} = \trc{A^\top B}$, and the associated Frobenius norm is~$\norm{\cdot}_F^2$. 
The identity matrix of size $n$ will be written as $I_n$, or~$I$ when $n$ is clear. 
For a (Fréchet) differentiable function~$f \colon X \to Y$, we denote its differential at~$x \in X$ in the direction~$v$ by $\dif{}f(x)[v]$. 
Finally, $\Crit(f)$ is the set of critical points of $f$, i.e.\ the set of points at which the differential of $f$ is $0$.

\section{Linear networks and their gradient dynamics}
\label{sec:prelim-results}
\label{sec:flow-linear-net}

We consider a linear network with $d_0$ inputs and $N$ layers of widths
$d_1,\ldots, d_N$, which is a model of linear functions of the form 
$$ 
x\mapsto W_N\cdots W_1 x,
$$ 
parametrized by the weight matrices $W_j \in \bR^{d_j \times
d_{j-1}}$, for all $j \in [N]$.  We will denote the tuple of weight matrices by $\W = (W_1, \ldots, W_N)$ and the space of all such tuples by $\Theta$. 
This is the \emph{parameter space} of our model. To slightly simplify the notation we will also denote the input and output dimensions by $m \equiv d_0$ and $n \equiv d_N$, respectively, and write $W \defeq  W_{N}\cdots W_1$ for the end-to-end matrix. 
For any $1 \leqslant i \leqslant j \leqslant N$, we will also write $W_{j:i} \defeq W_j \cdots W_i$ for the matrix product of layer $i$ up to $j$. 
We note that the represented function is linear in the network input $x$, but the parametrization
is not linear in the parameters $\W$. We denote the network's parametrization map by \begin{multline*} \mu \colon \Theta \to \bR^{d_N \times d_0};\\ \W
=(W_1,\ldots, W_N) \mapsto W_{N:1} = W_N\cdots W_1. 
\end{multline*} 
The \emph{function space} of the network is the set of linear functions it can represent. 
This corresponds to the set of possible end-to-end matrices, which are the $n\times m$ matrices of rank at most $\ud \defeq \min\{d_0,\ldots,
d_N\}$.  When $\ud= \min\{d_0,d_N\}$, the function space is a vector space.
Otherwise, when there is a bottleneck such that $\ud< \min\{d_0,d_N\}$, it is a non-convex subset of $\mathbb{R}^{m\times n}$ determined by polynomial constraints, namely the vanishing of the $(\ud+1)\times (\ud+1)$ minors.

Next, we collect a few results on the gradient dynamics of linear networks for general differentiable losses, which have been established in previous works with focus on the squared error loss \citep{Kawaguchi2016Deep,Bah2021Learning,Chitour2022Geometric,Arora2018Optimization}. 
In the interest of conciseness, here we only provide the main takeaways and defer a more detailed discussion to~\Cref{app:linear-networks}. 
For the remainder of this section, let $\cL^1\colon \bR^{n \times m} \to \bR$ be \emph{any} differentiable loss and $\cL^N$ be defined through the parametrization $\mu$ as $\cL^N(\W) \defeq \cL^1 \circ \mu(\W)$. 
For such a loss, the gradient flow $t \mapsto \W(t)$ is defined by 
\begin{equation}
\begin{aligned}
    \label{eq:GF}
    \od{}{t}\W(t) &= - \nabla \cL^N(\W(t)) 
     \\
     & \iff
     \\
    \forall j \in [N],\quad \od{}{t}W_j(t)
                  &= - \nabla_{W_j} \cL^N(W_1(t), \ldots,
    W_N(t)). 
    \end{aligned}
\end{equation}
This governs the evolution of the parameters. 
Furthermore, we observe that the partial derivative of $\cL^N$ with respect to
$W_j$, for all $j \in [N]$, is given by 
\begin{equation} 
\begin{aligned}
        \nabla_{W_j}& \cL^N(W_1, \ldots, W_N) \\
        &= W_{j+1}^\top \cdots W_N^\top
        \nabla \cL^1(W) W_1^\top \cdots W_{j-1}^\top.
        \label{eq:GD}
\end{aligned}
\end{equation} 
As it turns out, the gradient flow dynamics preserves the difference of the Gramians of subsequent layer weight matrices, 
which are thus invariants of the gradient flow; i.e. 
$$
\od{}{t} (W_{j+1}(t)^\top W_{j+1}(t)) = \od{}{t} (W_{j}(t) W_{j}(t)^\top).  
$$
The notion of \emph{balancedness} for the weights of linear networks was first introduced by~\citet{Fukumizu1998Effect} in the shallow case and generalized to the deep case by~\citet{Du2018Algorithmic}. 
This is useful as it removes the redundancy of the parametrization when investigating the dynamics in function space and has been considered in numerous works. 

\begin{definition}[Balanced weights]
    \label{def:balanced-weights}
    The weights $W_1, \ldots, W_N$ are said to be \emph{balanced} if, for all~$j
    \in [N-1],~~ W_j W_{j}^\top = W_{j+1}^\top W_{j+1}$.
\end{definition}

Assuming balanced initial weights, if the flow of each $W_j$ is defined and bounded, then the rank of the end-to-end matrix $W$ remains constant during training~\citep[Proposition 4.4]{Bah2021Learning}. 
Moreover, the products $W_{N:1} W_{N:1}^\top$ and $W_{N:1}^\top W_{N:1}$ can be written in a concise manner; namely, $W_{N:1}W_{N:1}^\top = {(W_N W_N^\top)}^N$ and $W_{N:1}^\top W_{N:1} = {(W_1^\top W_1)}^N$, which simplifies computations.

\begin{remark} 
Some attempts to relax the balanced initial weights assumption include the notion of approximate balancedness \citet{Arora2019Convergence}, which only requires that there exists $\delta>0$ such that $\norm{W_{j} W_{j}^\top - W_{j+1}^\top W_{j+1}}_F\leqslant \delta$ for $j \in [N-1]$. 
Our proofs in this paper use exactly balanced initial weights for simplicity, but they would also work under the approximate balancedness setting. 
Further initializations have been proposed by e.g.\
\citet{GBL2019Implicit,Yun2021unifying}. 
We defer the analysis of such cases for future work favoring here a focused discussion of the Bures-Wasserstein loss with balanced initial weights. 
\end{remark}

\section{Wasserstein generative linear networks} 
\label{sec:wass-lin-gan}

\subsection{The Bures-Wasserstein loss}\label{sec:def-BW}

The Bures-Wasserstein (BW) distance is defined on the space of positive
semi-definite matrices (or \emph{covariance space}) $\cS_+(n)$. 
We collect definitions and key properties of the gradient. 
\begin{definition}[Bures-Wasserstein distance]
    \label{def:BW}
    Given two symmetric positive semidefinite matrices~$(\Sigma_0,\, \Sigma) \in
    {(\cS_+(n))}^2$, the squared Bures-Wasserstein distance between $\Sigma_0$ and $\Sigma$ is defined as 
    \begin{equation}
        \label{eq:BW}
        \cB^2(\Sigma, \Sigma_0) = \trc{\Big(\Sigma + \Sigma_0 - 2(\Sigma_0^{1/2}
    \Sigma \Sigma_0^{1/2})^{1/2}\Big)}.
\end{equation}
\end{definition} 
\citet[Lemma A.3]{Kroshnin2021Statistical} shows that the matrix square root is differentiable on the set of positive definite matrices. In turn, we can differentiate the BW distance at $\Sigma \in \cS_{++}(n)$. 
 However, the mapping $\Sigma \mapsto \cB^2(\Sigma, \Sigma_0)$ is not differentiable at all $n \times n$ matrices. 
Indeed, if we let $\Gamma Q \Gamma^\top$ be a spectral decomposition of $\Sigma_0^{1/2}\Sigma\Sigma_0^{1/2}$, then \eqref{eq:BW} can be written as 
\begin{equation}
    \label{eq:BW-2}
    \cB^2(\Sigma, \Sigma_0) = \norm{\Sigma^{1/2}}^2_F +
    \norm{\Sigma_0^{1/2}}^2_F - 2 \trc{Q^{1/2}}.
\end{equation}
Due to the square root on $Q$, the map $\Sigma \mapsto \cB^2(\Sigma, \Sigma_0)$ is not differentiable when the number of positive eigenvalues of $Q$, i.e.\ the rank of $\Sigma_0^{1/2}\Sigma \Sigma_0^{1/2}$, changes. More specifically, while one can compute the gradient over the set of matrices of rank $k$ for any given $k$, the norm of the gradient blows up if the matrix changes rank. 
We describe the gradient of $\cB^2$ restricted to the set of full-rank matrices in \Cref{app:BW}. 
We refer the reader to \citet{bhatia2019bures} for further details on the BW distance. 

\subsection{Linear Wasserstein GAN}  
\label{sec:lin-wgan}
The distance defined in~\eqref{eq:BW} corresponds to the 2-Wasserstein distance between two zero-centered Gaussians. It can be used as a loss for training models of Gaussian distributions, in particular generative linear networks. 
Recall that a zero-centered Gaussian distribution is completely specified by its covariance matrix. 
Given a bias-free linear network and a latent Gaussian distribution $\cN(0, I_m)$, a linear network generator computes a push-forward of the latent distribution, which is again a Gaussian distribution. 
If $Z \sim \cN(0, I_m)$ and $X = WZ$, then 
$$
X \sim
\cN(0, WW^\top) \eqdef \nu. 
$$
Given a target distribution $\nu_0 = \cN(0, \Sigma_0)$ (or simply a covariance matrix $\Sigma_0$, which may be a sample covariance matrix), 
one can select $W$ by minimizing $\cB^2(WW^\top, \Sigma_0) = \cW_2^2(\nu, \nu_0)$. %
We will denote the map that takes the end-to-end matrix $W$ to the covariance matrix $WW^\top$ by $\pi\colon \bR^{n\times m} \to \bR^{n\times n}; \; W \mapsto WW^\top$. 

\paragraph{Loss in covariance, function, and parameter spaces} 
We consider the following losses, which differ only on the choice of the search variable, taking either a covariance space, function space, or parameter space viewpoint. 
\begin{itemize}[leftmargin=*]
\item First,  we denote the loss over covariance matrices $\Sigma \in \cS_+(n)$
    as  $L\colon \Sigma
    \mapsto  \cB^2(\Sigma, \Sigma_0)$. 
\item 
    Secondly, given $\pi \colon W \mapsto WW^\top \in
    \cS_+(n)$, we define the loss in the function space, i.e.\ over end-to-end matrices $W \in \bR^{n\times m}$, as $L^1 \colon  W \mapsto  L \circ \pi(W)$. 
This is given by, for $W \in \bR^{n \times m}$, 
\begin{equation}\label{eq:L1}
    \quad L^1(W) = \trc{\Big(WW^\top + \Sigma_0 - 2(\Sigma_0^{1/2} WW^\top \Sigma_0^{1/2})^{1/2}\Big)}. 
\end{equation}
This loss is \emph{not} convex in $\bR^{n \times m}$, which can be seen even in the scalar case.

\item Lastly, for a tuple of weight matrices $\W = (W_1, \ldots, W_N)$, we
    compose $L^1$ with the parametrization map $\mu \colon \W \mapsto W_{N:1}$
    to define the loss in the parameter space as $L^N \colon \W \mapsto L \circ
    \pi \circ \mu(\W)$, for $\W \in \Theta$.  Observe that this is, again, a
non-convex loss.  \end{itemize} Thus, for $\W \in \Theta$, $L^N(\W) =
L^1(\mu(\W))= L(\pi( \mu(\W)) =  \cB^2(\pi\circ \mu(\W), \Sigma_0)$.  While the
gradient flow~\eqref{eq:GF} is defined on the parameters $\W$, viewing the
problem in the covariance space is useful since then the objective function is
convex, even if it may be subject to non-convex constraints.  One of our goals
is to translate properties between $L$, $L^1$, and $L^N$.

\paragraph{Smooth perturbative loss} 
As mentioned before, the Bures-Wasserstein loss is non-smooth at covariance
matrices with vanishing eigenvalues.  As a result, the usual analysis tools to
prove uniqueness and convergence of the gradient flow do not apply here.  To
tackle this issue, we introduce a smooth perturbative version of the loss.
Consider the perturbation map $\varphi_\tau \colon \Sigma \mapsto  \Sigma + \tau
I_n$, where $\tau>0$ plays the role of a regularization strength.  Then the
perturbative loss in the covariance space is defined as $L_\tau = L \circ
\varphi_\tau$, and the perturbative loss in the function space as $L^1_\tau =
L_\tau \circ \pi$.  More explicitly, we let 
\begin{equation}\label{pertloss} 
\begin{aligned}
    L^1_{\tau}(W) &= \trc\Big(W W^{\top} + \tau I_n + \Sigma_0\\
& -2(\Sigma_0^{1/2}(WW^\top+\tau I_n)\Sigma_0^{1/2})^{1/2}\Big).
\end{aligned}
\end{equation}
This function is smooth and allows us to apply usual convergence arguments for the gradient flow. 
Likewise, $L^N_\tau \defeq L_\tau
\circ \pi \circ \mu$ is well-defined and smooth on $\Theta$. 

\begin{remark}
The perturbative loss~\eqref{pertloss}, as well as the original loss on
fixed-rank matrices, are differentiable. 
    Many results of \citet{Bah2021Learning} can be carried over for these
    differentiable Bures-Wasserstein losses. 
    For example, the uniform boundedness at any time $t \geqslant 0$ of the end-to-end matrix holds,
    $\norm{W(t)} \leqslant \sqrt{2L^1(W(0)) + \trc\Sigma_0}$.
    Similar observations may apply for the case of $L^1$ in the case that the matrix $WW^\top$ remains positive definite throughout training, in which case the gradient flow 
    remains well-defined and the loss is monotonically decreasing. 
    We expand on this in~\Cref{app:linear-networks}. 
\end{remark}

The next lemma, proved in \Cref{prf:diff-losses}, provides a quantitative bound
for the difference between the original and the perturbative loss. To compare
the two losses, we set the parameters --- and hence, the end-to-end matrices ---
to a fixed, common value.

\begin{lemma}\label{lem:diff-losses}
    Let $W\in\bR^{n\times m}$ and  $\tau > 0$.  Assume that $\rank{(\Sigma_0)} =
    n$. Then, with $L^{1}(W)$ given by (\ref{eq:L1}) and $L^1_\tau(W)$ given by
    (\ref{pertloss}), we have 
    \begin{equation}\label{eq:bound} |L^1_\tau(W) -
        L^1(W)| \leqslant n\sqrt{\tau} \left(\sqrt{\tau}+
        \frac{2\lambda_{\max}(\Sigma_0^{1/2})}{\lambda_{\min}(\Sigma_0^{1/2})}\right),
    \end{equation} 
    with $(\lambda_{\max}(\Sigma_0^{1/2}), \lambda_{\min}(\Sigma_0^{1/2}))$ the
    maximum and minimum eigenvalues of~$\Sigma_0^{1/2}$. 
\end{lemma}
We observe that the upper bound \eqref{eq:bound} is tight in $\tau$ in the sense that it goes to zero as $\tau$ goes to zero.

\section{Critical points}
\label{sec:critical-points}

In this section, we characterize the critical points of the Bures-Wasserstein loss restricted to matrices of a given rank. 
The proofs of results in this section are given in \Cref{app:proofs-critical-points}. 

For $k \in \bN$, denote $\cM(k)$ as the manifold of rank-$k$ matrices of size $n \times m$: 
\begin{equation}
\cM(k) \equiv \cM(k; n, m) \defeq \{ W \in \bR^{n\times m} \mid \rank{W} = k\}. \label{eq:def-Mk} 
\end{equation}
Similarly, we denote by $\cM({\leqslant k}) \equiv \cM(\leqslant k; n, m)$ the set of $n \times m$ matrices of rank at most $k$. 
The manifold $\cM(k)$ is an embedded submanifold of the linear space $(\bR^{n\times m}, \ang{\cdot, \cdot}_F)$, with codimension $(n-k)(m-k)$ (\citealt[Proposition 4.5]{HS1995Critical};\ \citealt[\S 2.6]{boumal2022intromanifolds}).  
Given a function $f \colon \bR^{n\times m} \to \bR$, its 
\emph{restriction} on $\cM(k)$ is denoted by $f|_{\cM(k)} \colon \cM(k) \ni W \mapsto f(W)$. 
A function $f$ may not differentiable everywhere on $\mathbb{R}^{n\times m}$ but still have a restriction on $\cM(k)$ that is differentiable.

\begin{definition}
\label{def:crit-restricted}
Let $\cM$ be a smooth manifold. Let $f \colon \bR^{n\times m} \to \bR$ be any
function such that its restriction on $\cM$ is differentiable. A point $W \in
\cM$ is said to be a \emph{critical point of $f|_{\cM}$} if the differential of $f|_{\cM}$ at $W$ is the zero function, i.e.\ $\dif{}f|_{\cM}(W) = 0$. 
\end{definition}

\subsection{Critical points of \texorpdfstring{$L^1$}{L1} over \texorpdfstring{$\cM(k)$}{Mk}} 

Given a matrix $A \in \bR^{n \times n}$ and a set $\cJ_k \subseteq [n]$, where the subscript indicates the cardinality of the set, $k= |\cJ_k|$, 
we denote by $A_{\cJ_k} \in \bR^{n \times k}$ the sub-matrix of $A$ consisting of the columns with index in $\cJ_k$. 
If the matrix $A$ is diagonal, we let $\bar{A}_{\cJ_k} \in \bR^{k \times k }$ be the diagonal sub-matrix which extracts the rows and columns with index in $\cJ_k$. 
The following result characterizes the critical points of the loss in function space. 
It can be regarded as a type of Eckart-Young-Mirsky result for the case of the Bures-Wasserstein loss. 

\begin{theorem}[Critical points of $L^1$]
    \label{thm:critical-points-L1}
        Assume $\Sigma_0$ has $n$ distinct, positive eigenvalues.   Let $
        \Sigma_0 = \Omega \Lambda \Omega^\top $ be a  spectral decomposition of
        $\Sigma_0$ (so $\Omega \in \cO(n)$), with eigenvalues ordered
        decreasingly.  Let $k \in [\min\set{n, m}]$.  A matrix $W^* \in \cM(k)$ is a critical point of $L^1|_{\cM(k)}$ if and only if $W^* = \Omega_{\mathcal{J}_k} \bar\Lambda_{\mathcal{J}_k}^{1/2} V^\top$ for some $\mathcal{J}_k \subseteq [n]$ with $|\cJ_k| = k$ and $V \in
        \bR^{m \times k}$ with $V^\top V = I_k$. 
        The minimum over $\cM(\leqslant k)$ is attained precisely when $\cJ_k=[k]$. 
        In particular, $\inf_{\cM(k)}L^1(W) = \min_{\cM(k)}L^1(W)$ and $\min_{\cM(k)} L^1(W) =\min_{\cM(\leqslant k)} L^1(W)$. 
\end{theorem}

\begin{remark}
    Notice that %
    there are $\binom{n}{k}$ critical points %
    up to right rotation by an arbitrary orthonormal matrix (the trivial symmetry of $W\mapsto WW^\top$). 
\end{remark}

    The proof relies on evaluating the zeros of the gradient $\nabla
    L^1|_{\cM(k)}$ (see \Cref{lem:gradient-L1-svd}). 
    Then evaluating the loss at these critical points allows us to identify which of them attain the minimum. 

\begin{remark} 
Interestingly, the  critical points and the minimizer of $L^1$ characterized in the above result agree with those of the squared error loss \citep{Eckart1936approximation, mirsky1960symmetric}. 
    Nonetheless, we observe that \eqref{eq:BW} is only defined for positive
    semi-definite matrices. Hence the notion of unitary invariance considered by
    \citet{mirsky1960symmetric} only makes sense for left and right
    multiplication by the same matrix. Moreover, while we can establish unitary
    invariance for a variational expression of the distance (see
    \Cref{lem:variational-BW}), this is still not a norm in the sense that there
    is no function $B \colon \cS_+(n) \to \bR$ such that  $\cB(\Sigma,\Sigma_0)= B(\Sigma-\Sigma_0)$, and hence it does not fall into the framework of \citet{mirsky1960symmetric}. 
    We offer more details about this in Appendix~\ref{app:BW}. 
\end{remark}

\subsection{Critical points of the perturbative loss}
\label{sec:crit-points-pertloss}

For the critical points of the perturbative loss $L_{\tau}^1(W)$ we obtain the following results. 

\begin{theorem}[Critical points of $L^1_\tau$]
    \label{thm:critical-points-L1tau}
    Assume $\Sigma_0$ has $n$ distinct, positive eigenvalues. Let $\Sigma_0 = \Omega \Lambda \Omega^\top $ be a spectral decomposition of $\Sigma_0$, with eigenvalues ordered decreasingly. A point $W^* \in \cM(k)$ is a critical point of $L^1_\tau|_{\cM(k)}$ if and only if 
    $W^* = \Omega_{\cJ_k} {(\bar{\Lambda}_{\cJ_k} - \tau
    I_k)}^{1/2} V^\top$ for some $\cJ_k \subseteq [n]$ with $|\cJ_k| =
k$ and %
$V \in \bR^{n \times k}$ 
with $V^\top V=I_k$. %
Moreover, the value at such a point is
    $L^1_\tau(W^*) = \sum_{j \not\in \cJ_k} (\sqrt{\lambda_j} -
    \sqrt{\tau})^2$.
The %
minimum over $\cM(\leqslant k)$ is therefore attained precisely when $\cJ_k = [k]$. 
In particular, $\inf_{\cM(k)}L^1_\tau(W) = \min_{\cM(k)}L^1_\tau(W)$ and $\min_{\cM(k)} L^1_\tau(W) =\min_{\cM(\leqslant k)} L^1_\tau(W)$. 
\end{theorem}

Note that the above characterization of the critical points imposes an upper bound on $\tau$: for a given $W^*$ to be a critical point, one must have that 
$\tau \leq \lambda_j$ for all $j\in \cJ_k$, because the eigenvalues of 
$\bar{\Lambda}_{\cJ_k} - \tau
I_k$ have to be nonnegative.

In order to link the critical points 
in the parameter space to the critical points in the function space, we appeal to
the correspondence drawn by \citet[Propositions 6 and 7]{Trager2020Pure}.  
For the Bures-Wasserstein loss, this allows to conclude the following. 

\begin{proposition}[Critical points in parameter space are critical points in 
function space]
    \label{prop:crit-param-func}
    Assume a full-rank target $\Sigma_0$ with spectral decomposition $\Sigma_0 =
    \Omega \Lambda \Omega^\top$ and distinct eigenvalues $\lambda_1 > \cdots >
    \lambda_n > 0$ ordered decreasingly. Let $\tau \in (0, \lambda_n]$. If $\W^*
    \in \Crit(L^N_\tau)$, then $W^* = \mu(\W^*)$ is a critical point of the loss
    $L^1_\tau|_{\cM(k)}$, where $k = \rank W^*$. Moreover, when $k = \ud = \min_{i
    \in [N]} \set{d_i}$, then $\W$ is a local minimizer of the loss $L^N_\tau$
    if and only if $W^* = \mu(\W^*)$ is a local minimizer, and therefore the
    global minimizer, of $L^1_\tau|_{\cM(\ud)}$. In this case,
    $\Sigma^*_\tau  = W^*{W^*}^\top + \tau I_n$ is the $\tau$-best $\ud$-rank
    approximation of the target in the covariance space, in the sense that
    $\Sigma^*_\tau = \Omega \begin{pmatrix} \Lambda_{[\ud]} & \\ & \tau
    \end{pmatrix} \Omega^\top$. 
\end{proposition}

\Cref{prop:crit-param-func} ensures that, under the assumption that the
solution of the gradient flow is a (local) minimizer in the parameter space
and has the highest possible rank $\ud$ for the given network architecture, the
solution in the covariance space is the best $\ud$-rank approximation of the
target in the sense of the $\tau$-smoothed Bures-Wasserstein distance.  The
fact that  any local minimizer of $L^1_\tau|_{\cM(\ud)}$ is indeed a global minimizer is not immediate (since neither the loss $L^1_\tau$ nor the set
$\cM(\ud)$ are convex), but can be shown as we do in
\Cref{lem:strict-saddle-L1tau}.

\begin{remark}
    \label{rem:full-rank-limit}
    Under the balancedness assumption, one can show that the rank of the
    end-to-end matrix does not drop during training \citep[Proposition
    4.4]{Bah2021Learning}, and that the trajectory almost surely escapes the strict saddle points \citep[Theorem 6.3]{Bah2021Learning}. 
    If the initialization of the network has rank $\ud$, the matrices $W(t),\, t >0$, maintain rank $\ud$
    throughout training.  There can be issues in the limit, since $\cM(\ud)$ is
    not closed. Proving whether or not %
    the limit point also belongs to $\cM(\ud)$ is an
    interesting open problem. %
\end{remark}

\section{Convergence analysis} 
\label{sec:convergence}

The Bures-Wasserstein distance can be viewed through the lens of the Procrustes metric \citep{dryden2009non, masarotto2019procrustes}. In fact, it can be obtained by the following minimization problem. 
\begin{lemma}[{\citealt[Theorem 1]{bhatia2019bures}}]
\label{lem:variational-BW}
For $(\Sigma, \Sigma_0) \in {(\cS_+(n))}^{2}$, 
\begin{equation}
    \label{eq:variational-BW}
    \cB^2(\Sigma, \Sigma_0) = \min_{U \in \cO(n)}\norm{\Sigma^{1/2} - \Sigma_0^{1/2} U}_F^2,  
\end{equation} 
where $\cO(n)$ denotes the set of $n\times n$ orthogonal matrices. 
Moreover, the minimizer $\bar U$ occurs in the polar decomposition of $\Sigma^{1/2}\Sigma_0^{1/2}$. 
\end{lemma}

We emphasize that in the above description of the Bures-Wasserstein distance, the minimizer $\bar U$ depends on $W$, so that $\cB^2$ fundamentally differs from a squared Frobenius norm. 
Moreover, the square root on $\Sigma$ can lead to singularities when differentiating the loss. 
Nonetheless, based on~\eqref{eq:variational-BW} we can formulate the following deficiency margin concept to avoid such singularities. 
\begin{definition}[Modified deficiency margin]
    \label{def:MDM}
    Given a target matrix $\Sigma_0\in\bR^{n\times n}$ and a positive constant $c>0$, we say that $\Sigma\in\bR^{n\times n} $ has a modified deficiency margin $c$ with respect to $\Sigma_0$ if 
    \begin{equation}\label{eq:c-bound}
        \min_{U\in \cO(n)}\norm{\Sigma^{1/2} - \Sigma_0^{1/2}U}_F\leqslant \sigma_{\min}(\Sigma_0^{1/2})-c.
    \end{equation}
\end{definition}
With a slight abuse of terminology, we will say that $W$ has a modified deficiency margin if $WW^\top$ does. The deficiency margin idea can be traced back to \citet{Arora2019Convergence}. 
Note that we can write $\sqrt{WW^{\top}} = \Sigma^{1/2}$, and this square root
can be realized by Cholesky decomposition. 
If we initialize the parameters so that $\Sigma$ is close to the target $\Sigma_0$, then (\ref{eq:c-bound}) holds trivially. 
In fact, if the initial value $W(0)$ satisfies the modified deficiency margin condition, then %
the least singular value of $W(t)$ remains bounded below by $c$ 
along %
gradient flow or gradient descent trajectories with decreasing $L^N$: 
\begin{lemma}\label{lem:marginc}
    Suppose $W(0) W(0)^{\top}$ has a modified deficiency margin $c$ with respect to $\Sigma_0$. Then 
    \begin{equation}
        \label{eq:smin-bound}
        \sigma_{\min}\Big(\sqrt{W(t) W(t)^{\top}}\Big)\geqslant c,\quad ~\text{for}~t\geqslant 0.
    \end{equation}
\end{lemma}
The proof of this and all results in this section are provided in
\Cref{app:proofs-convergence}. We note that, while the modified deficiency margin assumption is sufficient for \Cref{lem:marginc} to hold, it is by no
means necessary. We will assume that the modified  deficiency margin assumption
holds for simplicity of exposition, but the gradient flow analysis in the
next paragraph only requires the less restrictive \Cref{lem:marginc} to hold.

\subsection{Convergence of gradient flow for the smooth loss} 

Since we cannot exclude the possibility that the rank of $WW^{\top}$ drops along the gradient flow of the BW loss, we consider the smooth perturbation introduced in \Cref{sec:lin-wgan} as a way to avoid singularities. 
We consider the gradient flow \eqref{eq:GF} for the perturbative loss. %
The gradient of \eqref{pertloss} is %
\begin{equation*}
    \begin{split}
           &\nabla L^1_\tau (W) = \\
           &2\Big(W -\Sigma_0^{1/2} \big(\Sigma_0^{1/2} (WW^\top +\tau I_n) \Sigma_0^{1/2}\big)^{-1/2} \Sigma_0^{1/2} W\Big). 
       \end{split}
\end{equation*}

The perturbation $\tau I_n$ 
ensures that $\lambda_{\min}(\Sigma_\tau) \geqslant \tau > 0$, which in turn makes $L_\tau$ strongly-convex, as shown next.  

\begin{lemma}
    \label{lem:strong-cvx}
    The Hessian operator $\bG_{\tau}$ of the loss $L_\tau$ at $ \Sigma \in \cS_{+}(n)$ satisfies
    $\lambda_{\min} (\bG_{\tau}) \geqslant K_{\tau}$ for any $\Sigma \in
    \cS_{+}(n)$, with $K_{\tau} \defeq \frac{\sqrt{\tau \lambda_{\min}(\Sigma_0)}}{2
    C_{\tau}^2}$, where $C_{\tau} \defeq 2(L_\tau(\Sigma(0)) + \trc{\Sigma_0})$.
\end{lemma}
This is proven in \cref{lem:bound-hessian-G}.

Let us denote the minimizer of the perturbative loss $L(\Sigma_{\tau})$ by $\Sigma_{\tau}^{*}$. 
Equipped with the strong convexity of the loss $L_\tau$ given by Lemma~\ref{lem:strong-cvx}, we are ready to show
that the gradient flow has convergence rate $O(e^{-\tilde{K}_{c, N}K_\tau t})$
to the global minimizer of $L_\tau$, 
where $K_\tau$ is the
constant from the Hessian bound given by Lemma~\ref{lem:strong-cvx}, and
$\tilde{K}_{c,N}$ is a constant which depends on the modified margin deficiency
and the depth of the network. 
Recall that for $t \geqslant 0$, $\Sigma_{\tau}(t) =
W_{N:1}(t)W^{\top}_{N:1}(t) + \tau I_n$, so we prove convergence of gradient flow on
the loss under the parametrization $\Sigma_\tau(t) = \varphi_\tau(\pi(\mu(\W(t))))$.
\begin{theorem}[Convergence of gradient flow for the smooth loss]
    \label{thm:convergenceGF}
Let $\Delta_{\tau}^{*} \defeq \Sigma_{\tau}(0) - \Sigma_{\tau}^{*}$ be the
distance from the initialization %
to the minimizer $\Sigma_\tau^* \defeq 
\argmin_{\Sigma \in \cS_{++}(n)} L_\tau(\Sigma) = \Sigma_0 - \tau I_n$.
    Assume both balancedness (\Cref{def:balanced-weights}) and the
    modified deficiency margin (\Cref{def:MDM}). 
    Then the gradient flow $\dot \W(t) =
    -\nabla L^N_\tau(\W(t))$ converges as 
    \begin{equation}\label{GFconvpert}
        L(\Sigma_{\tau}(t)) -L(\Sigma_{\tau}^{*})\leqslant
        e^{-8N c^{\frac{2(2N-1)}{N}}K_\tau t}\Delta_{\tau}^{*},
    \end{equation}
    where $K_\tau = \frac{\sqrt{\tau \lambda_{\min}(\Sigma_0)}}{2 C_{\tau}^2}$ is the
    strong convexity parameter from \Cref{lem:strong-cvx}, with $C_{\tau} = 
    2(L(\Sigma_\tau(0)) + \trc(\Sigma_0))$.
\end{theorem} 

\begin{remark} \label{rem:GForig}
    Under the modified margin assumption (\Cref{def:MDM}), the
    parametrized covariance matrix $\Sigma = WW^\top$ has its eigenvalues lower-bounded by $c^2$ at all times, 
    as per \Cref{lem:marginc}. Therefore, the
    convergence result obtained in \Cref{thm:convergenceGF} can be extended to
    the original loss, with $(\Sigma_\tau(t), \Sigma_\tau^*)$ replaced with
    $(\Sigma(t), \Sigma^*)$, $\Delta_\tau^*$ replaced with $\Delta^* \defeq
    \Sigma(0) - \Sigma^*$, $K_\tau$ replaced with $K_{c^2} = \frac{\sqrt{c^2 \lambda_{\min}(\Sigma_0)}}{2 C^2}$, and $C_\tau$
    replaced with $C_0 =  2(L(\Sigma(0)) + \trc(\Sigma_0))$. More details about this are given in \Cref{sec:proofgradflowconv}.
\end{remark}

\subsection{Convergence of gradient descent for the BW loss}
Assuming that the initial end-to-end matrix $W$ 
has a modified deficiency margin, we can establish the following convergence
result for gradient descent with finite step sizes, which is valid for both the
perturbed loss and the original (non-perturbed) loss. 
Given an initial value $\W(0)$, we consider the gradient descent iteration 
\begin{equation} 
    \W(k+1) = \W(k) - \eta \nabla L^N(\W(k)), \quad k=0,1,\ldots, 
    \label{eq:GDupdate}
\end{equation} 
where $\eta>0$ is the learning rate or step size and the gradient is given by~\eqref{eq:GD}. 

\begin{theorem}[Convergence of gradient descent]
\label{thm:gradientdescent}
    Assume that the initial values $W_i(0)$, $1\leqslant i\leqslant N$, are balanced and $W(0) = W_{N:1}(0)$ has a modified deficiency margin $c$. 
    If the learning rate $\eta>0$ satisfies
    \begin{equation*}
        \eta \leqslant \min\left\{\frac{c^2}{8M\sqrt{L^1(W(0))}},~ \frac{N c^{\frac{2(N-1)}{N}}}{2\Delta},~ \frac{1}{4Nc^{\frac{2(N-1)}{N}}}\right\},
    \end{equation*}
    where $\Delta= \frac{2^{N+1}}{c^{2N}}N^2M^{(4N-3)/N}\lambda^{1/2}_{\max}(\Sigma_0)+8N(N-1)M^{(3N-4)/N}\Big(M^{1/N}+\norm{\Sigma_0^{1/2}}_F\Big)$, and $M=\sqrt{2{L^1(W(0)) + \norm{\Sigma_0^{1/2}}^2_F}}$, 
    then, for any $\epsilon>0$, one achieves $\epsilon$ loss by the gradient descent \eqref{eq:GDupdate} at iteration 
    \begin{equation*}
        k\geqslant \frac{1}{2\eta  Nc^{\frac{2(N-1)}{N}}}\log\left(\frac{
    L^1(W(0))}{\epsilon}\right).
    \end{equation*}
\end{theorem}

\begin{remark}
    Theorems~\ref{thm:convergenceGF} and~\ref{thm:gradientdescent} show that the depth of the network can accelerate the convergence of the gradient algorithms. 
    We verify this experimentally in \Cref{sec:exp}.
\end{remark}

\subsection{Experimental evaluation of the convergence rate}
\label{sec:exp}

\begin{figure}[t]
    \centering
    \begin{subfigure}[t]{.465\columnwidth}
    \centering
    \includegraphics[width=\columnwidth]{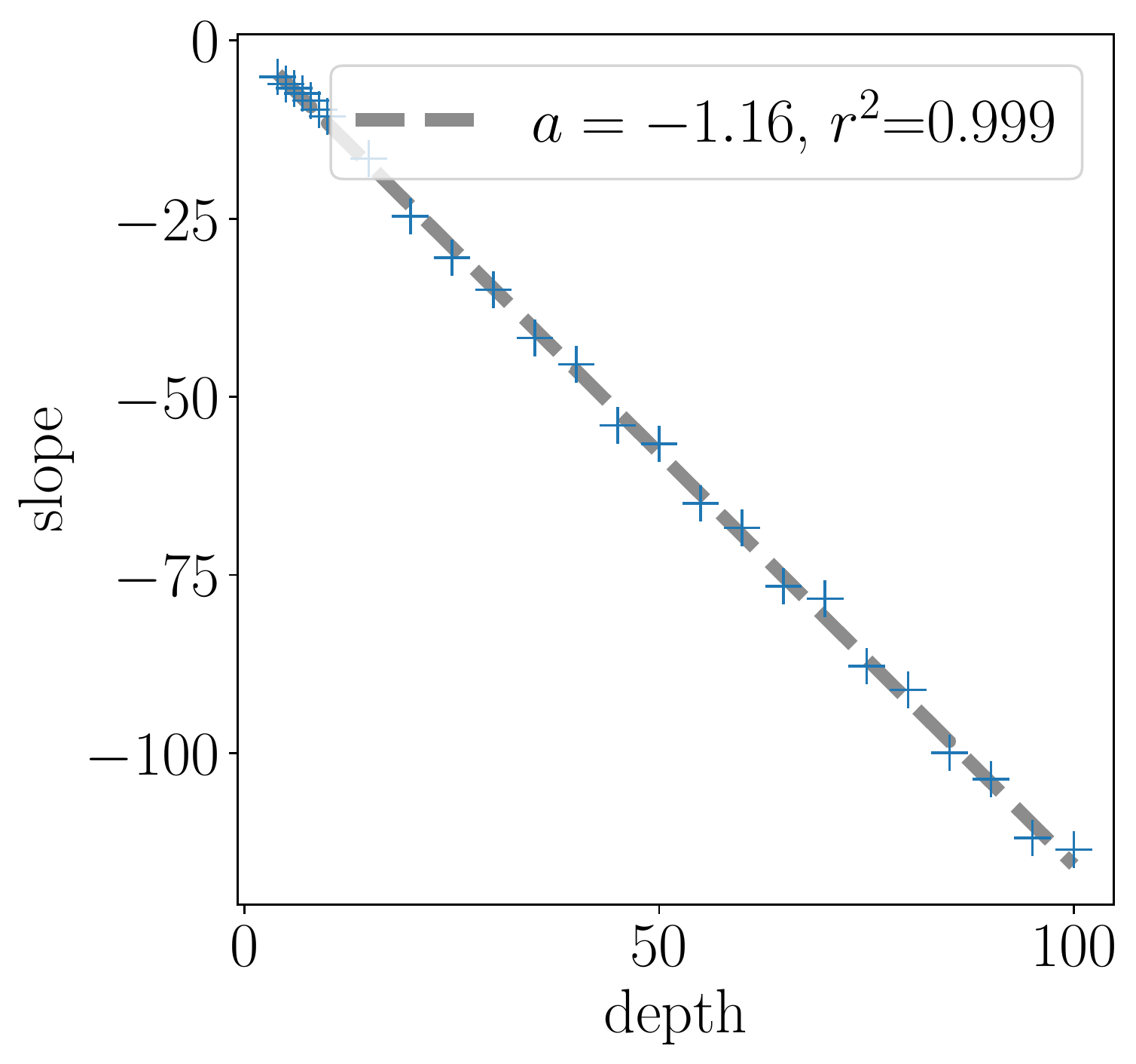}
    \caption{}
    \label{fig:slope-a}
\end{subfigure}
\begin{subfigure}[t]{.523\columnwidth}
    \centering
    \includegraphics[width=\columnwidth]{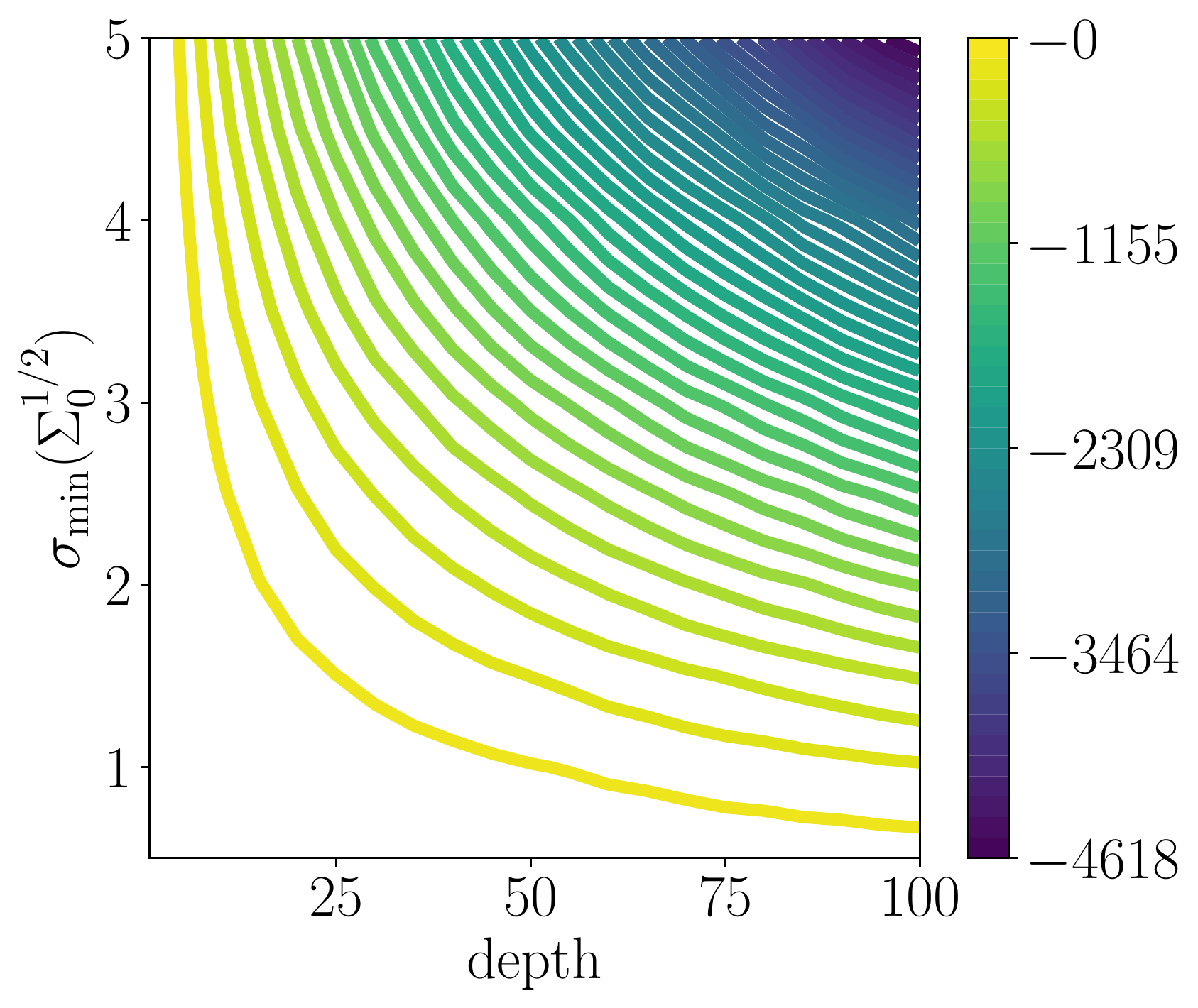}
    \caption{}
    \label{fig:slope-b}
\end{subfigure}
\caption{Logarithm of the linear convergence rate as a function of the depth $N$ and the minimum singular value 
$\sigma_{\min}(\Sigma_0^{1/2})$. 
In (a), $\sigma_{\min}(\Sigma_0^{1 / 2}) \approx 0.7078$ is
fixed; convergence rate and its linear regression as a function of the depth
$N$. In (b), both the depth and $\sigma_{\min}$ are varying, and the rate is
shown in a contour plot. The hyperbolas indicate a linear dependency on both the
depth $N$ of the network and the minimum singular value of the target square
root $\sigma_{\min}(\Sigma_0^{1 / 2})$, which is coherent with the
upper-bound given by \cref{thm:convergenceGF}.}
\label{fig:slope}
\end{figure}

We conduct numerical experiments to
illustrate our theoretical results\footnote{The source code for the experiments
    can be found at
\href{https://github.com/brechetp/BW-linear-networks}{https://github.com/brechetp/BW-linear-networks}.}. 
We observe empirically (\cref{fig:slope}) the linear dependency of the
asymptotic rate of convergence as a function of the depth of the network $N$ and
the minimum singular value square root of the target $\sigma_{\min}(\Sigma_0^{1/2})$.

\paragraph{Setup}

The target covariance matrix is sampled as $\Sigma_0 \defeq \Omega \Lambda
\Omega^\top$, where $\Omega \in \bR^{n \times n}$ is a random orthogonal matrix,
and the eigenvalues in $\Lambda$ follow a Zipf distribution, $\lambda_j \propto
{1}/{j}$ for $j \in [n]$. The input data dimension is set to be $n = 20$.  We vary the minimum
eigenvalue for the target $\lambda_{\min}$ and set $\lambda_j  = n/j \cdot
\lambda_{\min}$ for $j \in [n]$.  We consider constant width networks with $d_i = n
= m = 20$, for each $i \in [N]$.

To fulfill the modified deficiency margin assumption (\cref{def:MDM}), we
initialize the parameters close to the target $\Sigma_0$. If $\Sigma_0 = \Omega
\Lambda \Omega^\top$, then the weights are initialized in a way such that the
initial covariance matrix is $\Sigma(0) =  (\Sigma_0 - \tau I_n) + A$,
with $A$ being a random perturbation. More precisely, we choose $A = \Gamma D
\Gamma^\top$, where $\Gamma$ is a random orthogonal matrix, and $D$ is a
diagonal matrix with small eigenvalues  --- so that the overall distance between
the initialization and the target is bounded by $\sigma_{\min} - c$, for some $c
>0$.  With this initialization and the balancedness protocol explained
by~\citet{Arora2019Convergence}, the network satisfies both the balancedness and
modified deficiency margin assumptions.  In this case we expect
\cref{thm:convergenceGF} to hold for small step sizes. We estimate the
asymptotic linear convergence rate numerically.

\paragraph{Result} We compute the rate of convergence as follows.  First, the
network (initialized as detailed above) is trained with a small enough learning
rate $\eta$. Then, we compute $\log{(L(\Sigma(t) - L(\Sigma^*)))}$.
\Cref{thm:convergenceGF} states that this should be a linear function of the
time $t$.  Therefore, a linear regression is performed, and the slope taken as
the empirical rate of convergence.  According to~\cref{thm:convergenceGF}, this
rate should be linear in the depth $N$ and linear in the strong convexity
parameter $K_\tau$, which suggests that it could be linear in
$\sigma_{\min}(\Sigma_0^{1/2}) \equiv \sigma_{\min}$. 
Hence, we compute the empirical rate of convergence for varying depths and $\sigma_{\min}$, reported in~\cref{fig:slope}. 
In~\cref{fig:slope-a} the linear dependence in the depth is clearly visible, and
\cref{fig:slope-b} indicates a linear dependence in $\sigma_{\min}$ too.  Our
\Cref{thm:convergenceGF} only provides an upper bound on the convergence rate
and hence we compare with the empirical rates. The results suggest that this is
indeed the actual behavior in practice. 

\section{Conclusion} 

In this work, we studied the training of generative linear neural networks using
the Bures-Wasserstein distance. We characterized the critical points and
minimizers of this loss in function space,  over the set of matrices of fixed
rank $k$.  We introduced a smooth approximation of the BW loss, obtained by
regularizing the covariance matrix, and characterized its critical points in
function space as well.  Furthermore, under the assumption of balanced initial weights
satisfying a modified deficiency margin condition, we established a convergence
guarantee to the global minimizer for the gradient flow of both losses, with exponential rate of convergence.  Finally, we also considered the finite
step-size gradient descent optimization and established a linear convergence
result for both losses too, provided the step size is small enough depending on
the modified deficiency margin condition.  We collect our results
in~\Cref{tab:sumup-crit-pnts,tab:sumup-conv} in~\cref{app:table}.  These results contribute to the ongoing efforts to better characterize the optimization problems that arise in learning with deep neural networks beyond the commonly considered discriminative settings with the square loss.  

In future work, it would be interesting to refine our characterization of critical points of the Bures-Wasserstein loss in the parameter space, and to relax the modified deficiency margin condition that we invoked in order to establish our convergence results, as this constrains the parametrization to be of full rank.

\subsection*{Acknowledgments} 
This project has been supported by DFG SPP~2298 grant 464109215, ERC Starting
Grant 757983, and BMBF in DAAD project 57616814. GM has been supported by NSF
CAREER 2145630 and NSF 2212520. We would like to warmly thank the anonymous
reviewers for their questions and helpful comments. 

\bibliography{references}
\bibliographystyle{icml2023}

\newpage
\appendix
\onecolumn
\section*{Appendix} 

 The appendix is organized as follows. 
 \begin{itemize}[leftmargin=*]
     \item Appendix~\ref{app:table} gives a quick summary of the different geometrical and convergence results. 
     \item Appendix~\ref{app:BW} provides background on the Bures-Wasserstein loss and related optimal transport topics. 
     \item Appendix~\ref{app:linear-networks} presents general properties of linear neural networks and classical results on convergence in parameter space. 
     \item Appendix~\ref{app:proofs-critical-points} presents the proofs of results about critical points from~\Cref{sec:critical-points}. 
     \item Appendix~\ref{app:proofs-convergence} presents the proofs of results about convergence from~\Cref{sec:convergence}. 
     \item Appendix~\ref{app:hessian-losses} evaluates the Hessian of the loss.
 \end{itemize}

 \section{Summary of the results}
 \label{app:table}
 
\Cref{tab:sumup-crit-pnts,tab:sumup-conv} present a summary of the results obtained in this paper. 

\begin{table}[h!]
    \centering
    \begin{tabular}{cccc}
        \toprule
        Loss & Parametrization & Critical points  & Ref\\
        \midrule
        $L^1$ & $W$ & $\Omega_{\cJ_k} \bar{\Lambda}_{\cJ_k}^{1/2} V^\top$ &  \Cref{thm:critical-points-L1} \\
        $L^1_\tau$ & $W$ & $\Omega_{\cJ_k} {(\bar{\Lambda}_{\cJ_k} - \tau I_k)}^{1/2} V^\top$ &\Cref{thm:critical-points-L1tau} \\
        \bottomrule
    \end{tabular}
    \caption{Summary of critical point results. The target is assumed full rank with
    distinct eigenvalues and spectral decomposition $\Sigma_0 = \Omega \Lambda \Omega^\top$. 
    Here $V \in \bR^{m \times k}$ is any semi-orthogonal matrix and $\cJ_k \subset [n]$ is an index set of cardinality $k$. 
}
    \label{tab:sumup-crit-pnts} 
\end{table} 

\begin{table}[h!]
    \centering
    \begin{tabular}{cccccc}
        \toprule
        Loss & Parametrization &  Initialization &
        Convergence rate & Ref\\
         \midrule
        $L^N_\tau $ & $\W$ &   Balanced, MDM & GF:
        Exponential & \Cref{thm:convergenceGF} \\
        $L^N$ & $\W$ &  Balanced, MDM & GD: $O(\log(1/\epsilon))$ & \Cref{thm:gradientdescent} \\
        \bottomrule
    \end{tabular}
    \caption{Summary of  convergence results. 
    Here ``Balanced'' stands for balanced weights (\Cref{def:balanced-weights}), ``MDM'' stands for modified deficiency margin (\Cref{def:MDM}), and $\epsilon$ is the precision we want to achieve
(\Cref{thm:gradientdescent}). 
}\label{tab:sumup-conv} 
\end{table}

\section{Properties of the Bures-Wasserstein distance}
\label{app:BW}

\subsection{BW and the 2-Wasserstein distance}
\label{sec:def-W2} 

The Bures-Wasserstein distance has a natural connection with the 2-Wasserstein distance on a metric space. In the case of zero-centered Gaussian measures, the two distances are identical. 
We briefly describe the general definition of the 2-Wasserstein distance. 

Given a metric space $(\cX, \norm{\cdot})$, the 2-Wasserstein distance is a well-known metric on the space of quadratically integrable probability measures $\cP_2(\cX) \defeq \set{\mu \in \cP(\cX) \mid \int{\norm{x}^2 \dif \mu(x)} < \infty}$. 

\begin{definition}[2-Wasserstein distance]
   The 2-Wasserstein distance between two measures $(\nu, \nu_0) \in
    {(\cP_2(\cX))}^2$ is defined as the solution to following minimization problem: 
    \begin{equation}
        \label{eq:W2}
        \cW_2^2(\nu, \nu_0) = \inf_{\pi \in \Pi(\nu, \nu_0)} \int \norm{x-y}^2
        \dif \pi(x, y),
    \end{equation}
    where $\Pi(\nu, \nu_0)$ is the set of distributions with fixed marginals $\nu$ and $\nu_0$, 
    $\Pi(\nu, \nu_0) = \{ \pi \in
    \cP_2( \cX \times \cX) \mid \pi_1 = \nu,\, \pi_2 = \nu_0 \} $, with $\pi_i$ denoting the marginal of $\pi$ along the $i$th variable. 
\end{definition}

It is known that the 2-Wasserstein distance metrizes the weak convergence on the space $\cP_2$ \citep[see, e.g.][Theorem 6.9]{villani2008optimal}. Therefore, it can used to compare probability
distributions in systems such as GANs. 
On the other hand, the cost of computing this loss can quickly become prohibitive~\citep[see, e.g.][]{Pele2009Fast}. Only in some cases, efficient ways to compute~\eqref{eq:W2} are known. 
In a usual WGAN \citep{Arjovsky2017Wasserstein}, an approximation of the $1$-Wasserstein distance is computed based on the dual expression of the ($1$-)Wasserstein distance using a neural network to approximate the dual variable, called the discriminator network. 

The 2-Wasserstein distance between two Gaussian measures has a closed-form expression (or a closed-form expression for the discriminator), so that adversarial training is not needed. 
We will consider two centered Gaussian distributions, which are
described by their covariance matrices. In the case of centered Gaussian distributions, the 2-Wasserstein distance reduces to the Bures-Wasserstein distance between the covariance matrices $\Sigma_0$ and $\Sigma$~\citep{Dowson1982Frechet}: 

\begin{lemma}
    If $\nu = \cN(\bm{m}, \Sigma)$ and $\nu_0 = \cN(\bm{m}_0, \Sigma_0)$, then 
    \begin{equation*}
        \cW_2^2(\nu, \nu_0) = \norm{\bm{m} - \bm{m}_0}^2 + \cB^2(\Sigma,
        \Sigma_0) . 
    \end{equation*}
\end{lemma}

It is well known (see~\citealt{10.2307/2626967}
or~\citealt[Theorem~1.3]{villani2003topics} or~\citealt[Theorem
5.10]{villani2008optimal}) that the squared 2-Wasserstein distance has the
following dual expression, also known as the Kantorovich duality: 
\begin{equation}
    \label{eq:wass2-dual}
    \begin{split}
    \cW_2^2(\nu_0, \nu_\theta) = \sup_{(f, g) \in L^1(\nu_\theta) \times
        L^1(\nu_0)} & \Big\{\int
    {f(x) \dif \nu_\theta(x)} + \int {g(x) \dif \nu_0(x)}  
    \mid \forall (x, y),\, f(x) + g(y) \leqslant \norm{x-y}^2 \Big\}, 
    \end{split}
\end{equation}
where $L^1(\nu)$ is the set of the integrable functions with respect to a
measure $\nu$. Therefore, the dual variables $f$ and $g$ are required to be integrable with
respect to the source and target measures, and to fulfill the cost inequality.

\begin{remark}
    In the context of WGANs it is common to consider the 1-Wasserstein distance with cost given by the distance $\|x-y\|$. 
    This has a dual expression, referred to as the Kantorovich-Rubinstein formula
\citep[\S 6.2]{villani2008optimal}, that allows for a more tractable
computation in practice, with for instance only one dual variable. 
Nonetheless, in general there is no closed-form solution known when $c(x,y) = \norm{x-y}$. 
\end{remark}

    \subsection{BW and the Eckart-Young-Mirsky theorem}
In this section, we provide further background on the Bures-Wasserstein distance. 
First, we show that, except in some particular cases (\Cref{lem:hellinger}), the
Bures-Wasserstein distance between two covariance matrices is not translation
invariant (\Cref{lem:BW-not-transl-inv}), which implies that it cannot be
expressed as the norm (let alone unitary) of a difference between two matrices. Then, an explanation
as to why the critical points found in \Cref{thm:critical-points-L1} are the same
as the one found when using the squared Frobenius norm between $\Sigma$ and
$\Sigma_0$ is given.

\begin{lemma}
    \label{lem:hellinger}
    In the case that $\Sigma_0$ and $\Sigma$ commute, the Bures-Wasserstein distance reduces to the Hellinger distance: 
    \begin{equation*}
        \Sigma_0 \Sigma = \Sigma \Sigma_0 \quad \implies \quad \cB^2(\Sigma,
        \Sigma_0) = \norm{\Sigma ^ {1/2} - \Sigma_0^{1/2}}_F^2 . 
    \end{equation*}
\end{lemma}
\begin{proof}
    This follows from the fact that, if $\Sigma$ and $\Sigma_0$ commute, so do $\Sigma^{1/2}$ and $\Sigma_0^{1/2}$, so that $\Sigma_0^{1/2} \Sigma \Sigma_0^{1/2} = {(\Sigma_0^{1/2} \Sigma^{1/2})}^2$ and 
\begin{align*}\label{eq:commutative-case}
    \trc{((\Sigma^{1/2})^2 + (\Sigma_0^{1/2})^2 - 2{({\Sigma_0^{1/2}}{\Sigma^{1/2}})})} 
    &= \trc{({( {\Sigma^{1/2}} - {\Sigma_0^{1/2}})}( {\Sigma^{1/2}} - {\Sigma_0^{1/2}})^\top)} \\
    &= \norm{{\Sigma^{1/2}} - {\Sigma_0^{1/2}}}_F^2 , 
\end{align*}
as claimed. 
\end{proof}

From this, one remarks that the problem of minimizing the BW distance between covariance matrices that commute falls under the framework of the Eckart-Young-Mirsky theorem. 
In this case if the optimization variable is $\Sigma^{1/2} = {(WW^\top)}^{1/2}$, we obtain a formulation in terms of the squared error loss. 
Nonetheless, in the case where $\Sigma$ and $\Sigma_0$ do not commute, we do not have such a correspondence, as in general, the BW distance is not translation invariant, neither when considered as a function of $(\Sigma, \Sigma_0)$ nor when considered as a function of $(\Sigma^{1/2}, \Sigma_0^{1/2})$. 

\begin{lemma}[BW is not translation invariant]
    \label{lem:BW-not-transl-inv}
    There exist positive semidefinite matrices $(\Sigma, \Sigma_0) \in \cS_+(n)
    \times \cS_+(n)$ and  a translation $T \in \cS_+(n)$,  such that
    $\cB^2(\Sigma+T, \Sigma_0+T) \neq \cB^2(\Sigma, \Sigma_0)$.  The same
    statement also holds for the loss $\cE$ defined on the matrix square roots,
    $\cE(\Sigma^{1/2}, \Sigma_0^{1/2}) \defeq \cB^2(\Sigma, \Sigma_0)$. 
\end{lemma} 
\begin{proof}
    For the first part of the statement, taking 
    \begin{equation*}
    \Sigma = \begin{pmatrix} 1 & 0 \\ 0
        & 1 \end{pmatrix}, \quad 
        \Sigma_0 = \begin{pmatrix} 1 & 0 \\ 0 &2
        \end{pmatrix}, \quad 
        T = \begin{pmatrix} t & 0 \\ 0 & t \end{pmatrix},
        \quad t>0, 
        \end{equation*}
        then $\cB^2(\Sigma + T, \Sigma_0 + T) - \cB^2(\Sigma,
        \Sigma_0) = {(\sqrt{2+t} -
        \sqrt{1+t})}^2 - {(\sqrt{2} - 1)}^2 $, which is non-zero. 

        For the second part of the statement, if 
        $$
        \Sigma_0^{1/2} =
        \begin{pmatrix} 1 & 0 \\ 0 & 2 \end{pmatrix}, \quad \Sigma^{1/2} =
        \begin{pmatrix} 1 & 1 \\ 1 & 2 \end{pmatrix}, \quad 
        T = \begin{pmatrix}
    1 & 0 \\ 0 & 1 \end{pmatrix}, 
    $$
    one computes 
    \begin{align*}
        \cE(\Sigma^{1/2}, \Sigma_0^{1/2}) 
        =& \norm{\Sigma^{1/2}}_F^2 + \norm{\Sigma_0^{1/2}}_F^2 - 2 \trc{(\Sigma_0^{1/2} \Sigma
        \Sigma_0^{1/2})}^{1/2}
        \\
        =& 12 - 2\trc{ \begin{pmatrix} 2 & 6 \\ 6 & 20 \end{pmatrix}^{1/2}}
        \intertext{and}
        \cE(\Sigma^{1/2} + T, \Sigma_0^{1/2} + T) 
        =& \norm{\Sigma^{1/2} + T}_F^2 + \norm{\Sigma_0^{1/2} +T}_F^2 \\
        &- 2 \trc{((\Sigma_0^{1/2} +T)(\Sigma^{1/2} + T)(\Sigma^{1/2} +T)(
        \Sigma_0^{1/2} + T))}^{1/2}\\
        =& 28 - 2\trc{\begin{pmatrix} 20 & 30 \\ 30 & 90 \end{pmatrix}^{1/2}},
    \end{align*}
    which gives the difference $ \cE(\Sigma^{1/2} + T, \Sigma_0^{1/2} + T)  - \cE(\Sigma^{1/2}, \Sigma_0^{1/2}) \approx 0.121229 \neq 0$. 
\end{proof}

\Cref{lem:BW-not-transl-inv} implies that in general one cannot express the Bures-Wasserstein distance (either on the covariance or on their square roots) as a norm of a difference (otherwise, the loss would be translation invariant). 
This hinders a direct application of the Eckart-Young-Mirsky theorem, where the
problem is cast as $\min_{X} \norm{A - X}_*$ with a fixed $A$ for some unitary
invariant norm $\|\cdot\|_*$. 

Nonetheless, there is a close link between the Bures-Wasserstein distance and the (squared) Euclidean distance. This is best seen through the definition of the
2-Wasserstein distance between two zero-centered Gaussian distributions, as we will present next. 
We follow here an approach inspired by \citet[Theorem
1]{Feizi2020understanding}, for which we provide details in order to show a link
between the minimization of the Bures-Wasserstein distance over rank-constrained
covariance matrices and the Eckart-Young-Mirsky theorem (or $k$-PCA). 

Given $k \in [n]$, the set of rank-$k$ positive semi-definite matrices is denoted by $\cS_+(k; n)$. 
With $n \in \bN \setminus \set{0}$ and $k \in [n]$, 
we are interested in the minimization problem 
\begin{equation}
    \label{eq:EYM-Bures}
    \inf_{A \in \cS_+(k; n)} \cB^2(A, B).
\end{equation}
For any measure $\alpha$, denote $\supp(\alpha)$ its support, i.e.\ $\alpha(X)
= 0$ for $X \subseteq \bR^n \setminus \supp(\alpha)$. 
The following is a well known connection between covariance matrices and the support of the corresponding Gaussian probability distributions. 
\begin{lemma}
    \label{lem:spt-gauss}
    Let $A \in \cS_+(k; n)$ and $\alpha = \cN(0, A)$. 
    Then the support of $\alpha$ is equal to the column space of $A$, 
    $$
    \supp(\alpha) = \spn(A).
    $$
\end{lemma}
For $k \in [n]$, denote the set of linear subspaces of $\bR^n$ of
dimension $k$ by $\kL(\bR^n, k)$, and, for $\cC \in \kL(\bR^n, k)$, denote
by $\cN(\cC) \defeq \set{\cN(0, M) \mid M \in \cS_+(k; n),\, \spn(M) = \cC}$ the
set of all Gaussian distributions with mean $0$ and support $\cC$.

\Cref{lem:spt-gauss} allows to translate the problem~\eqref{eq:EYM-Bures} to a problem on linear subspaces of fixed dimension. Indeed, with $\alpha = \cN(0, A)$ and $\beta = \cN(0, B)$, we know that $\cB^2(A, B) = \cW_2^2(\alpha, \beta)$. Therefore, we can split the optimization problem as 
\begin{align}
    \label{eq:equiv-prob}
    \inf_{A \in \cS_+(k; n)} \cB^2(A, B) \iff \inf\, \set*{ \inf\, \set*{
    \cW_2^2(\alpha, \beta) \mid \alpha \in \cN(\cC)} \mid \cC \in \kL(\bR^n, k) }. 
\end{align}
Solving~\eqref{eq:EYM-Bures} is therefore equivalent to solving the right-hand side of~\eqref{eq:equiv-prob}, which is split in two parts: 
\begin{itemize}
\item 
For a given linear subset $\cC$ of dimension $k$, find the Gaussian distribution
that minimizes the $2$-Wasserstein distance to $\beta$. 
\Cref{lem:sol-alpha} below states that this $\alpha^*$ is the projection of $\beta$ onto $\cC$. 
\item 
Then, find the subset $\cC$ of required dimension that minimizes the variance of the projection of $\beta$ onto the orthogonal complement of $\cC$; or, equivalently, find $\cC$ that maximizes the variance of the projection of $\beta$ onto
$\cC$. The solution to this problem is the $k$-PCA decomposition of $\beta$, as stated in \Cref{lem:sol-C}.
\end{itemize}

Recall, given any $\beta \in \cP_2(\bR^n)$, that we are interested in solving
$\inf \set{\cW_2^2(\alpha, \beta) \mid \alpha \in \cN(\cC)}$. 
The next lemma
gives the solution this problem in $\alpha$. 
For any given linear subspace $\cC \subseteq \bR^n$, denote $p_\cC$
the orthogonal projection onto $\cC$. 

\begin{lemma}
    \label{lem:sol-alpha}
    Let $\beta \in \cP_2(\bR^n)$.
    One has $\inf\,\set{\cW_2^2(\alpha, \beta)
    \mid \alpha \in \cN(\cC)}  = \min\, \set{\cW_2^2(\alpha, \beta)
    \mid \alpha \in \cN(\cC)}$, and  the distribution
    $\alpha^*$ that achieves the minimum for a given $\cC$ is the orthogonal projection of $\beta$ onto
    $\cC$: $\alpha^* = {p_\cC}_{\#} \beta = \beta_\cC$.

\end{lemma}
\begin{proof}
Denote the admissible set of couplings with given marginals by $\Gamma(\alpha,
\beta) = \set{\pi \in \cP_2(\bR^n \times \bR^n) \mid \pi_1 = \alpha,\,
\pi_2 = \beta}$, with $\pi_i$ the marginal along the $i$th variable,
so that 
$$
\cW_2^2(\alpha, \beta) = \inf \set{\int \norm{x-y}^2 \dif
\pi(x,y) \mid \pi \in \Pi(\alpha, \beta)}.
$$
Then, for any given  linear subspace $\cC \subseteq \bR^n$, denote $p_\cC$
the orthogonal projection onto $\cC$.  Define $\mu_\cC \defeq
{p_\cC}_{\#}\mu $ for any $\mu \in \cP_2(\bR^n)$, and likewise $\pi_{\cC
\times \cC} \defeq {p_{\cC \times \cC}}_{\#} \pi$, for $\pi \in
\cP_2(\bR^n \times \bR^n)$, where, for $(\cX, \cY)$ two (measurable) spaces,
the push-forward $T_\# \mu \in \cP_2(\cY)$ of a measure $\mu \in \cP_2(\cX)$
by an operator $T \colon \cX \to \cY$ is such that, for any measurable set
$\cS \subseteq \cY$, $T_\#\mu(\cS) = \mu(T^{-1}(\cS))$. 

If $\supp(\alpha) = \cC$ (i.e.\ $\alpha
= \alpha_\cC$), one obtains
\begin{align}
    \cW_2^2(\alpha, \beta) 
        &= \inf_{\pi \in \Pi(\alpha, \beta)} \int \norm {x - y}^2 \dif \pi(x,y) \notag
        \\
        &= \inf_{\pi \in \Pi(\alpha, \beta)} \set*{\int \norm{x - y}^2
        \dif \pi_{\cC \times \cC}(x,y)} + 
        \int \norm{y}^2 \dif \beta_{\cC^\bot}(y) . 
        \label{eq:split-wass}
\end{align}
Thus, for a given $\cC \in \kL(\bR^n, k)$ one has 
    \begin{align*}
        \begin{split}
        &\inf \set*{\cW_2^2(\alpha, \beta) \mid \alpha \in \cN(\cC)} 
        \\
            = &\inf\set*{ \inf\set*{
                    \int\norm{x-y}^2 \dif \pi_{\cC \times \cC}(x,y) \mid \pi \in
            \Pi(\alpha, \beta)} \mid \alpha \in \cN(\cC)}
            + \int \norm{y}^2 \dif \beta_{\cC^\bot}(y).
                \end{split}
            \end{align*}
            We are interested in the term that is dependent on $\pi$ (and therefore
            $\alpha$), which is equivalent to
            \begin{equation*}
                \inf \set*{\inf \set*{ \int \norm{x - y}^2 \dif \pi(x,y) \mid \pi \in \Pi(\alpha, \beta_\cC)} \mid \alpha \in
                \cN(\cC)} = \inf \set*{ \cW^2_2(\alpha, \beta_\cC ) \mid \alpha \in
                \cN(\cC)}.
            \end{equation*}
            Since $\beta_\cC \in \cN(C)$, the solution is attained for $\alpha^*
            = \beta_\cC$.
        \end{proof}

        Then, the problem~\eqref{eq:EYM-Bures} is equivalent to 
        \begin{align*}
            \inf\, \set*{ \inf\, \set*{
            \cW_2^2(\alpha, \beta) \mid \alpha \in \cN(\cC) } \mid \cC \in \kL(\bR^n, k) }
            \iff &\inf \set*{\int
            \norm{y}^2\dif \beta_{\cC^\bot}(y) \mid \cC \in \kL(\bR^n, k)} 
            \\
                \iff &\sup \set*{\int \norm{y}^2 \dif \beta_\cC(y) \mid \cC \in \kL(\bR^n, k)}.
            \end{align*} 
            Therefore, the problem boils down to finding the linear subspace $\cC$ which
            maximizes the variance of the target when projected onto $\cC$.  
            The solution to this problem, also known as $k$-PCA, is given in the next
            lemma, of which we provide a proof for convenience. 
            \begin{lemma}
                \label{lem:sol-C}
                Let $\Omega \Lambda \Omega^\top = B$ be a spectral decomposition of $B
                \in \cS_{++}(n)$, with the eigenvalues in $\Lambda$ ranked in decreasing
                order. Let $k \in [n]$, and let $\Omega \eqdef \begin{pmatrix} \Omega_{[k]} &
                \Omega_{\bot} \end{pmatrix}$ be such that $\Omega_{[k]} \in \bR^{n
                \times k}$ corresponds to the $k$ highest eigenvalues of $B$.  Denote
                $\beta = \cN(0, B)$ and for a linear subspace $\cC$, denote by $\beta_\cC$
                the orthogonal projection of $\beta$ onto $\cC$. 
                Then 
                $$
                \sup\set{\int
                \norm{y}^2\dif\beta_\cC(y) \mid \cC \in \kL(\bR^n, k)} = \max\set{\int
            \norm{y}^2\dif\beta_\cC(y) \mid \cC \in \kL(\bR^n, k)} = \int \norm{y}^2
            \dif \beta_{\cC^*}(y), 
            $$
           where $\cC^* = \spn \Omega_{[k]}$. 
                \end{lemma}

                \begin{proof}
                    Recall that $\beta_\cC = {(p_\cC)}_{\#} \beta$, where $p_\cC$ is the
                    orthogonal projection onto any $\cC \in \kL(\bR^n, k)$. 
                    Then, 
                    \begin{align}
                        \int \norm{y}^2 \dif \beta_\cC(y) 
            &= \int \norm{p_\cC(y)}^2 \dif \beta(y)  \notag
            = \int y_\cC^\top y_\cC \dif \beta(y) \notag
            = \int \trc(y_\cC^\top y_\cC) \dif \beta(y) \notag 
            = \int \trc(y_\cC y_\cC^\top) \dif \beta(y) \notag
            \\
            &= \trc\Big(\int y_\cC y_\cC^\top \dif \beta(y)
            \Big),\label{eq:var-proj}
                    \end{align}
                    where the usual notation for $y_\cC = p_\cC(y)$ is used.

                    For $\cC \subseteq \bR^n$, there is equivalence between the two statements

                    \begin{enumerate}[label=(\roman*)]
                        \item $\cC \in \kL(\bR^n, k)$; and
                        \item $\exists C \in \bR^{n\times k}\colon C^\top C = I_k, \cC =
                            \spn C$.
                    \end{enumerate}
                    With such a $C$ spanning $\cC$, the projection onto $\cC$ can be written $p_\cC =
                    CC^\top$, and
                    \begin{align*}
                        \int y_\cC y_\cC^\top \dif \beta(y) 
                        = \int CC^\top y y^\top CC^\top \dif \beta(y)
                        = CC^\top \Big(\int y y^\top \dif \beta(y) \Big)CC^\top
                        = CC^\top B CC^\top,
                    \end{align*}
                    so that~\eqref{eq:var-proj} becomes
                    \begin{align*}
                        \trc\Big(\int y_\cC y_\cC^\top \dif \beta(y) \Big)
                        = \trc(CC^\top B CC^\top)  = \trc(C^\top B C)
                    \end{align*}

                    Therefore, with $\beta = \cN(0, B)$ and $\beta_\cC =
                    {(p_\cC)}_{\#}\beta$ for any $\cC \in \kL(\bR^n, k)$, the
                    following equivalence holds
                    \begin{equation}
                        \label{eq:sup-equiv}
                        \sup \set*{\int \norm{y}^2 \dif \beta_\cC(y) \mid \cC \in \kL(\bR^n,
                        k)} \iff \sup \set*{\trc(C^\top B C) \mid C \in \bR^{n\times
                        k},\, C^\top C = I_k}.
                    \end{equation}

                    Let $\Omega \Lambda \Omega^\top = \sum_{i=1}^n \lambda_i \omega_i
                    \omega_i^\top$ be a spectral decomposition of $B$ with
                    decreasing eigenvalues $\lambda_1 \geqslant \cdots \geqslant \lambda_n$. 

                    For any $C \in \bR^{n \times k}$ such that $C^\top C = I_k$, we compute 
                    \begin{align}
                        \label{eq:proj-sum}
                        \trc C^\top B C 
                        = \sum_{i=1}^n \lambda_i \trc (C ^\top \omega_i \omega_i^\top C)
                        = \sum_{i=1}^n \lambda_i \trc{( \omega_i^\top C C^\top \omega_i)}
                        = \sum_{i=1}^n \lambda_i \ang{\omega_i, C C^\top \omega_i}.
                    \end{align}

                    For each $i \in [n]$, by orthogonal decomposition $\omega_i = CC^\top \omega_i + (I_n -
                    CC^\top) \omega_i$, one has that $\ang{\omega_i, CC^\top \omega_i}
                    \leqslant 1$, with equality if and only if $CC^\top \omega_i =
                    \omega_i$, i.e., if and only if $\omega_i \in \spn C$.  The
                    sum~\eqref{eq:proj-sum} is therefore maximized for $C = \Omega_{[k]}$.

                    Therefore, the supremum in~\eqref{eq:sup-equiv} is attained for $C =
                    \Omega_{[k]} \iff \cC = \spn \Omega_{[k]}$, concluding the proof. 
                \end{proof}

Thus, the solution to \eqref{eq:equiv-prob} can be given as follows. 
                \begin{proposition}
                    Let $k \in [n]$, and let $\begin{pmatrix} \Omega_{[k]} & \Omega_{\bot}
                        \end{pmatrix}\begin{pmatrix} \Lambda_{[k]} & \\ &
                        \Lambda_{\bot} \end{pmatrix} \begin{pmatrix} \Omega_{[k]}^\top
                    \\ \Omega_{\bot}^\top \end{pmatrix} = B$  be a spectral
                    decomposition of $B  \in \cS_{++}(n)$, with the $k$ largest eigenvalues in
                    $\Lambda_{[k]}$.
                    Using the same notations as before, the problem~\eqref{eq:equiv-prob} is solved for $\cC = \spn(
                    \Omega_{[k]})$. In this case, $\alpha^* = \cN(0, B|_k)$, where
                    $B|_k = \Omega_{[k]} \Lambda_{[k]} \Omega_{[k]}^\top$. 
                    \end{proposition}

                    \begin{proof}
                        \Cref{lem:sol-C} already shows that the supremum is
                        obtained for $\cC = \spn \Omega_{[k]}$. In this case, the optimal
                        $\alpha^*$, the projection of $\beta$ onto $\cC = \spn \Omega_{[k]}$, has
                        covariance matrix
                        \begin{align*}
                            A^* 
            &= \int x x^\top \dif \alpha ^*(x) = \int x x ^\top \dif
            {(p_\cC)}_{\#} \beta(x)
            \\
            &= \int x_\cC x_\cC ^\top  \dif \beta(x)
            \\
            &= \Omega_{[k]} \Omega_{[k]}^\top \Sigma \Omega_{[k]}
            \Omega_{[k]}^\top 
            \\
            &= \Omega_{[k]} \Lambda_{[k]} \Omega_{[k]}^{\top}
            \\
            &= B|_{k} . 
                        \end{align*}
                    \end{proof}

\subsection{Gradient of the Bures-Wasserstein loss}
\label{app:gradient-BW}

We give here the gradient of the squared-Bures-Wasserstein distance between two
full-rank covariance matrices. 

\begin{notation}[Differential]
    \label{not:differential}
We denote the differential of $f$ at $X$
    in the direction $H$ by $\dif {} f(X)[H]$. 
    Sometimes, with $Y = f(X)$, the shorthand notation $\dif{} Y$ is preferred, %
    and it is assumed that the direction $H$ is a small perturbation $\dif X$ around $X$. 
    For instance, if $Y = f(X) = XX^\top$, then $\dif{} Y = \dif{}X X^\top + X \dif{}X^\top$ 
    is one way to write $\dif{} f(X)[H] = HX^\top + XH^\top$. 
\end{notation}

\begin{lemma}[Differential of $L$]
    \label{lem:fod-L1tilde}
The differential of $L$ on $\cS_{++}(n)$ is 
    \begin{equation*}
        \forall\, \Sigma \in \cS_{++}(n),\, X \in \cS_{++}(n),\quad \dif{}L(\Sigma)[X] = \trc{(X -
        \Sigma_0^{1/2}{[\Sigma_0^{1/2}\Sigma
    \Sigma_0^{1/2}]}^{-1/2}\Sigma_0^{1/2}X)}.
    \end{equation*}
\end{lemma}

\begin{proof}
    \label{prf:gradient-BW}
    We will use the fact that, for $A \in \cS_{++}(n)$, $\dif{} \trc(A^{1/2}) =
    \frac12 \trc(A^{-1/2} \dif{}A)$.
    By the differential calculus rules, for~$\Sigma \in \cS_{++}(n)$, 
    \begin{align*}
        \dif{} L(\Sigma) 
        &= \dif{} \trc (\Sigma + \Sigma_0 -
        2(\Sigma_0^{1/2}\Sigma \Sigma_0^{1/2})^{1/2})
        \\
        &= \trc \dif \Sigma - 2 \trc\dif{} [(\Sigma_0^{1/2}\Sigma
        \Sigma_0^{1/2})^{1/2}]
        \\
        &= \trc ( \dif{} \Sigma - (\Sigma_0^{1/2}\Sigma \Sigma_0^{1/2})^{-1/2}
        \dif{}( \Sigma_0^{1/2} \Sigma \Sigma_0^{1/2}) )
        \\
        &= \trc (\dif{}\Sigma - \Sigma_0^{1/2}(\Sigma_0^{1/2}\Sigma \Sigma_0^{1/2})^{-1/2}
        \Sigma_0^{1/2} \dif \Sigma).
    \end{align*}\end{proof}

\begin{corollary}[Gradient of $L$]
    \label{lem:gradient-BW}
The gradient of $L$ on $\cS_{++}(n)$ is 
    \begin{equation*}
        \label{eq:grad-L1tilde}
        \forall\: \Sigma \in \cS_{++}(n), \quad \nabla L(\Sigma) = I - \Sigma_0^{1/2}{[\Sigma_0^{1/2}\Sigma
    \Sigma_0^{1/2}]}^{-1/2}\Sigma_0^{1/2}.
    \end{equation*}
\end{corollary}

\subsection{Difference between BW and its smooth version} 
In this section, we provide the proof of \Cref{lem:diff-losses} stated in \Cref{sec:wass-lin-gan}.

\begin{proof}[Proof of \Cref{lem:diff-losses}]
    \label{prf:diff-losses}
Let $\Sigma = WW^\top$ and $\Sigma_\tau = WW^\top + \tau I_n$. 
In view of \eqref{def:BW}, the difference between the perturbative and the original loss is given by 
\begin{align}
    |L^1_\tau(W) - L^1(W)| = |L(\Sigma_{\tau}) - L^{1}(W)| &= \bigg|\tau n -
    2\trc{\left(\left(\Sigma_0^{1/2}\Sigma_{\tau}\Sigma_{0}^{1/2}\right)^{1/2}-\left(\Sigma_0^{1/2}WW^{\top}\Sigma_{0}^{1/2}\right)^{1/2}\right)}\bigg| \notag 
    \\
 &\leqslant \tau n +
 2\abs{\trc{\left(\left(\Sigma_0^{1/2}\Sigma_{\tau}\Sigma_{0}^{1/2}\right)^{1/2}
 - \left(\Sigma_0^{1/2}WW^{\top}\Sigma_{0}^{1/2}\right)^{1/2}\right)}}.  \label{eq:diff}
\end{align}
Let $A \defeq \Sigma_0^{1/2}\Sigma \Sigma_0^{1/2}$ and $A_\tau \defeq
\Sigma_0^{1/2}\Sigma_{\tau}\Sigma_0^{1/2}$. Note that $A_\tau = A + \tau
\Sigma_0$. We aim to bound
\begin{equation*}
    \abs{\trc{\left(A_\tau^{1/2} - A^{1/2}\right)}} =
    \abs{\trc{\left({(A + \tau \Sigma_0)}^{1/2} - A^{1/2}\right)}}.
\end{equation*}

We can bound the absolute value of the trace of a matrix by its spectral norm (defined as
$\norm{X}_* \defeq \sqrt{\lambda_{\max}(X X^\top)}$ for any matrix $X$)  as
$\abs{\trc{(X)}} \leqslant n \norm{X}_*$ for any matrix $X \in \bR^{n \times m}$. Then, a bound
on $\norm{X}_*$ can be found.

First assume that $\Sigma$ (hence $A$) is full rank, so that $A \succcurlyeq
\lambda_{\min}(A) I \succ 0$. We can then use the
perturbation inequality from~\citet[Lemma 2.2]{Sch1992Perturbation}
and find that 
\begin{align*}
    \norm{{(A + \tau \Sigma_0)}^{1/2} - A^{1/2}}_* 
    &\leqslant \frac{1}{\sqrt{\lambda_{\min}(A + \tau \Sigma_0)} +
\sqrt{\lambda_{\min}(A)}} \norm{\tau \Sigma_0^{1/2}}_*.
\intertext{Since}
    \lambda_{\min}(A + \tau \Sigma_0) 
    &\geqslant \lambda_{\min}(A) + \lambda_{\min}(\tau \Sigma_0) ,
    \\
\lambda_{\min}(A) 
    &\geqslant \lambda_{\min}(\Sigma) \lambda_{\min}(\Sigma_0),
    \\
    \text{and} \qquad \forall (a,b) \in {(\bR_{\geqslant 0})}^2,\quad  \sqrt{a} + \sqrt{b}
    &\geqslant \sqrt{a+b},
\intertext{one gets}
    \norm{{(A + \tau \Sigma_0)}^{1/2} - A^{1/2}}_* &\leqslant \frac{\tau
        \norm{\Sigma_0^{1/2}}_*}{\lambda_{\min}(\Sigma_0^{1/2}) \sqrt{
    \tau + 2\lambda_{\min}(\Sigma) }}\leqslant  \sqrt{\tau}
    \frac{\lambda_{\max}(\Sigma_0^{1/2})}{\lambda_{\min}(\Sigma_0^{1/2})},
\intertext{so that}
\abs{\trc\left({(A + \tau \Sigma_0)}^{1/2} - A^{1/2} \right)} &\leqslant n
    \sqrt{\tau} \frac{\lambda_{\max}(\Sigma_0^{1/2})}{\lambda_{\min}(\Sigma_0^{1/2})}.
    \intertext{Plugging it into~\eqref{eq:diff} yields the following bound}
\abs{L(\Sigma_\tau) - L(\Sigma)} &\leqslant n\sqrt{\tau} \left(\sqrt{\tau}+
    \frac{2\lambda_{\max}(\Sigma_0^{1/2})}{\lambda_{\min}(\Sigma_0^{1/2})}\right).
    \label{eq:bound-losses} \numberthis
\end{align*}

Now, in the case where $\Sigma$ has a rank deficiency, by continuity of the
function $X \mapsto \trc{(X^{1/2})}$, the bound found in~\eqref{eq:bound-losses}
still holds, since it does not depend on $\Sigma$. This completes the proof.
\end{proof}

\section{General results for linear networks}
\label{app:linear-networks}

This section deals with general properties of linear networks and their first- and second-order differential in parameter space. 
We first recall results that hold for any differentiable loss $\cL^1$ on $\bR^{n\times m}$ and its parametrization $\cL^N = \cL^1 \circ \mu$ on $\Theta$. 
These results have a long history in the linear neural networks literature
\citep{Baldi1989Neural,Kawaguchi2016Deep,Arora2018Optimization,Arora2019Convergence,Chitour2022Geometric,Bah2021Learning};
we report them here borrowing the presentation from~\citealp{Bah2021Learning}.
By convention, the product of matrices $W_{q:p} \defeq  W_{q} W_{q-1}
\cdots W_p$ is equal to $I_{d_q}$ when $q < p$.

\begin{lemma}[Gradient flow, {\citealt[Lemma 2.1]{Bah2021Learning}}]\label{lem:flow-prop}
    For any differentiable loss $\cL^1$, and parametrization $\cL^N = \cL^1 \circ \mu$, such that $\mu(W_1, \ldots, W_N) = W_N \cdots W_1$, one has: 
    \begin{enumerate}[leftmargin=*]
        \item \label{lem:chain-rule}
            For all $j \in [N]$,
    \begin{equation}\label{eq:gradient-LN}
        \nabla_{W_j} \cL^N(W_1, \ldots, W_N) = W_{N:j+1}^\top \nabla \cL^1(W)
        W_{j-1:1}^\top. 
    \end{equation}
        \item \label{lem:flow-w}
            If each of the $W_i(t)$ satisfies the
            flow~\eqref{eq:GF}, then the
            product $W_{N:1} = W_N \cdots W_1$ satisfies 
    \begin{equation}
        \label{eq:flow-w}
        \od{W(t)}{t} = - \sum_{j=1}^N W_{N:j+1} W_{N:j+1}^\top \nabla \cL^1(W)
        W_{j-1:1}^\top W_{j-1:1} 
    \end{equation}
        \item \label{lem:invariance}
    For all $j \in [N-1]$ and all $ t \geqslant 0$, 
    \begin{equation}
        \od{}{t} (W_{j+1}^\top(t) W_{j+1}(t)) = \od{}{t}
        (W_j(t)W_j^\top(t)) . 
    \end{equation}
\item
    \label{lem:balanced-flow}
    If $W_1(0), \ldots, W_N(0)$ are balanced, then, for all $j\in [N-1]$ and
    all $t \geqslant 0$, $W_{j+1}^\top(t) W_{j+1}(t) = W_j(t)W_j^\top(t)$ and 
    \begin{equation}
        \label{eq:balanced-flow}
        R(t) \defeq \od{W(t)}{t} + \sum_{j=1}^N
        {(W(t)W^\top(t))}^{\frac{N-j}{N}} \nabla \cL^1(W) {(W^\top(t)
        W(t))}^{\frac{j-1}{N}} = 0.
    \end{equation}
    \end{enumerate}
\end{lemma}

In the case of a twice-differentiable loss $\cL^1$ and the parametrization $\cL^N = \cL^1 \circ \mu$, one can express the second-order differential as follows. 
\begin{lemma}[Second-order differential]
    \label{lem:sod-loss} 
    Let $(\overrightarrow{U}, \overrightarrow{V}) \in \Theta\times \Theta$ be two parameters, $\overrightarrow{U} = (U_1, \ldots, U_N),\,
    \overrightarrow{V}=(V_1, \ldots, V_N)$. The second-order differential of the loss $\cL^N$ at $\overrightarrow{W} = (W_1,
    \ldots, W_N) \in \Theta$ is 
    \begin{equation}
        \label{eq:sod-loss}
        \begin{aligned}
            &\dif{}^2 \cL^N(\overrightarrow{W})[\overrightarrow{U}, \overrightarrow{V}] 
            =  \sum_{i=1}^N \sum_{j \neq i}
                                       \ang{U_i, W_{i+1}^\top
            \cdots V_j^\top \cdots W_N^\top \nabla \cL^1(W) W_1^\top \cdots
        W_{i-1}^\top} \\
        & \quad + \sum_{i=1}^N \sum_{j=1}^N {\vc(U_i)}^\top \Big( W_{i-1:1} \otimes {(W_{N:i+1})}^\top \cdot \nabla^2 \cL^1(W) \cdot
            {(W_{j-1:1})}^\top \otimes (W_{N:j+1}) \Big)\vc(V_j),
    \end{aligned}
    \end{equation}
where $\ang{A, B} = \trc A B^\top$ for two matrices of compatible sizes and $\nabla^2 \cL^1(W) \in \bR^{n^2\times n^2}$ is the matrix such that,
    $\forall (U,V) \in
    {(\bR^{n \times n})}^2 ,\, \dif{}^2 \cL^1(W)[U, V] = {\vc(U)}^\top  \nabla^2
    \cL^1(W) \vc(V)$. 
\end{lemma}
\begin{proof}
    The second-order differential for the parametrization $\phi$ is, for two parameters $(\overrightarrow{U},\overrightarrow{V}) \in \Theta\times \Theta$, 
    \begin{align*}
        \dif{}^2 \phi(\overrightarrow{W})[\overrightarrow{U}, \overrightarrow{V}] 
        &= \dif{}(\overrightarrow{W} \mapsto \dif{} \phi(\overrightarrow{W})[\overrightarrow{U}])[\overrightarrow{V}] = \{\dif{}\phi(\overrightarrow{W}
        + \overrightarrow{V})[\overrightarrow{U}] - \dif{}\phi(\overrightarrow{W})[\overrightarrow{U}] \}\big|_{\text{lin}}
        \\ 
        &= \Big\{ \sum_{i=1}^N (W_N + V_N) \cdots   U_i \cdots (W_1
        + V_1) - \sum_{i=1}^N W_N \cdots U_i \cdots
    W_1 \Big\} \Big|_{\text{lin}}
        \\
        &= \sum_{i=1}^N \sum_{j \neq i}^N W_N \cdots V_j \cdots U_i \cdots W_1 . 
    \end{align*}
    Here, $f(\overrightarrow{U}, \overrightarrow{V})|_{\text{lin}}$ refers to the linear part of $f$ with
    respect to each $U_i, V_j$. 
    From the chain rule for second-order differentials, 
    \begin{align*}
        \dif{}^2& \cL^N(\overrightarrow{W})[\overrightarrow{U}, \overrightarrow{V}] \nonumber \\
        &= \dif{}^2 (\cL^1 \circ \phi) (\overrightarrow{W})[\overrightarrow{U}, \overrightarrow{V}] 
        \\
        &=  \dif{}^2 \cL^1(W)\Big[\dif{}\phi(\overrightarrow{W})[\overrightarrow{U}],
            \dif{}\phi(\overrightarrow{W})[\overrightarrow{V}]\Big] +
        \dif{}\cL^1(W)\Big[\dif{}^2 \phi(\overrightarrow{W})[\overrightarrow{U}, \overrightarrow{V}]\Big]
        \\
        &=  \sum_{i=1}^N \sum_{j=1}^N \dif{}^2 \cL^1(W)[W_N \cdots U_i \cdots
        W_1,\, W_N \cdots V_j \cdots W_1] +
        \sum_{i=1}^N \sum_{j\neq i} \dif{}\cL^1(W)[W_N \cdots V_j \cdots U_i \cdots W_1]
        \\
        &=  \sum_{i=1}^N \sum_{j=1}^N {\vc(W_N \cdots U_i \cdots W_1)}^\top
        \nabla^2 \cL^1(W) \vc(W_N \cdots V_j \cdots W_1) 
        + \sum_{i=1}^N \sum_{j\neq i} \ang{\nabla \cL^1(W), W_N \cdots V_j \cdots
        U_i \cdots W_1}
        \\
        &=  \sum_{i=1}^N \sum_{j=1}^N {\Big({(W_{j-1} \cdots W_1)}^\top \otimes (W_N
                \cdots W_{i+1}) \vc(U_i)\Big) }^\top \nabla^2 \cL^1(W) \Big( {(W_{j-1}
        \cdots W_1)}^\top \otimes (W_N \cdots W_{i+1})\Big)   \vc(V_j) \nonumber 
     \\ & \qquad + \sum_{i=1}^N \sum_{j\neq i} \ang{W_{i+1}^\top \cdots V_j^\top \cdots W_N^\top \nabla \cL^1(W)
     W_1^\top  \cdots W_{i-1}^\top, U_i}
        \\
        &=  \sum_{i=1}^N \sum_{j=1}^N {\vc(U_i)}^\top
        {\Big({(W_{j-1} \cdots W_1)} \otimes {(W_N
                \cdots W_{i+1})}^\top \Big) } \nabla^2 \cL^1(W) \Big( {(W_{j-1}
        \cdots W_1)}^\top \otimes (W_N \cdots W_{i+1})\Big) \vc(V_j) 
     \nonumber \\ & \qquad + \sum_{i=1}^N \sum_{j\neq i} \ang{W_{i+1}^\top \cdots V_j^\top \cdots W_N^\top \nabla \cL^1(W)
     W_1^\top  \cdots W_{i-1}^\top, U_i} . 
    \end{align*}
\end{proof}

\begin{corollary}[Hessian of the Loss]
    \label{cor:hessian-loss}
    The Hessian of $\cL^N$, $\nabla^2 \cL^N(\theta)$, can be represented as a $d_\theta \times d_\theta$
    matrix. 
    It is a block matrix with blocks corresponding to different layers. 
    Each block $\nabla^2_{W_i,W_j} \cL^N (\overrightarrow{W})$  has dimension
    ${d_i}d_{i-1}  \times d_{j}d_{j-1}$, and corresponds to the differential $\dif{}^2
    \cL^N(\overrightarrow{W})[\overrightarrow{U}_i, \overrightarrow{U}_j]$, where
    $\overrightarrow{U}_i = (0, \ldots, 0, U_i, 0, \ldots, 0)$. 
    The diagonal block elements are 
    \begin{align}
        \label{eq:hess-diag}
        \nabla^2_{W_i} \cL^N(\overrightarrow{W}) = 
        ( W_{i-1:1}
                \otimes {(W_{N:i+1})}^\top) \cdot \nabla^2 \cL^1(W) \cdot
            {(W_{i-1:1})}^\top \otimes (W_{N:i+1}) ,
        \end{align}
        and the off-diagonal blocks are 
    \begin{equation}
        \label{eq:hess-offdiag}
        \begin{split}
        \nabla^2_{W_i,W_j} \cL^N(\overrightarrow{W})  &= ( W_{i-1:1}
                \otimes {(W_{N:i+1})}^\top) \cdot \nabla^2 \cL^1(W) \cdot
            ({(W_{j-1:1})}^\top \otimes W_{N:j+1}) \\
                                                    & \qquad + \Big[ (W_{i-1} \cdots W_1
        {\nabla \cL^1(W)}^\top W_N \cdots  W_{j+1}) \otimes (W_{i+1}^\top \ldots
        W_{j-1}^\top ) \Big] K_{d_j d_{j-1}}, 
    \end{split}
    \end{equation}
    where $K_{pq}$ is the $pq$-commutation matrix (for $X \in \bR^{p \times q},\, K_{pq}
    \vc{X} = \vc{X^\top}$).
\end{corollary}

\begin{proof}
    The evaluation of the second-order differential
    $\dif{}^2L^N(\overrightarrow{W})[\overrightarrow{U}, \overrightarrow{V}]$
    given in~\eqref{eq:sod-loss} at $[\overrightarrow{U}_i,
    \overrightarrow{U}_i]$ readily provides the diagonal blocks of the Hessian. 
    For the off-diagonal blocks, the expression 
    \begin{align*}
        &\ang{U_i, W_{i+1}^\top \cdots U_j^\top \cdots W_N^\top \nabla \cL^1(W)
        W_1^\top \cdots W_{i-1}} 
    \end{align*}
        {can be transformed into}
        \begin{align*}
                                 &{\vc{(U_i)}} ^\top \vc{(W_{i+1}^\top \cdots U_j^\top \cdots W_N^\top \nabla \cL^1(W)
    W_1^\top \cdots W_{i-1})}  
        \\
        &= \vc{(U_i)} ^\top \Big[ (W_{i-1} \cdots W_1 \nabla \cL^1(W) W_N \cdots
        W_{j+1}) \otimes (W_{i+1}^\top \cdots W_{j-1}^\top) \Big] \vc( U_j^\top)
        \\
        &= \vc{(U_i)} ^\top \Big[ (W_{i-1} \cdots W_1 \nabla \cL^1(W) W_N \cdots
            W_{j+1}) \otimes (W_{i+1}^\top \cdots W_{j-1}^\top) \Big] K_{d_j
        d_{j-1}}\vc( U_j),
    \end{align*}
    proving~\eqref{eq:hess-offdiag}. 
\end{proof}

Now, for the smooth BW loss, we would like to show convergence to a critical point of $L^N_\tau$ under the gradient flow update of the parameters. 
We first show that the BW loss $L^1$ restricted to the matrices $W$
of full row-rank $\cM_*$ satisfies the
so-called \L{}ojasiewicz inequality (meaning there exist constants $c>0$, $\mu >0$ such that, for all $W \in \cM_*$ in a neighbourhood of a critical
point $W^* \in \cM_*$, $\norm{\nabla L^1 (W)} > c \norm{L^1(W) - L^1(W^*)}^\mu$).

\begin{lemma}\label{equivalency}
    For any\ $W\in \cM_*$ (such that $WW^\top \in
    \cS_{++}(n)$), and for the loss $L^1$ defined
    in~\eqref{eq:L1}, we have
    \begin{equation*}\label{DW_equality}
        \norm{\nabla_W L^1(W)}_F^2 = 4 L^1(W).
    \end{equation*}
\end{lemma}
\begin{proof}
This equality can be obtained by direct computation. Since 
\begin{equation*}
    \label{eq:gradient-L1-n}
    \nabla L^1(W) = 2W - 2\Sigma_{0}^{1/2}(\Sigma_0^{1/2} WW^\top\Sigma_0^{1/2} )^{-1/2}\Sigma_0^{1/2} W,
\end{equation*}
we have
 \begin{equation*}
     \begin{aligned}
         \norm{&\nabla_W L^1(W)}_F^2\\
         &= 4\trc \Big(\big(W - \Sigma_{0}^{1/2}(\Sigma_0^{1/2} W W^{\top}\Sigma_0^{1/2})^{-1/2}\Sigma_{0}^{1/2} W\big)\big(W^{\top} - W^{\top}\Sigma_{0}^{1/2}(\Sigma_0^{1/2} W W^{\top}\Sigma_0^{1/2})^{-1/2}\Sigma_{0}^{1/2} \big) \Big)\\
         &=4\trc(WW^{\top}) - 4\trc \Big(WW^{\top}\Sigma_{0}^{1/2}(\Sigma_0^{1/2} W W^{\top}\Sigma_0^{1/2})^{-1/2}\Sigma_{0}^{1/2}\Big)\\
         &~~~ -4\trc\Big(\Sigma_{0}^{1/2}(\Sigma_0^{1/2} W W^{\top}\Sigma_0^{1/2})^{-1/2}\Sigma_{0}^{1/2}WW^{\top}\Big) + 4\trc(\Sigma_0).
     \end{aligned}
 \end{equation*}   
 Note that the mid two terms above are the same, and they can be simplified as 
 \begin{equation*}
 \begin{aligned}
    \trc \Big(WW^{\top}\Sigma_{0}^{1/2}(\Sigma_0^{1/2} W W^{\top}\Sigma_0^{1/2})^{-1/2}\Sigma_{0}^{1/2}\Big) &=  \trc\Big(\Sigma_{0}^{1/2}(\Sigma_0^{1/2} W W^{\top}\Sigma_0^{1/2})^{-1/2}\Sigma_{0}^{1/2}WW^{\top}\Big)\\
    &= \trc\Big((\Sigma_0^{1/2} W W^{\top}\Sigma_0^{1/2})^{1/2}\Big).
    \end{aligned}
 \end{equation*}
 Combining all the terms together, we get the equality \eqref{DW_equality}.  
\end{proof}

The conservation quantity described in \Cref{lem:flow-prop} item \ref{lem:invariance} for the gradient flow~\eqref{eq:GF} is key in numerous analyses. 
Another useful property is the following, which %
ensures that the gradient flow~\eqref{eq:GF} converges to a critical point of $\cL^N$. 
Namely, if the trajectory $t \mapsto \W(t)$ remains bounded for all $t \geqslant 0$, and if $\cL^1$ is an analytic function (i.e.\ locally given by a power series), 
then~\eqref{eq:GF} converges to a critical point of $\cL^N$, i.e.\ a point
$\theta^*$ so that $\nabla \cL^N(\theta^*) = 0$. 
This is stated in the next theorem. 

\begin{theorem}[Gradient flow converges to a critical point of $\cL^N$]\label{thm:flow-convergence}
    Let $\cL^1$ be analytic and suppose the trajectory $t \mapsto \mu(\theta(t))$ remains bounded under the gradient flow evolution $\dot{\theta} = -\nabla {[\cL^1 \circ \mu]}(\theta)$. 
    Then, the flows of $W_i(t)$ and $W(t)$ given by~\eqref{eq:GF} and~\eqref{eq:flow-w} are defined and bounded for all $t \geqslant 0$ and ${(W_1, \ldots, W_N)}$ converges to a critical point of $\cL^N = \cL^1 \circ \mu$ as $t \to \infty$. 
\end{theorem}

\begin{proof}
    This result is proven by~\citet[Thereoem 3.2]{Bah2021Learning} for the
    squared error loss, but it can be stated for an arbitrary analytic loss. 
    It relies on the \L{}ojasiewicz argument for the convergence of gradient flows~\citep[Theorem 2.2]{Absil2005Convergence}, and the fact that each of the weights $W_i$ is bounded in norm as long as the end-to-end product $\mu(\theta) = W_{N:1}$ is. 
    This last claim is proven by \citet[Theorem 3.2]{Bah2021Learning} and does not depend on the particular loss, as long as it is differentiable (so that the gradient flow is well defined). 
\end{proof}

The boundedness of $\norm{W}$ can be shown depending on the loss that is
considered. For example, it holds for the regularized loss $L^1_\tau$ as we discuss next. 
For the loss $L^1_\tau$ introduced in~\eqref{pertloss}, one can indeed bound the norm of $W$ throughout training as stated in \Cref{lem:bound-L1tau} below. 
Since the loss $L^1_\tau$ is analytic, one immediately gets the following result. 
We give a simple test to show the boundedness of a trajectory under~\eqref{eq:GF}, using the decrease of the loss along training. 

\begin{lemma}
    \label{lem:lemma1} Let $\cL^1 \colon \bR^{n \times m} \to \bR$ be a given
    loss, let $\mu \colon \Theta \to \bR^{n\times m}$ be the linear network
    parametrization, and denote $W(t) = \mu(\theta(t))$ for $\theta \colon \bR
    \to \Theta$ a path on the parameter space. Assume that there exists an increasing function $f\colon \mathbb{R}\to\mathbb{R}$ such that, for any $t \geqslant 0$, one has $\norm{W(t)}
    \leqslant f(\cL^1(W(t)))$. Then, the trajectory $t \mapsto W(t)$
    under the gradient flow \eqref{eq:GF} is bounded. 
\end{lemma}

\begin{proof}
    Under gradient flow, for any $t \geqslant 0,\, \cL^1(W(t)) \leqslant
    \cL^1(W(0))$. Indeed, writing the chain rule and the gradient flow~\eqref{eq:flow-w},
    \begin{align*}
        \od{}{t} \cL^1(W(t)) &= \sum_{j} D_{W_j}\cL^N{(W_1(t), \ldots, W_N(t))}
        \od{W_j(t)}{t} \\
                           &= -\sum_j \norm{\nabla_{W_j} \cL^N{(W_1, \ldots,
                           W_N)}}_F^2 \leqslant 0.
    \end{align*} 
    Therefore, for any $t \geqslant 0$, $\cL^1(W(t)) \leqslant \cL^1(W(0))$. 
    Now, let $f \colon \bR \to \bR$ be an increasing function, so that $f(\cL^1(W(t)))
    \leqslant f(\cL^1(W(0)))$. 
    Therefore, if for any $ t \geqslant 0,\, \norm{W(t)}
    \leqslant f(\cL^1(W(t)))$, then  $\norm{W(t)} \leqslant f(\cL^1(W(t))) \leqslant
    f(\cL^1(W(0)))$ is bounded. 
\end{proof}

The assumption in~\Cref{lem:lemma1} is satisfied for a couple of losses, including the squared error loss \citep{Bah2021Learning} and the $L^1_\tau$ loss, 
as shown in~\Cref{lem:bound-L1tau} below. This allows us to consider losses that ``grow with the weights'', so that the end-to-end matrix is bounded when the loss converges to zero. 
We now show the boundedness of the weights when considering the Bures-Wasserstein loss~\eqref{eq:L1}. 

\begin{lemma}[Boundedness for the BW loss $L$]
    \label{lem:bound-BW-L} 
    Given a target $\Sigma_0$, the loss $L(\Sigma)$ is lower-bounded by~${\frac12 \trc\Sigma - \trc \Sigma_0}$. 
\end{lemma}
\begin{proof}
 By the dual expression of the Wasserstein
 distance~\eqref{eq:wass2-dual}, 
 $$
 L(\Sigma) = \cW_2^2(\nu_0, \nu_\theta) = \sup_{f\in \cL^1(\nu_\theta)} \int f(x) \dif\nu_\theta + \int f^{\norm{\cdot}^2}(y) \dif\nu_0(y),
 $$ 
 with $\nu_\theta = \cN(0, \Sigma),\, \nu_0  = \cN(0, \Sigma_0)$ and $f^{\norm{\cdot}^2}$ 
 the
 $\norm{\cdot}^2$-transform of $f$ defined as $\forall y \in \bR^d,\,
 f^{\norm{\cdot}^2}(y) = \inf_{x\in \bR^d} \norm{x-y}^2 - f(x)$.

    With $\tilde{f} \colon x \mapsto \frac12 \norm{x}^2 $, the
    $\norm{\cdot}^2$-transform of $\tilde{f}$ is 
    $\tilde{f}^{\norm{\cdot}^2} \colon y \mapsto -\norm{y}^2 $, and we get 
    \begin{alignat}{2}
        &&  L(\Sigma) = \cW_2^2(\nu_0, \nu_\theta) 
        &\geqslant \int \frac12\norm{x}^2 \dif
        \nu_\theta(x) - \int \norm{y}^2 \dif\nu_0(y)  = \frac12 \trc{\Sigma} -
        \trc{\Sigma_0}, \label{eq:W2-bound}
    \end{alignat}
as claimed. 
\end{proof}

\begin{lemma}[Boundedness for the loss $L^1_\tau$] 
    \label{lem:bound-L1tau}
    The norm of the end-to-end matrix $W$ is upper-bounded when using the loss $L^1_{\tau}$ defined in~\eqref{pertloss}. 
\end{lemma}
\begin{proof}
    With $\varphi_\tau(\Sigma) = \Sigma + \tau I_n \eqdef \Sigma_\tau$,  the loss $L^1_\tau$ satisfies 
    \begin{alignat*}{1}
        L^1_\tau(W) 
        = L(\varphi_\tau(\pi(W)))
         & \geqslant
    \frac12 \trc{\Sigma_\tau}  - \trc{\Sigma_0}  =  \frac12 \trc{WW^\top} - \trc{\Sigma_0} + \frac{n}{2}  \tau
        \\
        \implies \quad
         \sqrt{2L^1_\tau(W) + 2 \trc \Sigma_0 - n \tau} 
        &\geqslant \norm{W} . 
    \end{alignat*}
    Therefore, there exists an increasing function $f$ such that $\norm{W} \leqslant f(L^1_\tau(W))$. Since the loss decreases under gradient flow, one has 
    \begin{equation}
        \label{eq:bound-W}
        \norm{W(t)} \leqslant   \sqrt{2L^1_\tau(W(0)) + 2 \trc \Sigma_0 - n
        \tau},
   \end{equation}
 and the boundedness of $t \mapsto W(t)$ is shown. 
\end{proof}

\begin{corollary}\label{eq:W_lower}
    For the Bures-Wasserstein loss $L$, if $W W^\top$ is positive definite, so
    that the loss is differentiable, then, the norm of the end-to-end matrix
    $W(t) = \mu(\theta(t))$ is uniformly bounded throughout the flow: 
    \begin{equation}
    \forall t \geqslant 0,\, \norm{W(t)}\leqslant    \sqrt{2L^1(W(0)) + 2\trc{\Sigma_0}},
    \end{equation}
    by using similar arguments as in the proof of \Cref{lem:bound-L1tau}.
\end{corollary}

\Cref{eq:W_lower} will be useful in the proof of \Cref{thm:gradientdescent}. 

\begin{lemma}
    \label{lem:GF-LNtau}
    The gradient flow~\eqref{eq:GF} on the perturbative loss~\eqref{pertloss}
    converges to a critical point $\theta^*$ of~$L^N_\tau$. 
\end{lemma}

This property of the gradient flow is necessary in order to prove the
convergence of the training to a minimizer of $L^1_\tau$. 
At first glance, there is no immediate reason to expect that the critical points of $L^N_\tau$ correspond to critical points of $L^1_\tau$, since the parametrization $\mu$ could introduce critical points. 
This last aspect led~\citet{Trager2020Pure} to distinguish between the \emph{pure} and \emph{spurious} critical points of a linear network; i.e.\ points that are critical for both $L^N$ and $L^1$, and those that are critical only for $L^N$, and study conditions under which spurious local minima can be excluded. 

\section{Proofs of \Cref{sec:critical-points}}
\label{app:proofs-critical-points}

In this section, we provide the proofs of the statements about the critical points of the loss functions in function space, $L^1|_{\cM(k)}$ and $L^1_\tau|_{\cM(k)}$. 
We characterize the critical points and show that all saddles are strict.

\subsection{Critical points of \texorpdfstring{$L^1|_{\cM(k)}$}{L1|Mk}}

First, the loss $L^1$ is expressed on the manifolds $\cM(k)$
(\Cref{lem:reformulation-loss}), where it is differentiable
(\Cref{lem:gradient-L1-svd}). 
Then, necessary conditions (\Cref{lem:necessary-fooc-svd}) on the critical points can be expressed, leading to the first part of \Cref{thm:critical-points-L1}. 
The second part of 
\Cref{thm:critical-points-L1} is then proven by evaluating the loss at the critical points found, and ranking them. 

Recall \Cref{def:crit-restricted} of the critical points of a function restricted to a manifold. Computing the differential of the restriction $L^1|_{\cM(k)}$ will allow to characterize the different critical points. 

\begin{definition}[Gradient]
    Given an embedded manifold $\cM$ and a function $f$ with a differentiable restriction $f|_{\cM}$, the gradient of 
    $f|_{\cM}$ 
    at $x \in \cM$ is the (unique) element of the
    tangent space $T_x \cM$ such that, for all $v \in T_x \cM$, $\dif{}
    f|_{\cM}(x)[v] = \ang{\nabla f|_{\cM}(x), v}$.
\end{definition}

We begin by expressing the loss $L^1|_{\cM(k)}$ with the Singular Value Decomposition (SVD) of $\Sigma_0^{1/2}W$.

\begin{lemma}\label{lem:reformulation-loss}

    Let $USV^\top = \Sigma_0^{1/2} W$ be a thin SVD of $\Sigma_0^{1/2}W$, so
    that $U\in \bR^{n \times k},\, V\in \bR^{m \times k},\, U^\top U = V^\top V
    = I_k,\, S = \Diag(s_1, \ldots, s_k) \in \bR^{k \times k}$, where $k =
    \rank \Sigma_0^{1/2} W  = \rank W$.
    The loss $L^1$ from~\eqref{eq:L1} on $\cM(k)$ can be expressed as
    \begin{equation}\label{eq:loss-svd}
        L^1|_{\cM(k)}(W) = \norm{W}^2_F + \norm{\Sigma_0^{1/2}}_F^2 - 2 \trc{S} .
    \end{equation} 
\end{lemma}

\begin{proof} 
If $U S V^\top = \Sigma_0^{1/2}W$ is a thin SVD of $\Sigma_0^{1/2}W$, then ${(\Sigma_0^{1/2}W{(\Sigma_0^{1/2}W)}^\top)}^{1/2} = U S U^\top$. 
Therefore, the expression of the loss $L^1$ given by~\eqref{eq:L1} can be written as 
    \begin{equation*}
        L^1|_{\cM(k)}(W) = \trc WW^\top + \trc \Sigma_0 - 2\trc{USU^\top} =
    \norm{W}^2_F + \norm{\Sigma_0^{1/2}}_F^2 - 2 \trc{S} , 
    \end{equation*} 
    as claimed. 
\end{proof}

With this description at hand, we now give the gradient of $L^1|_{\cM(k)}$. 

\begin{lemma}[Gradient of $L^1|_{\cM(k)}$]\label{lem:gradient-L1-svd} 
Let $(n, m) \in {(\bN \setminus \set{0})}^2$,  and let $k \leqslant \min{\set{n, m}}$. The
    loss $L^1|_{\cM(k)}$ (as given by~\eqref{eq:loss-svd}) is twice
    continuously differentiable on $\cM(k)$.  With $W \in \cM(k)$ and $USV^\top =
    \Sigma_0^{1/2}W$ a thin SVD of $\Sigma_0^{1/2}W$, its gradient is 
    \begin{equation}\label{eq:gradient-L1-svd} 
        \nabla L^1|_{\cM(k)}(W) = 2W - 2\Sigma_0^{1/2} UV^\top . 
    \end{equation} 
\end{lemma}

In order to prove \Cref{lem:gradient-L1-svd}, we need the differential
expression for the singular values appearing in the SVD of a matrix. Recall the
notation given in \cref{not:differential} for the differential.

\begin{lemma}[Differential of the SVD]\label{lem:diff-svd}

    Let $k \leqslant \min\set{n, m}$ and let $X \in \cM(k)$ be a matrix
    with $\rank{X} = k$.  Let $USV^\top = X$ be a thin SVD of $X$, with $U \in
    \bR^{n \times k},\, S \in \bR^{k \times k},\, V \in \bR^{m \times k}$, $S$
    diagonal and $U^\top U = V^\top V = I_k$. Then, the differential $\dif S$ is 
    \begin{equation*}
        \dif S = I_k \odot (U^\top\dif X V),
    \end{equation*}
    where $A \odot B$ denotes the Hadamard product between $A$ and $B$. 
\end{lemma}

\begin{proof} 
    Let $USV^\top = X$ be the decomposition as given in the lemma statement. The differential rules ensure that 
    \begin{alignat*}{2}
        && \dif X &= \dif{U} S V^\top + U \dif{S} V^\top + U S \dif{V^\top}.
        \\
        \intertext{This implies that }
        && U^\top \dif X V 
        &= U^\top \dif U S V^\top V + U^\top U \dif
        S V^\top V + U^\top U S \dif {V^\top} V
        \\
        && &= U^\top \dif U S  +  \dif S  + S \dif {V^\top} V
        \\
        & \implies & \dif S &= U^\top \dif X V - U^\top \dif U S - S \dif V^\top
        V.
    \end{alignat*} 
    Since $U^\top U = I_k$, $\dif{U^\top} U + U^\top \dif U = 0$, and $A \defeq
    U^\top \dif U = - \dif U^\top U = - A^\top$. Likewise, $B \defeq V^\top \dif V$
    is also antisymmetric.  The matrices $A$ and $B$ being antisymmetric, their
    diagonals are null; hence so are the diagonals of $AS$ and $SB$, i.e. $I_k
    \odot (AS) = I_k \odot (SB) = 0$.  Since $S$ is constrained to be
    diagonal, $\dif S$ must also be diagonal, i.e.\ $I_k \odot \dif S = \dif
    S$.  Therefore,  
    \begin{equation*}
        \dif S = I_k \odot (U^\top \dif X V) , 
    \end{equation*}  
as was claimed. 
\end{proof}

Now that the differential of the singular values is available, we are ready to prove
\Cref{lem:gradient-L1-svd}.

\begin{proof}[Proof of~\cref{lem:gradient-L1-svd}]
    For $W \in \cM(k)$, let $USV^\top = \Sigma_0^{1/2} W$ be a
    thin SVD of $\Sigma_0^{1/2} W \eqdef X$. \Cref{lem:reformulation-loss} ensures that 
    \begin{align}
        L^1|_{\cM(k)}(W) 
        &=\norm{W}_F^2 + \norm{\Sigma_0^{1/2}}_F^2 - 2 \trc{S}.
    \end{align}

    According to~\Cref{lem:diff-svd}, the matrix $S$ is differentiable and has
    differential $\dif S = I_k \odot (U^\top \dif X V)$. Therefore, the loss
    $L^1|_{\cM(k)}$ is differentiable. With the fact that $\dif\, \trc S =
    \trc \dif S$ \citep[see, e.g.][Chap.~8, Eq.~18]{Magnus2019Matrix}, we
    can compute 
    \begin{align*}
        \dif\,{\trc S}   &= \trc \dif S  
                     =  \trc {(I_k \odot (U^\top \dif X V))}
                     =  \trc {(U^\top \dif X V)}
                     \\
                     &=  \ang{UV^\top, \dif X}
                     =  \ang{UV^\top, \Sigma_0^{1/2}\dif W}
                     =  \ang{\Sigma_0^{1/2}UV^\top, \dif W}.
    \end{align*}
    Moreover, $\dif {\norm{W}^2_F} = 2\ang{W, \dif W}$, and so 
    \begin{align*}
        \dif {L^1|_{\cM(k)}(W) }
        &= \dif {\norm{W}^2_F} - 2 \dif {\trc S}
        = 2 \ang{W - \Sigma_0^{1/2}UV^\top, \dif W},
    \end{align*}
    and
    \begin{equation*}
        \nabla L^1|_{\cM(k)}(W) = 2(W - \Sigma_0^{1/2}UV^\top).
    \end{equation*}
Since matrices 
$U$ and $V$ are continuously differentiable on $\cM(k)$, $\nabla L^1|_{\cM(k)}(W) = 2(W - \Sigma_0^{1/2}UV^\top)$ is again continuously differentiable, and $L^1|_{\cM(k)}$ is twice continuously differentiable.  
\end{proof}

We are now ready to give the proof of~\Cref{thm:critical-points-L1}. We divide the
proof into necessary and sufficient conditions for a point to be a critical
point of $L^1|_{\cM(k)}$.

\begin{lemma}[Necessary condition on the critical points of 
$L^1|_{\cM(k)}$]\label{lem:necessary-fooc-svd}

    Assume $\Sigma_0$ has $n$ distinct eigenvalues. Let $W^* \in \cM(k)$ be a
    critical point of $L^1|_{\cM(k)}$. Then, with $U^* S^* {V^*}^\top =
    \Sigma_0^{1/2}W^*$ a thin SVD of $\Sigma_0^{1/2}W^*$, and $\Omega \Lambda
    \Omega^\top = \Sigma_0$ a spectral decomposition of $\Sigma_0$ (i.e.\ with
    $\Omega \in \cO(n)$),  there exists $\cJ_k \subseteq [n]$, such that  $S^* = \bar{\Lambda}_{\cJ_k}$ and $U^* = \Omega_{\cJ_k}$.  
\end{lemma}
\begin{proof} 
Since $W^*  \in \cM(k)$,  and $U^* S^* {V^*}^\top = \Sigma_0^{1/2} W^*$ is a thin
SVD of $\Sigma_0^{1/2} W^*$, this means that $S^* \in \bR^{k \times k}$. 
Then, 
\begin{alignat*}{3}
    \nabla L^1(W^*) = 0 
    & \implies 
    & W^* 
    &= \Sigma_0^{1/2} U^* {V^*}^\top, &\qquad &\text{by \eqref{eq:gradient-L1-svd}}
    \\
    &\implies
    & \Sigma_0^{1/2} W^*  
    &= \Sigma_0U^* {V^*}^\top 
    \\
    & \implies
    & U^* S^* {V^*}^\top 
    &= \Sigma_0U^* {V^*}^\top  
    \\ 
    &\implies
    &  S^*
    &= {U^*}^\top \Sigma_0 U^\ast, && {U^*}^\top U^* = I_k,\, {V^*}^\top V^* =
    I_k.
\end{alignat*}
Therefore, ${U^*}^\top \Sigma_0 U^*$ must be diagonal; and since $U^*$ is
semi-orthogonal, this is the case if and only if the vectors in $U^*$ are
eigenvectors for $\Sigma_0$, by uniqueness of the spectral decomposition of
$\Sigma_0$. Therefore, there exist $j_1,
\ldots, j_k$ indices between $1$ and $n$ such that $U^* = \begin{pmatrix}
\omega_{j_1} & \cdots & \omega_{j_k} \end{pmatrix}  = \Omega_{\cJ_k}$, in which
case 
\begin{equation*}
    S^* = {\Omega_{\cJ_k}}^\top \Sigma_0 \Omega_{\cJ_k} = 
    \begin{pmatrix} 
        \lambda_{j_1} & & \\
                      & \ddots & \\
                      & & \lambda_{j_k}
    \end{pmatrix}  = \bar{\Lambda}_{\cJ_k}.
\end{equation*} 
\end{proof}

Now we are ready to prove the first part of \Cref{thm:critical-points-L1}. 

\begin{proof}[Proof of \Cref{thm:critical-points-L1}, first part] 
Consider the expression for the gradient of $L^1|_{\cM(k)}$ given in \eqref{eq:gradient-L1-svd}.  
The necessary condition follows from~\Cref{lem:necessary-fooc-svd}, since  
    \begin{alignat*}{2}
        \nabla L^1|_{\cM(k)}(W^*) = 0 
        &\implies 
        &\Sigma_0^{1/2}W^* 
        &= \Omega_{\cJ_k}\bar\Lambda_{\cJ_k} V^\top 
        \\
        &\implies
        & W^*
        &= \Sigma_0^{-1/2} \Omega_{\cJ_k} \bar\Lambda_{\cJ_k} V^\top 
        \\
        &&
        &= \Omega \Lambda^{-1/2} \Omega^\top \Omega_{\cJ_k} \bar\Lambda_{\cJ_k} V^\top
        \\
        &&
        &= \Omega \Lambda^{-1/2}  \Lambda_{\cJ_k} V^\top
        \\
        &&
        &= \Omega \Lambda^{1/2}P_{\cJ_k}  V^\top
        \\
        &&
        &= \Omega P_{\cJ_k} \bar\Lambda^{1/2}_{\cJ_k} V^\top
        \\
        &&
        &= \Omega_{\cJ_k} \bar\Lambda^{1/2}_{\cJ_k} V^\top,
    \end{alignat*}
    which corresponds to the necessary condition in
    \Cref{thm:critical-points-L1}. 

    The sufficient condition can be verified as follows. With $W^* =
    \Omega_{\cJ_k} \bar\Lambda_{\cJ_k}^{1/2} V^\top$, one has $\Sigma_0^{1/2}W^*
    = \Omega \Lambda^{1/2} \Omega^\top \Omega_{\cJ_k} \bar\Lambda_{\cJ_k}^{1/2}
    V^{\top} = \Omega_{\cJ_k} \bar\Lambda_{\cJ_k} V^\top$, and, as this is a 
    correct thin SVD of $\Sigma_0^{1/2}W^*$, \Cref{lem:gradient-L1-svd} gives
    \begin{align*}
        \nabla L^1|_{\cM(k)}(W^*) 
        &= 2 (W^* - \Sigma_0^{1/2}\Omega_{\cJ_k} V^\top).
        \intertext{Further,}
        \Sigma_0^{1/2}\Omega_{\cJ_k} 
        &=  \Omega \Lambda^{1/2} \Omega^\top \Omega_{\cJ_k}
        \\
        &= \Omega \Lambda^{1/2} P_{\cJ_k} 
        \\
        &= \Omega P_{\cJ_k} \bar\Lambda^{1/2}_{\cJ_k}
        \\
        &= \Omega_{\cJ_k} \bar\Lambda^{1/2}_{\cJ_k}. 
            \intertext{Hence}
        \nabla L^1|_{\cM(k)}(W^*) = 2 (W^* - \Sigma_0^{1/2}\Omega_{\cJ_k} V^\top) 
        &= 2(\Omega_{\cJ_k} \bar \Lambda^{1/2}_{\cJ_k} V^\top - \Omega_{\cJ_k}
        \bar \Lambda^{1/2}_{\cJ_k} V^\top) = 0,
    \end{align*}
    and the sufficient condition is verified. 
\end{proof}

Now, the loss can be evaluated at the critical points in order to identify its minimizers. 

\begin{corollary}[Value of $L^1$ at the critical points] 
    \label{cor:value-crit-loss} The value of the loss $L^1$ at a critical
    point $W^* = \Omega_{\cJ_k} \bar{\Lambda}^{1/2}_{\cJ_k} V^\top$ is
    $L^1(W^*) = \trc \Lambda - \trc \bar{\Lambda}_{\cJ_k} = \sum_{i \notin
    \cJ_k} \lambda_i$. 
\end{corollary}

\begin{proof}
    For $k \geqslant 0$, let $W^*$ be a critical point of $L^1 |_{\cM(k)}$.
    From~\Cref{thm:critical-points-L1}, with $\Sigma_0 =
    \Omega \Lambda \Omega^\top$ a spectral decomposition of $\Sigma_0$, there exists a set $\cJ_k$ and a
    semi-orthogonal matrix $V \in \bR^{n \times k}$ such that $W^*
= \Omega_{\cJ_k} \bar{\Lambda}^{1/2}_{\cJ_k} V^\top$.  One can then compute the
value of the loss at $W^*$:
\begin{align*}
    L^1(W^*) 
    &= \trc{W^* {W^*}^\top} + \trc{\Sigma_0} - 2 \trc{\Big(
    {(\Sigma_0^{1/2} W^* )}{(\Sigma_0^{1/2}W^*)}^\top\Big)}^{1/2}
    \\
    &= \trc{\Omega_{\cJ_k} \bar{\Lambda}_{\cJ_k} \Omega_{\cJ_k}} +
    \trc{\Lambda} - 2 \trc{\Big(
    \Omega_{\cJ_k} \bar{\Lambda}_{\cJ_k}^2 \Omega_{\cJ_k}^\top \Big)}^{1/2}
    \\
    &= \trc{\bar{\Lambda}_{\cJ_k}} + \trc{\Lambda}  - 2 \trc{\bar{\Lambda}_{\cJ_k}}
    \\
    &= \trc{\Lambda} - \trc{\bar{\Lambda}_{\cJ_k}} . 
\end{align*}
\end{proof}

We now have all ingredients needed to prove the second part of \Cref{thm:critical-points-L1}. 

\begin{proof}[Proof of \Cref{thm:critical-points-L1}, second part]
    \label{prf:EYM-BW}
    The first part of the statement is readily implied by
    \Cref{cor:value-crit-loss}, as the eigenvalues are in decreasing order. The
    second part is implied by the fact that the minimum $L^1|_{\cM(k)}$ is
    indeed achieved for any $k \leqslant n$ (by selecting the $k$ largest
    eigenvalues of $\Sigma_0$) and the optimal value of the loss $L^*_k$ is
    smaller when considering more eigenvalues, i.e.\ $\min_{\cM(k)} L^1
    \leqslant \min_{\cM(<k)} L^1$.
\end{proof}

Next we show that only one point per set $\cM(k)$ is a minimizer of the loss $L^1|_{\cM(k)}$ and all other points are (strict) saddle points. 
We recall the definition of a strict saddle point: a point where there exists a descent direction. 

\begin{definition}[Strict saddle point]
    \label{def:strict-saddle-point}
A critical point $x$ of a function $f$ is said to be a \emph{strict saddle point} if the Hessian of
$f$ at $x$ has a
    strict negative eigenvalue. If all critical points of $f$ are either
    a strict saddle point or 
    a global minimizer, the we say that $f$ satisfies the \emph{strict saddle point property}.
\end{definition}

If the gradient flow can be expressed on a manifold, with a Riemannian gradient corresponding to a given metric, there is an equivalent definition of those saddle points, which will be handy to use. We refer to \citet[\S 6.1]{Bah2021Learning} for the details. 

\begin{proposition}
    \label{prop:strict-saddle-L1}
    The loss $L^1|_{\cM(k)}$ satisfies the strict saddle point property. 
\end{proposition}
\begin{proof}
    Let $\Sigma_0 = \Omega\Lambda \Omega^\top$ be the spectral decomposition of $\Sigma_0$
    with decreasing eigenvalues. For $k\in \bN$, according to
    \Cref{thm:critical-points-L1}, $W^*$ is a critical point of $L^1|_{\cM(k)}$
    if and only if there exists $\cJ_k \subset [n]$, such that
    $W^* = \Omega_{\cJ_k} \Lambda^{1/2}_{\cJ_k} V^\top$, with any $V \in \bR^{m
    \times k}$ so that $V^\top V = I_k$. If $\cJ_k =
    [k]$, $W^*$ is a global minimum of $L^1|_{\cM(k)}$, as shown
    in \Cref{cor:value-crit-loss}, and the proposition holds.

    Assume $\cJ_k \neq [k]$, then there exists  $j_0\in\cJ_k$ such that
    $\lambda_{j_0}<\lambda_k$, and there exists $j_1\notin \cJ_k$ but
    $j_1\in[k]$ such that $\lambda_{j_1}>\lambda_{j_0}$. We will
    show that $W^*$ is a strict saddle point of $L^1|_{\cM(k)}$. 

    The critical point $W^*$ can equivalently be expressed as 
\begin{equation}
    W^* = \Sigma_0^{-1/2}\sum_{i\in \cJ_k} \lambda_i  \omega_i  v_i^\top,
\end{equation}
 where $\omega_i, v_i$ are corresponding orthonormal vectors in $\Omega$ and $V$, and $\lambda_i$ are eigenvalues in $\Lambda$.

For $t\in(-1,1)$, we define 
\begin{equation*}
    \omega_{j_0}(t) = t \omega_{j_1}+\sqrt{1-t^2} \omega_{j_0} 
\end{equation*}
and the curve $\gamma :(-1,1)\mapsto \cM(k)$. 
We look at the perturbed matrix 
\begin{equation*}
    \gamma(t) =  \Sigma_0^{-1/2}\Big( \lambda_{j_0} \omega_{j_0}(t) v_{j_0}^\top+\sum_{i\in \cJ \setminus\{j_0\}} \lambda_i \omega_i  v_i^\top\Big).
\end{equation*} 
Note that $\gamma(0)=W$. Recall $L^1(W) = \trc\big(W W^\top + \Sigma_0 -2(\Sigma_0^{1/2} W W^\top\Sigma_0^{1/2})^{1/2}\big)$.  
It is enough to show that \citep[\S 6.1]{Bah2021Learning}:
\begin{equation*}
    \frac{d^2}{dt^2} L^1(\gamma(t))\Big|_{t=0}<0.
\end{equation*} 
We check it term by term, 
\begin{equation*}
\begin{aligned}
    \trc\Big(\gamma(t)\gamma(t)^{\top}\Big) &= \trc\Big(\Sigma_0^{-1/2}\big( \lambda_{j_0} \omega_{j_0}(t) v_{j_0}^\top+\sum_{i\in \cJ \setminus\{j_0\}} \lambda_i \omega_i  v_i^\top\big)\big( \lambda_{j_0} \omega_{j_0}(t) v_{j_0}^\top+\sum_{i\in \cJ \setminus\{j_0\}} \lambda_i \omega_i  v_i^\top\big)^{\top} \Sigma_0^{-1/2}\Big)\\
    & = \trc\Big(\Sigma_0^{-1}\big(\lambda_{j_0}^2 \omega_{j_0}(t)\omega_{j_0}(t)^{\top}+\sum_{i\in \cJ \setminus\{j_0\}} \lambda_i^2 \omega_i \omega_i^{\top}\big)\Big)\\
    & = \trc\Big(\big(\sum_{1\leqslant i\leqslant n}\lambda_i^{-1} \omega_i \omega_i^{\top} \big)\big(\lambda_{j_0}^2 \omega_{j_0}(t)\omega_{j_0}(t)^{\top}+\sum_{i\in \cJ \setminus\{j_0\}} \lambda_i^2 \omega_i \omega_i^{\top}\big)\Big)\\
    & = \frac{\lambda_{j_0}^2}{\lambda_{j_1}} t^2 +\lambda_{j_0} (1-t^2) + \sum_{i\in \cJ \setminus\{j_0\}} \lambda_i^2,
\end{aligned}
\end{equation*}
and 
\begin{equation*}
\begin{aligned}
    \trc\Big(&\big(\Sigma_0^{1/2}
\gamma(t) \gamma(t)^\top\Sigma_0^{1/2}\big)^{1/2}\Big)\\
&= \trc\Big(\big(\big( \lambda_{j_0} \omega_{j_0}(t) v_{j_0}^\top+\sum_{i\in \cJ \setminus\{j_0\}} \lambda_i \omega_i  v_i^\top\big)\big( \lambda_{j_0} \omega_{j_0}(t) v_{j_0}^\top+\sum_{i\in \cJ \setminus\{j_0\}} \lambda_i \omega_i  v_i^\top\big)^{\top}\big)^{1/2}\Big)\\
& = \trc\Big( \big(\lambda_{j_0}^2 \omega_{j_0}(t)\omega_{j_0}(t)^{\top}+\sum_{i\in \cJ \setminus\{j_0\}} \lambda_i^2 \omega_i \omega_i^{\top}\big)^{1/2}\Big)\\
&=\trc\Big( \big(t^2\lambda_{j_0}^2 \omega_{j_1}\omega_{j_1}^{\top}+(1-t^2)\lambda_{j_0}^2u_{j_0}\omega_{j_0}^{\top}+\sum_{i\in \cJ \setminus\{j_0\}} \lambda_i^2 \omega_i \omega_i^{\top}\big)^{1/2}\Big)\\
&= t|\lambda_{j_0}| + \sqrt{1-t^2}|\lambda_{j_0}|+\sum_{i\in \cJ\setminus\{j_0\}} |\lambda_i|.
\end{aligned}
\end{equation*}
Thus, since $\lambda_{j_1}>\lambda_{j_0}$, 
\begin{equation*}
    \frac{d^2}{dt^2} L^1(\gamma(t))\Big|_{t=0} = 2(
    \lambda_{j_0}^2 \lambda_{j_1}^{-1}- \lambda_{j_0}) -|\lambda_{j_0}|<0.
\end{equation*}
This completes the proof. 
\end{proof}

\subsection{Critical points of the perturbative loss $L^1_\tau|_{_k}$}
\label{app:critical-L1tau}

In this section, we provide the 
derivations for \Cref{sec:crit-points-pertloss}. The structure of reasoning is
similar to the one found in the proof of \Cref{thm:critical-points-L1}: first the gradient of $L^1_\tau$ is computed, 
then the critical points are characterized 
and ordered. 

\begin{lemma}[Gradient of $L^1_\tau$]
    The loss  $L^1_\tau$ has the following gradient 
    \begin{equation}
        \label{eq:grad-L1tau}
        \forall\: W \in \bR^{n\times m},\quad \nabla L^1_\tau (W) = 2\big(W - \Sigma_0^{1/2}{\big[ \Sigma_0^{1/2} (
    WW^\top + \tau I_n) \Sigma_0^{1/2}\big]}^{-1/2} \Sigma_0^{1/2} W\big). 
\end{equation}
\end{lemma}

\begin{proof}
    This results comes from the chain rule for the loss $L^1_\tau(W) =
    L \circ \varphi_\tau \circ \pi(W)$. With $\Sigma = \pi(W) = WW^\top$ and $\Sigma_\tau =
    \varphi_\tau(\Sigma) = \Sigma + \tau I_n$, 
    and since 
    $\dif{}\pi(W)[Z] = WZ^\top + ZW^\top$ and $\dif{}\varphi_\tau(\Sigma) =
    \id$, one has 
    \begin{alignat*}{2}
        &&\dif{} L^1_\tau(W)[Z]
        &= \dif{} (L \circ \varphi_\tau
        \circ \pi) (W) [Z]
        \\
        &&&= \dif{} L(\Sigma_\tau)\bigg[
        \dif{}\varphi_\tau(\Sigma)\Big[\dif{}\pi (W)[Z]\Big]\bigg]
        \\
        &&&= \dif{} L(\Sigma_\tau)[WZ^\top + ZW^\top]
        \\
        &&\ang{\nabla L^1_\tau(W), Z} &= \ang{\nabla L(\Sigma_\tau), WZ^\top + ZW^\top}
        \\
        &\iff \quad
        & \nabla L^1_\tau(W) &= (\nabla L(\Sigma_\tau) +
        {\nabla L(\Sigma_\tau)}^\top)W 
        \\
        &&&= 2 (W - \Sigma_0^{1/2}{[\Sigma_0^{1/2}\Sigma_\tau
    \Sigma_0^{1/2}]}^{-1/2}\Sigma_0^{1/2}W).
    \end{alignat*}
\end{proof}

With the expression of the gradient of $L^1_\tau$ available, \Cref{thm:critical-points-L1tau}
can be proven.

\begin{proof}[Proof of~\Cref{thm:critical-points-L1tau}]
    \label{prf:critical-points-L1tau}
The eigenvectors of $WW^\top + \tau$ are the same as $WW^\tau$, and the
eigenvalues are shifted by $\tau$. Therefore, the expression of the critical
points in the original loss can be adapted, so that the modified critical
points have the same left singular vectors and shifted singular values. This
leads to having $W^* = \Omega_{\cJ_k}{(\bar{\Lambda}_{\cJ_k} - \tau
I_k)}^{1/2} V_{\parallel}^\top =
    \begin{pmatrix}\Omega_{\cJ_k}&\bm{0}_{n \times n-k}\end{pmatrix}  \begin{pmatrix} {(\bar{\Lambda}_{\cJ_k} - \tau
        I_k)}^{1/2} & \\ & \bm{0}_{n-k \times m-k} \end{pmatrix}
        \begin{pmatrix}V_{\parallel} &V_{\bot}\end{pmatrix}^\top $, with $V =
        \begin{pmatrix} V_{\parallel} & V_\bot \end{pmatrix} \in \bR^{m \times
        m}$, such such that $V^\top V = V V^\top= I_m$.In the following, we will
        make sure that  $\nabla L^1_\tau(W^*) = 0$.

        Indeed, assume without loss of generality that $\Omega = \begin{pmatrix} \Omega_{\cJ_k}
    & \Omega_{\cJ_k^c}\end{pmatrix}$ (and $\Lambda = \begin{pmatrix}
        \bar{\Lambda}_{\cJ_k} & \\ & \bar{\Lambda}_{\cJ_k^c} \end{pmatrix}$), where
        $\cJ_k^c \defeq [n] \setminus \cJ_k$ for $\cJ_k \subseteq [n]$. Then,
        \begin{align*}
            W^*{W^*}^\top 
        &= 
        \Omega_{\cJ_k}( \bar{\Lambda}_{\cJ_k} - \tau I_k)
        \Omega_{\cJ_k}^\top = \Omega \begin{pmatrix} \bar{\Lambda}_{\cJ_k} -
            \tau I_k & \\
                     & \bm{0}_{n-k \times n-k} \end{pmatrix}\Omega^\top , \\
                     \Sigma^*_\tau 
                     \defeq W^*{W^*}^\top  + \tau I_n 
        &=  W^*{W^*}^\top + \tau \Omega \Omega^\top =
            \Omega \begin{pmatrix} \bar\Lambda_{\cJ_k} & \\ & \tau I_{n-k} \end{pmatrix} 
            \Omega^\top, 
            \\
            \Sigma_0^{1/2} \Sigma^*_\tau \Sigma_0^{1/2}
                                                       &=\Omega \Lambda^{1/2} \Omega^\top\Omega \begin{pmatrix} \bar\Lambda_{\cJ_k} & \\ & \tau I_{n-k} \end{pmatrix} 
                                                       \Omega^\top\Omega \Lambda^{1/2} \Omega^\top =  \Omega \begin{pmatrix} \bar\Lambda_{\cJ_k}^2 
        & \\ & \tau \bar\Lambda_{\cJ_k^c} \end{pmatrix}\Omega^\top,
        \\
        \Sigma_0^{1/2}{(\Sigma_0^{1/2} \Sigma^*_\tau \Sigma_0^{1/2})}^{-1/2}
        \Sigma_0^{1/2} 
                                                       &= \Omega \Lambda^{1/2} \Omega^\top \Omega \begin{pmatrix}
                                                           \bar\Lambda_{\cJ_k}^{-1} 
       & \\ & {(\tau \bar \Lambda_{\cJ_k^c})}^{-1/2} \end{pmatrix}  \Omega^\top
       \Omega \Lambda^{1/2} \Omega^\top   
       = \Omega \begin{pmatrix} I_k
       & \\ & \tau^{-1/2} \bar \Lambda_{\cJ_k^c}^{1/2} \end{pmatrix}
       \Omega^\top,
       \intertext{and}
       I_n - \Sigma_0^{1/2}{(\Sigma_0^{1/2} \Sigma^*_\tau \Sigma_0^{1/2})}^{-1/2} \Sigma_0^{1/2} 
                                                       &=
            \Omega \begin{pmatrix} \bm{0}_{k \times k} &  \\ & I_{n-k} - \tau^{-1/2}
            \bar \Lambda_{\cJ_k^c}^{1/2} \end{pmatrix} \Omega^\top.
            \end{align*}
            Since $\Omega^\top \Omega_{\cJ_k} = I_{\cJ_k} = \begin{pmatrix} A \\
            \bm{0}_{n-k \times k} \end{pmatrix} $, with $A \in \bR^{k \times k}$, the
            gradient evaluates to 
            \begin{align*}
                \nabla L^1_\tau(W^*) 
        &= 2(I_n - \Sigma_0^{1/2}{(\Sigma_0^{1/2}\Sigma_\tau^*\Sigma_0^{1/2})}^{-1/2}
        \Sigma_0^{1/2}) \Omega_{\cJ_k}{(\bar\Lambda_{\cJ_k} - \tau I_k)}^{1/2}
        V_{\parallel}^\top 
        \\
        &= 
        2\Omega \begin{pmatrix} \bm{0}_{k \times k} 
        &  \\ & I_{n-k} - \tau^{-1/2} \bar \Lambda_{\cJ_k^c}^{1/2} \end{pmatrix} 
        \begin{pmatrix} A \\ \bm{0}_{n-k \times k} \end{pmatrix} {(\bar \Lambda_{\cJ_k} - \tau I_k)}^{1/2} V_{\parallel}^\top
        \\
        &= 0.
            \end{align*}
    For such a critical point $W^* = \Omega_{\cJ_k}{(\bar{\Lambda}_{\cJ_k} - \tau
    I_k)}^{1/2} V^\top $, with regularized covariance $\Sigma_\tau^* = W^*
    {W^*}^\top + \tau I_n$, the value of the loss is 
    \begin{align*}
        L^1_\tau(W^*) &= \trc{\Sigma^*_\tau} + \trc{\Sigma_0} -  2\trc
        {(\Sigma_0^{1/2} \Sigma_\tau^* \Sigma_0^{1/2})}^{1/2} 
        \\
                      &= \sum_{j \in \cJ_k} \lambda_j + \tau (n-k) +
                      \sum_{j \in \cJ_k} \lambda_j + \sum_{j \in \cJ_k^c}
                      \lambda_j -  2 (\sum_{j\in \cJ_k} \lambda_j + 
                      \sum_{j\in \cJ_k^c}  \sqrt{\tau\lambda_j})
                      \\
                      &= \sum_{j \in \cJ_k^c} \lambda_j + \tau -
                      2\sqrt{\tau\lambda_j } 
                      \\
                      &= \sum_{j \in \cJ_k^c } {(\sqrt{\lambda_j}-\sqrt{\tau})}^2,
    \end{align*} which is uniquely minimized of $\cJ_k$ for $\cJ_k
    = [k]$ when the eigenvalues of $\Sigma_0$ are distinct and in descending
    order.

    Moreover, as in the unregularized case, we have the increasing sequence of
    minimizers $\min_{ \cM(k)} L^1_\tau \leqslant  \min_{ \cM(< k)} L^1_\tau$
    which, together with the identity $\cM(\leqslant k) = \cM(k) \cup \cM(<k)$,
    implies that $\min_{\cM(\leqslant k)} L^1_\tau = \min_{\cM(k)} L^1_\tau$.
\end{proof}

The loss $L^1_\tau$ satisfies the strict-saddle point property in a similar fashion as \Cref{prop:strict-saddle-L1} for $L^1$. 
\begin{lemma}
    \label{lem:strict-saddle-L1tau}
    The loss $L^1_\tau|_{\cM(k)}$ satisfies the strict saddle point property.
\end{lemma}
\begin{proof}[Proof of \Cref{lem:strict-saddle-L1tau}]
    \label{prf:strict-saddle-L1tau}
    The proof of \Cref{prop:strict-saddle-L1} can be adapted, with the
    expression of the critical points as, if $\Sigma_0 = \Omega \Lambda
    \Omega^\top$, and with $V \in \bR^{n \times k}$ any semi-orthogonal matrix, $W^* = {(\Sigma_0 - \tau I_n)}^{-1/2}
    \sum_{j=1}^n (\lambda_i - \tau) \omega_i v_i^\top$.
\end{proof}

We are now ready to prove \Cref{prop:crit-param-func}.
\begin{proof}[Proof of \Cref{prop:crit-param-func}]
    \label{prf:crit-param-func}
    The fact that $W^* = \mu(\W^*)$ is a critical point of $L^1_\tau |_{\cM(k)}$
    (with $k = \rank W^*$) if and only if $\W^*$ is a critical point for
    $L^N_\tau$, as well as the fact that, when $k = \ud = \min_i(d_i)$, $W^*$ is
    a local minimizer of $L^1_\tau |_{\cM(\ud)}$ if and only if $\W^*$ is a local
    minimizer of $L^N_\tau$ are straightforwardly deduced from \citet[Proposition
    6]{Trager2020Pure}, since $L^1_\tau$ is smooth.

    The additional fact that any local minimizer of rank $\ud$ of
    $L^1_\tau|_{\cM(\ud)}$ is a global minimizer of $L^1_\tau|_{\cM(\ud)}$
    comes from \Cref{lem:strict-saddle-L1tau}: $L^1_\tau|_{\cM(\ud)}$ satisfies
    the strict saddle point property, therefore, the only critical points of
    $L^1_\tau|_{\cM(\ud)}$ are strict saddle points and the global minimizer.

    Now, the expression of such a global minimizer is given by
    \Cref{thm:critical-points-L1tau}: with $\Sigma_0 = \Omega \Lambda
    \Omega^\top$ a spectral decomposition of $\Sigma_0$ in descending order
    of the eigenvalues, there exists $V \in \cO(m)$ orthogonal, such that $W^*
    = \Omega_{[\ud]} {(\bar\Lambda_{[\ud]} - \tau I_{\ud})}^{1/2}
    V_{[\ud]}^\top$, and $\Sigma_\tau^* = W^*{W^*}^\top + \tau I_n = \Omega
    \begin{pmatrix} \Lambda_{[\ud]} & \\ & \tau \end{pmatrix} \Omega^\top$. 
\end{proof}

\section{Proofs of \Cref{sec:convergence}}
\label{app:proofs-convergence}

In this section, we provide the proofs of the convergence statements in \Cref{thm:convergenceGF,thm:gradientdescent}. 

\subsection{Bounds on the Hessian of $L^1_\tau$}
\label{app:bounds-hessian}
In this section, we provide bounds on the Hessian of the perturbative loss
$L^1_\tau$. We first compute the Hessian the loss $L$ as a function of the
covariance matrix, as given by \citet[Lemma A.2]{Kroshnin2021Statistical}. Then, a simple
chain rule for the differential allows to express the Hessian in the case the
loss is a function of the end-to-end matrix $W$.

Denoting $\Sigma_\tau := WW^{\top} + \tau I_{n}$ the regularized
covariance matrix, the loss $L$ can be expressed in terms of the optimal transport
plan between $\Sigma_{\tau}$ and $\Sigma_0$ \citep[Proposition 2.1]{Kroshnin2021Statistical}.  We
have 
\begin{equation}\label{opttrloss}
    \begin{aligned}
        L(\Sigma_\tau) =& \trc\big(\Sigma_\tau + \Sigma_0 - 2(\Sigma_0^{1/2}
        \Sigma_\tau \Sigma_0^{1/2} )^{1/2}\big) \\
            =&  \|\big(T_{\Sigma_{\tau}}^{\Sigma_0} -I\big)\Sigma_{\tau}^{1/2}\|_{F}^{2}  \\
            =& \trc{\big(T_{\Sigma_{\tau}}^{\Sigma_0} -I\big)\Sigma_{\tau}{\big(T_{\Sigma_{\tau}}^{\Sigma_0}} -I\big)},
        \end{aligned}
    \end{equation}
    where $T_{\Sigma_{\tau}}^{\Sigma_0} = \Sigma_0^{1/2}\big(\Sigma_{0}^{1/2}\Sigma_{\tau}\Sigma_{0}^{1/2}\big)^{-1/2}\Sigma_{0}^{1/2}=\Sigma_{\tau}^{-1/2}\big(\Sigma_{\tau}^{1/2}\Sigma_{0}\Sigma_{\tau}^{1/2}\big)^{1/2}\Sigma_{\tau}^{-1/2}$.

    This expression of the loss allows to compute its second order differential.

\begin{lemma}[Second-order differential of $L_\tau$, {\citealt[Lemma
A.6]{Kroshnin2021Statistical}}]

\label{lem:sod-L}

Let $W \in \bR^{n \times m}$ and let $\tau >0$. Define $\Sigma_\tau = WW^\top +
\tau I_n$ to be the regularized covariance matrix.  Given that $\Sigma_\tau
\succ 0$, the loss $L$ given by~\eqref{opttrloss} is twice continuously
differentiable at $\Sigma_\tau$. Let $\Gamma Q \Gamma^\top =
\Sigma_0^{1/2}\Sigma_\tau \Sigma_0^{1/2}$ be a spectral decomposition of
$\Sigma_0^{1/2}\Sigma_\tau \Sigma_0^{1/2}$, with $Q = \Diag{{(q_1, \ldots,
q_n)}}$. For $Y \in \cS_{++}(n)$, define $\Delta(Y) \in \cS(n)$ to be the matrix
with element ${\Delta(Y)}_{ij} = {(\sqrt{q_i} + \sqrt{q_j})}^{-1} (\Gamma^\top
\Sigma_0^{1/2}Y \Sigma_0^{1/2}\Gamma)_{ij}$. Let $\bG_{\tau}$ be the
linear operator defined as 
\begin{equation}
    \label{eq:def-G}
    \begin{array}{rcl}
        \bG_{\tau} \colon \quad \cS_{++}(n) &\longrightarrow& \cS(n) \\
        Y & \longmapsto & \bG_{\tau}(Y) =\Sigma_0^{1/2} \Gamma Q^{-1/2} \Delta(Y)  Q^{-1/2} \Gamma^\top
    \Sigma_0^{1/2}.
\end{array}
\end{equation}
Then, the second order differential of $L_\tau$ is given by
\begin{equation}
\forall(X,Y) \in {\cS_{++}(n)}^2,\quad \dif{}^2
    L_\tau(\Sigma_\tau)[X, Y] = \ang{X, \bG_{\tau}(Y)}.
\end{equation} 
\end{lemma}

\begin{proof}
    For completeness, we provide a proof of the statement different from the one by~\citet{Kroshnin2021Statistical}. 
    We begin by stating the first-order differential for the loss
    $L$ evaluated on the PD matrix $\Sigma_\tau$. This is given in \cref{lem:fod-L1tilde} 
    \begin{align*}
    \dif{} L(\Sigma_\tau) [X]
        &= \trc(X - \Sigma_0^{1/2} {(
        \Sigma_0^{1/2} \Sigma_\tau \Sigma_0^{1/2})}^{-1/2}
        \Sigma_0^{1/2} X)  
        \\
        &= \ang{I - \Sigma_0^{1/2} {(
            \Sigma_0^{1/2} \Sigma_\tau \Sigma_0^{1/2})}^{-1/2}
    \Sigma_0^{1/2}, X}.  
    \end{align*}

    Let $\GL(n) = \set{A \in \bR^{n\times n} \mid \det A \neq 0}$, and let $f
    \colon \GL(n) \ni F \mapsto F^{-1}$; then $f$ is differentiable with differential $\dif{} f(F)[X] = -F^{-1}X F^{-1}$ \citep[Theorem
    8.3]{Magnus2019Matrix}. 
    Let $g \colon \cS_{++}^n \ni A \mapsto A^{1/2}$ be the matrix square root.
    The function $g$ is differentiable on $ \cS_{++}(n)$, and its differential can be computed as
    follows \citep[Lemma A.1]{Kroshnin2021Statistical}. 
    Let $A \in \cS_{++}(n)$, and let
    $\Gamma Q \Gamma^\top$ be its spectral decomposition, with $Q =
    \Diag(q_1, \ldots, q_n)$. For $X \in \cS(n)$, define $\delta(X) \in \bR^{n\times n}$ to be the matrix with elements
    ${\delta(X)}_{ij} \defeq {(\sqrt{q_i} +
    \sqrt{q_j})}^{-1}(\Gamma^\top X \Gamma)_{ij}$. Then,  the differential of $g$ at $A$ in the direction $X$ is 
$\dif{}g(A)[X] =     \Gamma \delta (X) \Gamma^\top$.

Therefore,  the chain rule on the differentials gives
\begin{equation*}
    \dif{} (f\circ g)(A)[X] = \dif f(g(A)) [ \dif g(A)[X]]  = - A^{-1/2} \dif
{}g (A)[X]
    A^{-1/2} = -A^{-1/2} \Gamma \delta(X) \Gamma^\top A^{-1/2},
\end{equation*} 
and, with $A = \Sigma_0^{1/2} \Sigma_\tau \Sigma_0^{1/2}$, 
\begin{align*}
    \dif{}^2 L(\Sigma_\tau) [X, Y]
    &= \dif{}(\Sigma_\tau \mapsto \dif{} L(\Sigma_\tau)[X])[Y]
    \\
    &=  \dif{} (\trc(X - (\Sigma_0^{1/2} {(
        \Sigma_0^{1/2} \Sigma_\tau \Sigma_0^{1/2})}^{-1/2}
        \Sigma_0^{1/2} X)))[Y]
        \\
    &= - \trc( \Sigma_0^{1/2} (\dif{}{(
    \Sigma_0^{1/2} \Sigma_\tau \Sigma_0^{1/2})}^{-1/2}[Y])
        \Sigma_0^{1/2} X)
        \\
    &=- \trc( \Sigma_0^{1/2} (-A^{-1/2} \Gamma \delta(\dif{} (\Sigma_0^{1/2}
    \Sigma_\tau \Sigma_0^{1/2})[Y]) \Gamma^\top A^{-1/2}) \Sigma_0^{1/2} X)
    \\
    &= \trc( \Sigma_0^{1/2} \Gamma Q^{-1/2} \delta(\Sigma_0^{1/2}Y\Sigma_0^{1/2}) Q^{-1/2} \Gamma^\top
    \Sigma_0^{1/2} X)
    \\
    &= \ang{X, \bG_{\tau}(Y)}
\end{align*}
with 
$$
\bG_{\tau}(Y) = \Sigma_0^{1/2} \Gamma Q^{-1/2} \Delta(Y) Q^{-1/2} \Gamma^\top
\Sigma_0^{1/2}
$$ 
and 
$$
{\Delta(Y)}_{ij} = {\delta(\Sigma_0^{1/2} Y \Sigma_0^{1/2})}_{ij} = (\sqrt{q_i} + \sqrt{q_j})^{-1} (\Gamma^\top \Sigma_0^{1/2} Y \Sigma_0^{1/2} \Gamma)_{ij}.
$$
\end{proof}

In order to express the Hessian of the loss as a function of the end-to-end matrix $W$, we need the chain rule for the second-order differential. 
We first recall the chain rule for the second-order differential.

\begin{lemma}[Chain rule for second-order differential, {\citealt[Theorem
6.9]{Magnus2019Matrix}}]
\label{lem:sod-chainrule}
    Let $f\colon R \to S$ and $g\colon S \to T$ be two differentiable
    functions on open sets, such that $h = g \circ f\colon R \to T$ is always
    well defined. Then, given
    two directions $u, v$, the second-order differential of $h$ at $c$ is 
    \begin{equation}
    \label{eq:sod-chainrule}
        \dif{}^2 h(c)[u,v] = \dif{}^2 g(f(c))\big[\dif{} f(c)[u], \dif{}
            f(c)[v]\big] +
        \dif{} g(f(c))[\dif{}^2 f(c)[u, v]] . 
    \end{equation}
\end{lemma}

With this computation rule, we are able to give the second-order differential of $L^1_\tau = L_\tau  \circ \pi$.

\begin{lemma}[Second-order differential of $L^1_\tau$] 
    \label{lem:sod-L1}
    Let $W\in\bR^{n \times m}$. For any $U, V \in \bR^{n\times m}$, the second
    order differential of $L^1_\tau$ at W in the directions $U, V$ is 
    \begin{equation*}
        \dif{}^2 L^1_\tau(W)[U, V] = \ang{U, \bH_{\tau}(V)} , 
    \end{equation*}
    where 
    \begin{equation}
        \label{eq:def-H}
        \bH_{\tau}(V) =  2(\bG_{\tau}(VW^\top + WV^\top)W + (I -
        \Sigma_0^{1/2}{(\Sigma_0^{1/2} \Sigma_\tau \Sigma_0^{1/2})}^{-1/2}
        \Sigma_0^{1/2})V),
    \end{equation}
    and $\bG_{\tau}$ is defined as in~\eqref{eq:def-G}. 
\end{lemma}

\begin{proof}
    Applying the formula~\eqref{eq:sod-chainrule} to $L^1_\tau =
    L_\tau \circ \pi$ gives,
    with $\Sigma = \pi(W)$ and $\dif{}^2\pi(W)[U, V] = UV^\top + VU^\top$, 
    \begin{align*}
        \dif{}^2 L^1_\tau(W)[U, V] 
        &= \dif{}^2 L_\tau(\Sigma)[\dif{}
        \pi(W)[U], \dif{}\pi(W)[V]]  + \dif{}L_\tau(\Sigma) [\dif{}^2
        \pi(W)[U, V]]
        \\
        &= \ang{UW^\top + WU^\top, \bG_{\tau}(VW^\top + WV^\top)} + \trc{(UV^\top +
            VU^\top)}\\
            &~~~ - \trc{\Sigma_0^{1/2}{(\Sigma_0^{1/2}\Sigma_\tau
        \Sigma_0^{1/2})}^{-1}\Sigma_0^{1/2} (UV^\top + VU^\top)}
        \\
        &= 2 \ang{U, \bG_{\tau}(VW^\top + WV^\top)W + V - \Sigma_0^{1/2}{(\Sigma_0^{1/2}\Sigma_\tau
        \Sigma_0^{1/2})}^{-1}\Sigma_0^{1/2} V }
        \\
        &= \ang{U, \bH_{\tau}(V)}, 
    \end{align*}
where we used the symmetry of $\Sigma_0^{1/2}{(\Sigma_0^{1/2}\Sigma_\tau
\Sigma_0^{1/2})}^{-1/2}\Sigma_0$ to simplify the expression.
\end{proof}

The maximal eigenvalue of $\bH_{\tau}$ will be computed in \Cref{lem:bound-H}. But first,
we study the eigenvalues of $\bG_{\tau}$.

\subsection{Lipschitz-smoothness of \texorpdfstring{$L^1_\tau$}{L1τ}}
\label{sec:lipschitz-Ltau}

The  aim of this section is to study the Lipschitz-smoothness of the loss
$L^1_\tau$. For that, we will study the spectrum of its Hessian operator,
and the closely related Hessian operator of $L_\tau$. We first recall the
definition we take for the eigenvalues of those matrix operators.

\begin{definition}[Eigenvalues of matrix operators]
    \label{def:extremal-eigenvals}
    Let $\bF \colon \bR^{p\times q} \to \bR^{r \times s}$ be a linear operator. Then, its
    extremal eigenvalues $\lambda_{\max}(\bF),\, \lambda_{\min}(\bF)$ are
    defined as 
    \begin{equation*}
        \lambda_{\max}(\bF) \defeq 
        \sup_{U \in \bR^{p \times q}\colon \norm{U}_F=1} \ang{U,\bF(U)}, \quad
        \lambda_{\min}(\bF) \defeq \inf_{U \in \bR^{p \times q} \colon \norm{U}_F=1}
        \ang{U, \bF(U)}. 
    \end{equation*}
\end{definition} 
    
One can use the bounds of \citet[Lemma A.3]{Kroshnin2021Statistical} to bound the Hessian of the loss. 

\begin{lemma}[Bounds on the second-order differential, {\citealt[Lemma A.3]{Kroshnin2021Statistical}}]
    \label{lem:bound-sod}

    Let $\bG_{\tau}(X)$ be defined as in~\eqref{eq:def-G}. The second-order differential of
    $L_\tau$ respects
    the following bounds
    \begin{subequations}
\begin{align}
    \ang{X, \bG_{\tau}(X)} 
    &\leqslant \frac{\lambda_{\max}^{1/2}(\Sigma_0^{1/2} \Sigma_\tau
    \Sigma_0^{1/2})}{2} \norm{\Sigma_\tau^{-1/2} X \Sigma_\tau^{-1/2}}^2_F,
    \label{eq:bound-Htilde}
    \\
    \ang{X, \bG_{\tau}(X)} & \geqslant \frac{\lambda_{\min}^{1/2}(\Sigma_0^{1/2}
        \Sigma_\tau \Sigma_0^{1/2})}{2} \norm{\Sigma_\tau^{-1/2} X
    \Sigma_\tau^{-1/2}}^2_F.
    \label{eq:bound-Htilde-min}
\end{align}
\end{subequations}
\end{lemma}

Those in turn bound the extremal eigenvalues of the Hessian, as defined in
\Cref{def:extremal-eigenvals}.

\begin{lemma}[Bounds on the Hessian $\bG_{\tau}$]
    \label{lem:bound-hessian-G}

    Let $\bG_{\tau}$ be defined as in~\eqref{eq:def-G}.     Then, the extremal
    eigenvalues of $\bG_{\tau}$ are bounded as
    \begin{equation}
        \lambda_{\max}(\bG_{\tau}) \leqslant \frac{\sqrt{C_{\tau}
    \lambda_{\max}(\Sigma_0)}}{2\tau^2}, \qquad
        \lambda_{\min}(\bG_{\tau}) \geqslant
    \frac{\sqrt{\tau \lambda_{\min}(\Sigma_0)}}{2C_{\tau}^2}, 
    \end{equation}
    where $C_{\tau} = 2(L(\Sigma_\tau(0)) + \trc (\Sigma_0))$ is
    initialization-dependent. 
In particular, the loss $L_\tau$ is strongly convex, with parameter 
$$
K_\tau = \frac{\sqrt{\tau\lambda_{\min}(\Sigma_0)}}{2C_{\tau}^2}.
$$
\end{lemma}

\begin{proof}
    We first provide the proof for the maximal eigenvalue.  

    The maximal eigenvalue of the Hessian is defined as 
    \begin{equation*}
        \lambda_{\max}(\bG_{\tau}) = \sup_{X:\norm{X}_F=1} \ang{X, \bG_{\tau}(X)}.
    \end{equation*}
    From the upper-bound of $\ang{X, \bG_{\tau}(X)}$ in~\eqref{eq:bound-Htilde}, one has 
    \begin{align*}
        \sup_{X:\norm{X}_F=1} \ang{X, \bG_{\tau}(X)} 
        &\leqslant \sup_{X:\norm{X}_F=1} \frac{\lambda_{\max}^{1/2}(\Sigma_0^{1/2}
        \Sigma_\tau \Sigma_0^{1/2})}{2} \norm{\Sigma_\tau^{-1/2} X \Sigma_\tau^{-1/2}}^2_F
        \\
        &  =\frac{\lambda_{\max}^{1/2}(\Sigma_0^{1/2} \Sigma_\tau
    \Sigma_0^{1/2})}{2} \sup_{X:\norm{X}_F=1} \norm{\Sigma_\tau^{-1/2} X \Sigma_\tau^{-1/2}}^2_F
      \\
        &= \frac{\lambda_{\max}^{1/2}(\Sigma_0^{1/2} \Sigma_\tau
    \Sigma_0^{1/2})}{2} \sup_{X:\norm{X}_F=1} \norm{\Sigma_\tau^{-1} X}^2_F
    \\
        &= \frac{\lambda_{\max}^{1/2}(\Sigma_0^{1/2} \Sigma_\tau
        \Sigma_0^{1/2})}{2} \lambda_{\max}^2(\Sigma_\tau^{-1})
        \\
        &\leqslant  \frac{\lambda_{\max}^{1/2}(\Sigma_0^{1/2} \Sigma_\tau
        \Sigma_0^{1/2})}{2 \tau^2} . 
    \end{align*}
    The last inequality comes from the definition of $\Sigma_\tau$; if $\lambda_1 \geqslant \lambda_2 \geqslant \cdots \geqslant \lambda_k >0 $ are the
    positive eigenvalues of $WW^\top$, then $\Sigma_\tau^{-1} = {(WW^\top + \tau
    I_n)}^{-1}$ has eigenvalues $\underbrace{\tau^{-1} = \cdots  =\tau^{-1}}_{n-k \text{ times}}
    > {(\lambda_k + \tau)}^{-1} \geqslant \cdots \geqslant {(\lambda_1 +
    \tau)}^{-1}$. 

    For any positive definite matrices $A,B \in \cS_{++}(n)$ with
    \emph{increasing} eigenvalues, and for any $k \in
    [n]$, we know that
    \begin{equation*}
        \lambda_k(A) \lambda_1(B) \leqslant \lambda_k(AB) =
        \lambda_k(A^{1/2} B A^{1/2})
        \leqslant \lambda_k(A)\lambda_n(B).
    \end{equation*}
    Therefore, we have the bound 
$\lambda_{\max}^{1/2}(\Sigma_0^{1/2}\Sigma_\tau \Sigma_0^{1/2}) \leqslant
\lambda_{\max}^{1/2}(\Sigma_0) \lambda_{\max}^{1/2}(\Sigma_\tau)$. Moreover,
$\lambda_{\max}(\Sigma_\tau) \leqslant \trc{\Sigma_\tau}$, and from
\Cref{lem:bound-BW-L}, we know that $\trc{\Sigma_\tau} \leqslant 2(
L(\Sigma_\tau) - L(\Sigma_0)) \eqdef C_{\tau}$. Therefore, we obtain
\begin{equation*}
    \lambda_{\max}(\bG_{\tau}) \leqslant \frac{\sqrt{C_{\tau}
    \lambda_{\max}(\Sigma_0)}}{2\tau^2}.
\end{equation*}
The proof for the minimal eigenvalue is similar and follows from the bound
\eqref{eq:bound-Htilde-min}. In this case, the term
$\lambda_{\min}^{1/2}(\Sigma^{1/2}_0
\Sigma_\tau \Sigma_0^{1/2})$ can be lower bounded by $\sqrt{\tau
\lambda_{\min}(\Sigma_0)}$.
\end{proof}

\begin{remark}
    \label{rem:general-bound-sod}
    In a more generic situation, if $\cQ_{n,\varepsilon} = \cS_{++}(n) \cap
    \set{A \in \cS(n) \mid
    \lambda_{\min}(A) \geqslant \varepsilon}$, then, the original loss $L$ is
    differentiable on $\cQ_{n,\varepsilon}$. The bounds found in
    \Cref{lem:bound-sod} are valid if $\tau$ is replaced with $\varepsilon$.
    Specifically, since $\cQ_{n, \varepsilon}$ is convex, the loss $L$ is
    strongly convex on $\cQ_{n, \varepsilon}$, with strong-convexity constant
    $K_{\varepsilon} = \frac{\sqrt{\varepsilon \lambda_{\min}(\Sigma_0)}}{2C^2}$, where $C
    \defeq 2(L(\Sigma(0)) + \trc \Sigma_0)$.
\end{remark}

The above \Cref{rem:general-bound-sod} leads to stating the following lemma. 

\begin{lemma}
    \label{lem:gen-str-cvx}
    For $n \in \bN \setminus \set{0}$ and $\varepsilon \in \bR_+$, let
    $\cQ_{n,\varepsilon} \defeq
    \cS_{++}(n) \cap \set{ A \in \cS(n) \mid
    \lambda_{\min}(A) \geqslant \varepsilon}$. Then, ($\cQ_{n,\varepsilon}$ is convex and) the loss $L$ is strongly convex on
    $\cQ_{n,\varepsilon}$, with constant $K_{\varepsilon} = \frac{\sqrt{\varepsilon
    \lambda_{\min}(\Sigma_0)}}{2 C^2}$, where $C \defeq  2(L(\Sigma(0)) +
    \trc(\Sigma_0))$.
    \end{lemma}

\begin{proof} 
    $\cQ_{n,\varepsilon}$ is convex as the intersection of convex sets. On $\cQ_{n,\varepsilon}$, the
    Hessian of the loss $L$ has its spectrum lower-bounded as stated in
    \Cref{rem:general-bound-sod}. The proof of \Cref{lem:bound-sod} can
    therefore be adapted with $\varepsilon$ in place of $\tau$.
\end{proof}

The \Cref{lem:gen-str-cvx} will be useful to state the gradient flow convergence
result for the original loss $L$ in \Cref{thm:GF-bis}.

We now turn to the Hessian of $L^1_\tau$, $\bH_{\tau}$. 

\begin{lemma}[Spectral bound of $\bH_{\tau}$]
    \label{lem:bound-H}
    Let $\bH_{\tau}$ be defined as in~\eqref{eq:def-H}. The maximal eigenvalue for the Hessian of $L^1_\tau$ respects the
    following bound
    \begin{equation}
        \label{eq:bound-H}
        \lambda_{\max}(\bH_{\tau}) \leqslant
        \lambda_{\max}^{1/2}(\Sigma_0^{1/2}\Sigma_\tau \Sigma_0^{1/2}) 
        \frac{2C^2}{\tau^2} + 2(1 - \lambda_{\min}(\Sigma_0^{1/2}{(\Sigma_0^{1/2}\Sigma_\tau
    \Sigma_0^{1/2})}^{-1/2}\Sigma_0^{1/2} ))
    \end{equation}
\end{lemma}

\begin{proof}
    From~\eqref{eq:bound-Htilde}, one has for any $X \in \cS_{++}(n)$, 
    \begin{alignat*}{2}
    &&\ang{X, \bG_{\tau}(X)} 
    &\leqslant \frac{\lambda_{\max}^{1/2}(\Sigma_0^{1/2}
        \Sigma_\tau \Sigma_0^{1/2})}{2} \norm{\Sigma_\tau^{-1/2} X \Sigma_\tau^{-1/2}}^2_F.
        \intertext{Let $U \in \bR^{n\times m}$. With $X(U) = UW^\top + WU^\top$, the bound becomes }
    &&\ang{UW^\top + WU^\top, \bG_{\tau}(X(U))}
    &\leqslant \frac{\lambda_{\max}^{1/2}(\Sigma_0^{1/2}
        \Sigma_\tau \Sigma_0^{1/2})}{2} \norm{\Sigma_\tau^{-1/2} X(U) \Sigma_\tau^{-1/2}}^2_F
    \\
    &\iff&
    2\ang{UW^\top,  \bG_{\tau}(X(U))}
    &\leqslant \frac{\lambda_{\max}^{1/2}(\Sigma_0^{1/2}
        \Sigma_\tau \Sigma_0^{1/2})}{2} \norm{\Sigma_\tau^{-1/2} X(U) \Sigma_\tau^{-1/2}}^2_F
    \\
    &\iff&
    2\ang{U,  \bG_{\tau}(X(U))W}
    &\leqslant \frac{\lambda_{\max}^{1/2}(\Sigma_0^{1/2}
        \Sigma_\tau \Sigma_0^{1/2})}{2} \norm{\Sigma_\tau^{-1/2} X(U) \Sigma_\tau^{-1/2}}^2_F . 
    \end{alignat*} 
    Therefore, 
    \begin{alignat}{2}
        && \ang{U, &\bH_{\tau}(U)} 
        = 2\ang{U, \bG_{\tau}(X(U))W +
        (I-\Sigma_0^{1/2}{(\Sigma_0^{1/2}\Sigma_\tau
    \Sigma_0^{1/2})}^{-1/2}\Sigma_0^{1/2})U}
\nonumber    \\
        &&
        &\leqslant  \frac{\lambda_{\max}^{1/2}(\Sigma_0^{1/2}
        \Sigma_\tau \Sigma_0^{1/2})}{2} \norm{\Sigma_\tau^{-1/2} X(U) \Sigma_\tau^{-1/2}}^2_F + 2\ang{U,  (I-\Sigma_0^{1/2}{(\Sigma_0^{1/2}\Sigma_\tau
    \Sigma_0^{1/2})}^{-1/2}\Sigma_0^{1/2})U} .\label{eq:UHU}
    \end{alignat} 
    We proceed by bounding each of the summands.

    First consider the term $\norm{\Sigma_\tau^{-1/2} X(U)
    \Sigma_\tau^{-1/2}}^2_F = \norm{\Sigma_\tau^{-1} X(U)}^2_F$.
 If $U$ is such that $\norm{U}_F = 1$, then $\norm{X(U)}_F^2 =
    \norm{UW^\top + WU^\top}^2 \leqslant 4\norm{W}_F^2$. We know that
    $\norm{W}_F \leqslant C$ for some constant $C$, c.f.~\eqref{eq:bound-W}. Therefore,
    $\norm{U}_F = 1 \implies \norm{X(U)} \leqslant 2C$  and
    \begin{align*} 
        \sup_{U\colon \norm{U}_F=1} \norm{\Sigma_\tau^{-1} X(U) }^2_F 
        &\leqslant \sup_{X \colon \norm{X}_F \leqslant 2C} \norm{\Sigma_\tau^{-1} X(U)}^2_F 
        \\
        &=\sup_{X \colon \norm{X}=1 } 4C^2\norm{\Sigma_\tau^{-1} X}^2_F 
        \\
        & = 4C^2\lambda_{\max}^2(\Sigma_\tau^{-1}) = \frac{4C^2}{\tau^2}.
    \end{align*}
    Therefore, 
    \begin{align*}
        \sup_{U \colon \norm{U}_F=1} \frac{\lambda_{\max}^{1/2}(\Sigma_0^{1/2}
        \Sigma_\tau \Sigma_0^{1/2})}{2} \norm{\Sigma_\tau^{-1/2} X(U) \Sigma_\tau^{-1/2}}^2_F
        & \leqslant  \lambda_{\max}^{1/2}(\Sigma_0^{1/2}
    \Sigma_\tau \Sigma_0^{1/2}) \frac{2C^2}{\tau^2} . 
    \end{align*}
    The second summation in~\eqref{eq:UHU} can be bounded as 
    \begin{equation*}
    \begin{aligned}
        \sup_{U \colon \norm{U}_F=1}2\langle U,  (I-&\Sigma_0^{1/2}{(\Sigma_0^{1/2}\Sigma_\tau
\Sigma_0^{1/2})}^{-1/2}\Sigma_0^{1/2})U\rangle\\ 
    &= 2\lambda_{\max}(I-\Sigma_0^{1/2}{(\Sigma_0^{1/2}\Sigma_\tau
    \Sigma_0^{1/2})}^{-1/2}\Sigma_0^{1/2})
    \\
    &= 2(1-\lambda_{\min}(\Sigma_0^{1/2}{(\Sigma_0^{1/2}\Sigma_\tau
    \Sigma_0^{1/2})}^{-1/2}\Sigma_0^{1/2} )).
    \end{aligned}
\end{equation*}
\end{proof}

\begin{lemma}[Lipshitz-smoothness of $L^1_\tau$]
    For $\tau >0$, the loss $W \mapsto L^1_\tau(W)$ is Lipschitz smooth.
\end{lemma} 

\begin{proof}
    This directly follows from the boundedness of the Hessian showed previously and the convexity of $L_{\tau}^{1}$ using Taylor approximation.
\end{proof}

Once the Lipschitz-smoothness of the loss has been proven, one can turn to showing that the rank is preserved under balanced initial conditions. 

\begin{proposition}[{\citealt[Proposition 4.4]{Bah2021Learning}}]
    \label{prop:trajectory-in-Mk}
    Let ${\cL^1 \colon \bR^{n\times m} \to \bR}$ be a Lipschitz smooth function (i.e., a differentiable function with Lipschitz gradient).
    Suppose that $W_1(t), \ldots, W_N(t)$ are solutions of the gradient
    flow~\eqref{eq:GF} of $L^N$ with balanced initial values $W_j(0)$ and
    define the product $W(t) = \phi(\theta(t))  = W_N(t) \cdots W_1(t)$. If $W(0)$ is contained in
    $\cM(k)$ for some $k \in \bN$, then $W(t)$ is contained in $\cM(k)$ for all $t \geqslant 0$.
\end{proposition}

\begin{proof}
    Let $P(t) = {W_1(t)}^\top W_1(t) = {({W(t)}^\top W(t))}^{1/N}$ and $Q(t) =
    {W_N(t){W_N(t)}^\top} = {(W(t){W(t)}^\top)}^{1/N}$.
    The proof follows \emph{if} the gradient flow is locally Lipschitz continuous in
    $P,Q, W$, so that the curves $P, Q, W$ are uniquely determined by an initial
    datum $P(0), Q(0), W(0)$. From~\Cref{eq:GF,eq:gradient-LN},
    \begin{align*}
        \dot{P} &=  - W^\top \nabla \cL^1(W) - {\nabla \cL^1(W)}^\top W, \\
        \dot{Q} &=  - \nabla \cL^1(W)W^\top  - W{\nabla \cL^1(W)}^\top,  \\
        \dot{W} &= -\sum_{j=1}^N Q^{N-j} \nabla \cL^1(W) P^{j-1} . 
    \end{align*}
    Now, with the assumption of Lipschitz continuity of the flow, a given
    solution is uniquely determined by the initial data, %
    and the proof tools of~\citet[Proposition 4.4]{Bah2021Learning} can be used here as well.
\end{proof}

\begin{remark}
The loss $L^1_\tau$ satisfies the conditions of \Cref{prop:trajectory-in-Mk}; therefore, the flow on $L^1_\tau$ remains in the manifold $\cM(k)$ if $W(t_0) \in \cM(k)$ for some $t_0$.
\end{remark}

\subsection{Proofs of gradient flow convergence}
\label{sec:proofgradflowconv}

We first prove here the convergence of the gradient flow of $L^N$ to a parameter
corresponding to a covariance matrix that is a global minimizer of $L_\tau$,
under the assumptions of balanced weights (\Cref{def:balanced-weights}) and
modified deficiency margin (\Cref{def:MDM}). Then, in \Cref{thm:GF-bis}, we
state the theorem for the original loss, under the same assumptions on the
weights.

\begin{proof}[Proof of \Cref{thm:convergenceGF}]
    The idea of the proof is to transfer the strong convexity property from
    $L_\tau$ to the evolution of the parameters.
    Let us start by the inequality which holds due to strong convexity
    \begin{equation*}
        L(\Sigma_{\tau}) - L(\Sigma_{\tau}^{*}) \leqslant \frac{1}{2K_{\tau}}\|\nabla L(\Sigma_{\tau})\|^{2},
    \end{equation*} 
    where $K_{\tau}$ is the constant from Lemma~(\ref{lem:bound-hessian-G}).
    Rearranging the terms in the above equation, we have 
    \begin{equation}\label{eq:str-conv-2}
        -\|\nabla L(\Sigma_{\tau})\|^{2} \leqslant -2K_{\tau}\left(L(\Sigma_{\tau}) -
        L(\Sigma_{\tau}^{*})\right).
    \end{equation}

    On the covariance space, for the regularized loss, the gradient flow is written
    \begin{align*}
        \od{ L_\tau(\Sigma)}{t} 
        &= \ang{\nabla L_\tau (\Sigma),
        \od{}{t} \Sigma(t)} 
        \\
        &=  \ang{\nabla_\Sigma L_\tau (\Sigma), W \od{W}{t}^\top +
        \od{W}{t} W^\top}
        \\
        &= 2 \ang{\nabla_\Sigma L_\tau(\Sigma) W, \od{W}{t}} . 
    \end{align*}
    The expression of $\od{W}{t}$ is given
    in~\Cref{lem:flow-prop}.\ref{lem:flow-w}:
    \begin{equation*}
        \od{W}{t} = - \sum_{\ell=1}^N W_{N:j+1} W_{N:j+1}^\top \nabla L^1(W)
        W_{\ell-1:1}^\top W_{\ell-1:1}.
    \end{equation*}

    Since $\nabla L^1(W) = 2 \nabla L(\Sigma)W$, and from the balancedness
    assumption we have $W_{N:j+1}W_{N:j+1}^\top =
    {(WW^\top)}^{\frac{N-\ell}{N}}$ and $W_{\ell-1:1}^\top W_{\ell-1:1} =
    {(W^\top W)}^{\frac{\ell-1}{N}}$, we get 
    \begin{align*}
        \od{L_\tau(\Sigma(t))}{t} = - 4 \sum_{\ell=1}^N \ang{ \nabla
            L_\tau(\Sigma) W, {(WW^\top)}^{\frac{N-\ell}{N}} \nabla L_\tau(\Sigma) W
        {(W^\top W)}^{\frac{\ell-1}{N}}} .
    \end{align*}
    Now, let $USV^\top = W$ be a (thin) SVD of $W$, so that $WW^\top = U S^2
    U^\top$ and $W^\top W = V S^2 V^\top$.     For one layer $\ell \in [N]$, we then have 
    \begin{align*}
        \ang{\nabla L_\tau(\Sigma) W, {(WW^\top)}^{\frac{N-\ell}{N}} \nabla
        L_\tau(\Sigma) W {(W^\top W)}^{\frac{\ell-1}{N}}}
            &= \trc{(\nabla L_\tau(\Sigma) W{(W^\top W)}^{\frac{\ell-1}{N}}
            W^\top \nabla L_\tau(\Sigma)  {(WW^\top)}^{\frac{N-\ell}{N}})}
            \\
            &= \trc{(\nabla L_\tau(\Sigma) USV^\top
                V{S}^{\frac{2(\ell-1)}{N}}V^\top
            V S U^\top \nabla L_\tau(\Sigma)  U{S}^{\frac{2(N-\ell)}{N}}U^\top)}
            \\
            &= \trc{(U^\top \nabla L_\tau(\Sigma) U{S}^{\frac{2(N+\ell-1)}{N}}
            U^\top \nabla L_\tau(\Sigma)  U{S}^{\frac{2(N-\ell)}{N}})}
            \\
            &= \ang{U^\top \nabla L_\tau(\Sigma) U{S}^{\frac{2(N+\ell-1)}{N}}, 
            {S}^{\frac{2(N-\ell)}{N}}U^\top \nabla L_\tau(\Sigma)  U} . 
    \end{align*}
    Let $X \defeq U^\top \nabla L_\tau(\Sigma) U,\, D
    \defeq{S}^{\frac{2(N+\ell-1)}{N}}$, and ${E} \defeq{S}^{\frac{2(N-\ell)}{N}}
    $. We evaluate $\ang{XD, {E}X}$ for the diagonal $D, {E}$ as 
    \begin{align*}
        \ang{X{D}, {E}X} = \sum_{i, j} X_{i,j} {D}_j X_{i,j} {E}_i = \sum_{i, j} {E}_i {D}_j
        X_{i,\ell}^2 . 
    \end{align*}
    Since ${E}_i = s_i^{\frac{2(N-\ell)}{N}}$ and ${D}_j = s_j^{\frac{2(N+\ell -
    1)}{N}}$, and due to the 
    modified margin deficiency assumption, for all
    $(i,j) \in {[k]}^2$, we have ${E}_i
    \geqslant c^{\frac{2(N-\ell)}{N}}$ and ${D}_j \geqslant c^{\frac{2(N+\ell -
    1)}{N}}$, so that 
    \begin{align}
        \label{eq:MDM-bound}
        \ang{X{D}, {E}X} \geqslant c^{\frac{2(2N - 1)}{N}}\sum_{i,j} X_{i,j}^2 =
        c^{\frac{2(2N-1)}{N}} \norm{X}_F^2 .
    \end{align}
    Since $X = U^\top \nabla L_\tau(\Sigma) U^\top$, we have that $\norm{X}_F^2 =
    \norm{\nabla L_\tau(\Sigma)}^2_F$, so that in total
    \begin{equation*}
        \od{}{t}L_\tau(\Sigma(t)) \leqslant - 4 \sum_{\ell=1}^{N} c^{\frac{2(2N-1)}{N}} \norm{\nabla
        L_\tau(\Sigma)}^2_F = -4 N c^{\frac{2(2N - 1)}{N}} \norm{\nabla 
        L_\tau(\Sigma)}_F^2.
    \end{equation*}
    From the strong convexity of $L_\tau$ \eqref{eq:str-conv-2}, we get the bound 
    \begin{alignat*}{2}
    &&\od{}{t}L_\tau(\Sigma(t)) 
    &\leqslant -8 N c^{\frac{2(2N - 1)}{N}} K_{\tau} (L_\tau(\Sigma) - L_\tau(\Sigma^*))
    \\
    & \implies \quad & \frac{1}{L_\tau(\Sigma(t)) - L_\tau(\Sigma^*)} \od{}{t}(L_\tau(\Sigma(t)) -
    L_\tau(\Sigma^*)) & \leqslant -8 N c^{\frac{2(2N -1)}{N}} K_{\tau} . 
\end{alignat*}

Now, by integrating both sides from $0$ to $t$, 
\begin{equation*}\label{eq:int-GF}
    \ln\left(\frac{L(\Sigma_{\tau}(t)) -
            L(\Sigma_{\tau}^{*})}{L(\Sigma_{\tau}(0)) -
    L(\Sigma_{\tau}^{*})}\right)\leqslant -8Nc^{\frac{2(2N-1)}{N}}K_{\tau}t.
\end{equation*}
Let $\Delta_{\tau}^{*} = \Sigma_{\tau}(0) - \Sigma_{\tau}^{*}$ be the distance
from the initialization to optimality. Finally we get the desired exponential rate
\begin{equation*}
    L(\Sigma_{\tau}(t)) - L(\Sigma_{\tau}^{*}) \leqslant
    e^{-8N c^{\frac{2(2N -1)}{N}}K_{\tau}t}\Delta_{\tau}^{*},
\end{equation*}
which concludes the proof.
\end{proof}

\begin{remark}
    \label{rem:remove-reg}
    The modified deficiency margin assumption \Cref{def:MDM} is used only in order to lower-bound the singular values of the parametrized covariance matrix $WW^\top$ in~\eqref{eq:MDM-bound}. 
    Furthermore, under the MDM assumption, the parametrized matrix is always full-rank and, therefore, we do not need to regularize the loss in order to define the gradient flow and to prove convergence of the same. 
\end{remark}

\Cref{rem:remove-reg} suggests that we can adapt the statement of
\Cref{thm:convergenceGF} in two ways. 
Namely, we could substitute the MDM assumption with the weaker condition~\eqref{eq:smin-bound}, or we could keep the MDM assumption and substitute the regularized loss by the unregularized loss. In the following we briefly discuss the arguments for the latter of these two options. 

For $\varepsilon \geqslant 0$, let $\cQ(n, \varepsilon) \defeq  \cS_{++}(n) \cap \set{ A \in \cS(n) \mid
    \lambda_{\min}(A) \geqslant \varepsilon}$.

    From \Cref{lem:gen-str-cvx}, we know that the loss $L$ is strongly-convex on the set $\cQ_{n \varepsilon}$, for
$\varepsilon > 0$, and the convergence of the gradient flow will therefore be
linear on this set. Under the modified margin deficiency assumption, we know
that the parametrized covariance matrix remains in the set $\cQ_{n, c^2}$, as
stated in the next lemma.

\begin{corollary}[from \Cref{lem:marginc}]
    \label{lem:ww'-pd}
    If $W$ satisfies the modified deficiency margin assumption at some time
    $t$, then, $WW^\top \in \cQ_{n,c^2}$ for all time.
\end{corollary}

Therefore, the modified deficiency margin assumption allows to conclude on the
linear convergence of the gradient flow for the original loss $L$.

\begin{theorem}
    \label{thm:GF-bis}
    Assume both balancedness (\Cref{def:balanced-weights}) and the
    modified deficiency margin
    (\Cref{def:MDM}) conditions hold. 
    Then the gradient flow $\dot \W(t) = - \nabla
    L^N(\W(t))$  converges as 
    \begin{equation}\label{GFconvpertbis}
        L(\Sigma(t)) -L(\Sigma^{*})\leqslant
        e^{-8N c^{\frac{2(2N-1)}{N}}Kt}\Delta_{0}^{*},
    \end{equation}
    where $K = \frac{\sqrt{c^2 \lambda_{\min}(\Sigma_0)}}{2 C^2}$ is the
    strong convexity parameter from \Cref{lem:strong-cvx}, with $C =
    2(L(\Sigma(0)) + \trc(\Sigma_0))$, and $\Delta_{0}^{*} = \Sigma(0) -
    \Sigma^{*}$ is the distance from the optimum as initialization.
\end{theorem}

\begin{proof}
    Under the modified margin deficiency assumption, by \Cref{lem:ww'-pd}, the
    model covariance $\Sigma(t) = W(t)W(t)^\top$ has its eigenvalues lower-bounded by
    $c^2$
    at all time $t \geqslant 0$. Therefore, the proof of \Cref{thm:convergenceGF} can be
    adapted, with $c^2$ in place of~$\tau$. 
\end{proof}

\subsection{Proof of gradient descent convergence}
\label{app:GD-conv}
We start by proving Lemma~\ref{lem:marginc} so that with the 
modified margin deficiency assumption on the initial weights, $WW^{\top}$ does not degenerate along the gradient descent training algorithms. 
\begin{proof}[Proof of \Cref{lem:marginc}]
Let $\bar U(k):=\argmin_{U\in \cO(n)} \norm{\sqrt{W(k) W(k)^{\top}} - {\Sigma_0^{1/2}}U}_F^2 $ for each $k$, then as $L^1(W(k))\leqslant L^1(W(0))$ for all $k\geqslant 0$, we have
    \begin{equation}
        \begin{aligned}
      \sigma_{\min}\Big(\sqrt{W(k) W(k)^{\top}}\Big)&= \sigma_{\min}\Big(\sqrt{W(k) W(k)^{\top}}-\Sigma_0^{1/2}\bar U(k)+\Sigma_0^{1/2}\bar U(k)\Big)   \\
      &\geqslant \sigma_{\min}\Big(\Sigma_0^{1/2}\bar U(k)\Big)-\sigma_{\max}\Big(\sqrt{W(k) W(k)^{\top}}-\Sigma_0^{1/2}\bar U(k)\Big)\\
      &\geqslant \sigma_{\min}\Big(\Sigma_0^{1/2}\bar U(k)\Big)-\norm{\sqrt{W(k) W(k)^{\top}} - {\Sigma_0^{1/2}}\bar U(k)}_F\\
      &= \sigma_{\min}\Big(\Sigma_0^{1/2}\bar U(k)\Big)-\sqrt{L^1(W(k))}\\
      &\geqslant \sigma_{\min}\Big(\Sigma_0^{1/2}\bar U(k)\Big)-\sqrt{L^1(W(0))}\\
      & =\sigma_{\min}\Big(\Sigma_0^{1/2}\bar U(k)\Big) - \norm{\sqrt{W(0) W(0)^{\top}} - {\Sigma_0^{1/2}}\bar U(0)}_F\\
      &\geqslant \sigma_{\min}\Big(\Sigma_0^{1/2}\bar U(k)\Big) - \sigma_{\min}\Big(\Sigma_0^{1/2}\Big) +c = c.
        \end{aligned}
    \end{equation}
    The cancellation in the last equality works due to the fact that the multiplication with an arbitrary unitary matrix does not change singular values.
\end{proof}
Now we are ready to prove the finite step size gradient descent convergence of the BW loss.
We consider the perfect balancedness of initial values $W_i(0), 1\leqslant i\leqslant N$ in the remaining proof. The approximation balancedness case can also be carried out but require more complicated auxiliary estimates. We leave the approximate balancedness assumption as a future direction.
\begin{proof}[Proof of \Cref{thm:gradientdescent}]
Let us start from the gradient descent of the loss with respect to each layer
\begin{equation}\label{jth_gd}
\begin{aligned}
    W_j(k+1) &= W_j(k) -\eta \nabla_{W_j} L^N(W_1(k),\cdots W_n(k))\\
    &=W_j(k) - \eta W_{j+1:N}(k)^\top \nabla_W L^1(W(k)) W_{1:j-1}(k)^{\top}, \quad 1\leqslant j \leqslant N,
\end{aligned}
\end{equation}
with the boundary conditions $W_{1:0}(k)= I_{d_0}$ and $W_{N+1:N}(k) = I_{d_N}$ for all $k\geqslant 0$.

With the notations $\overrightarrow{W} = (W_1, W_2, \cdots,W_N)$ and 
$$
\nabla L^N(\overrightarrow{W}) = \begin{pmatrix}
\nabla_{W_1}L^N(\overrightarrow{W}) \\
\vdots\\
\nabla_{W_N}L^N(\overrightarrow{W})
\end{pmatrix} , 
$$
we consider to write the Taylor expansion in the form
\begin{equation}\label{taylorexpansion}
    \begin{aligned}
L^N(\overrightarrow{W}(k+1))& = L^N(\overrightarrow{W}(k))+\Big\langle \nabla L^N(\overrightarrow{W}(k)),  \overrightarrow{W}(k+1)-\overrightarrow{W}(k)\Big\rangle \\
&+ \frac{1}{2}\Big\langle (\overrightarrow{W}(k+1)-\overrightarrow{W}(k))^{\top} \nabla^2 L^N(\overrightarrow{A_{\xi}}(k)), \overrightarrow{W}(k+1)-\overrightarrow{W}(k)\Big\rangle,
\end{aligned}
\end{equation}
with 
\begin{equation*}
    \overrightarrow{A_{\xi}}(k) = \overrightarrow{W}(k) + \xi(\overrightarrow{W}(k+1)-\overrightarrow{W}(k)), \quad \text{for some }~\xi\in[0,1].
\end{equation*} 
Recall the relation (\ref{eq:gradient-LN}), for $1\leqslant j\leqslant N$,
    \begin{equation*}
        \nabla_{W_j} L^N(W_1, \ldots, W_N) = W_{j+1}^\top \cdots W_N^\top
        \nabla_W L^1(W) W_1^\top \cdots W_{j-1}^\top,
    \end{equation*}
then the first order term in (\ref{taylorexpansion}), under (\ref{jth_gd}), can be written as
\begin{equation}
    \begin{aligned}
 \Big\langle \nabla &L^N(\overrightarrow{W}(k)),  \overrightarrow{W}(k+1)-\overrightarrow{W}(k)\Big\rangle=       \sum_{j=1}^N
    \nabla_{W_j} L^N(\overrightarrow{W}(k))^{\top}(W_j(k+1)-W_j(k))\\
    &=-\eta \sum_{j=1}^N W_{j-1}\cdots W_1\nabla_W L^1(W(k))^{\top}W_N\cdots W_{j+1}W_{j+1}^\top \cdots W_N^\top
        \nabla_W L^1(W(k)) W_1^\top \cdots W_{j-1}^\top\\
        &= -\eta \sum_{j=1}^NW_{j-1}\cdots W_1\nabla_W L^1(W(k))^{\top} (W_N W_N^\top)^{N-j}\nabla_W L^1(W(k)) W_1^\top \cdots W_{j-1}^\top\\
        &\leqslant -\eta \sum_{j=1}^N \sigma_{\min}\Big((W_N W_N^\top)^{N-j}\Big)\sigma_{\min}\Big((W_1^\top W_1)^{j-1}\Big)\norm{\nabla_W L^1(W(k))}_F^2 . 
    \end{aligned}
\end{equation}
Throughout the computation above, $W_i = W_i(k)$ for all $1\leqslant i\leqslant N$. Moreover, we use the balancedness $W_j W_j^{\top} = W_{j+1}^{\top} W_{j+1}$ for all $1\leqslant i\leqslant N-1$ so that, in the symmetric structure above, 
\begin{align*}
    & W_N\cdots W_{j+1}W_{j+1}^\top \cdots W_N^\top = (W_N W_N^\top)^{N-j}\\
    & W_1^{\top}\cdots W_{j-1}^\top W_{j-1}\cdots W_1  = (W_1^\top W_1)^{j-1}.
\end{align*}
Therefore, thanks to Lemma \ref{lem:marginc}, 
\begin{align*}
    \sigma_{\min}\Big((W_N(k) W_N(k)^{\top})^N\Big)=\sigma_{\min}\Big((W_1(k)^{\top} W_1(k))^N\Big)=\sigma_{\min}\Big(W(k)W(k)^{\top}\Big)\geqslant c^2,
\end{align*}
from which we get
\begin{equation}
 \Big\langle \nabla L^N(\overrightarrow{W}(k)), \overrightarrow{W}(k+1)-\overrightarrow{W}(k)\Big\rangle \leqslant -\eta N c^{\frac{2(N-1)}{N}} \norm{\nabla_W L^1(W(k))}_F^2.  
\end{equation}
Let us mention that \citet[Theorem 1 and Claim 1]{Arora2018Optimization} provide rigorous derivations about the equalities above. The second order term in (\ref{taylorexpansion}) is more complicated to handle, as we have
\begin{equation}
    \nabla^2 L^N(\overrightarrow{W})[\overrightarrow{X},\overrightarrow{X}] =
    \sum_{j=1}^N \Big\langle X_j, \frac{\dif{}^2 L^N(\overrightarrow{W})}{\dif
    W_j^2}X_j\Big\rangle + \sum_{j=1}^N\sum_{i=1,i\neq j}^N \Big\langle X_j,
\frac{\dif{}^2 L^N(\overrightarrow{W})}{\dif W_i\dif W_j}X_i\Big\rangle.
\end{equation}
Thanks to \Cref{cor:hessian-loss}, we have expressions of $\frac{\dif{}^2 L^N(\overrightarrow{W})}{\dif W_j^2}$ and $\frac{\dif{}^2 L^N(\overrightarrow{W})}{\dif W_i\dif W_j}$ ready.

Note that we have the boundedness (\Cref{eq:W_lower})
\begin{equation}\label{upper_W}
    \norm{W}_F\leqslant \sqrt{2{(L^1(W) + \norm{\Sigma_0^{1/2}}^2_F)}}  \leqslant   \sqrt{2{(L^1(W(0)) + \norm{\Sigma_0^{1/2}}^2_F)}} =: M,
\end{equation}
and it is straightforward to see that
\begin{equation}\label{upper_Wi}
    \norm{W_i}_F^2\leqslant \norm{W}_F^{2/N}, \quad \text{ for all }~1\leqslant i\leqslant N.
\end{equation}
Moreover, for all $1\leqslant i\leqslant N$, since $\xi\in[0,1]$, 
\begin{equation*}
    A_{\xi, i}(k) = W_i(k) + \xi(W_i(k+1)-W_i(k)) = (1-\xi) W_i(k) + \xi  W_i(k+1),
\end{equation*}
we then have the uniform upper bound for all $k\geqslant 0$,
\begin{equation}\label{bound_Ai}
\norm{A_{\xi, i}(k)}_F\leqslant (1-\xi) \norm{W_i(k)}_F + \xi \norm{W_i(k+1)}_F \leqslant M^{1/N}.
\end{equation}
Using $A_{\xi, i}(k)=W_i(k) -\xi \eta W_{j+1:N}(k)^\top \nabla_W L^1(W(k)) W_{1:j-1}(k)^{\top}$, we can obtain a lower bound in terms of the minimum singular value,
\begin{equation}
\begin{aligned}
    \sigma_{\min}&\left(A_{\xi, i}(k)A_{\xi, i}(k)^{\top}\right) \\&\geqslant \sigma_{\min}\left(W_i(k)W_i(k)^{\top}\right)-2\xi \eta \norm{W_i(k)}_F\norm{W_{j+1:N}(k)}_F\norm{W_{1:j-1}(k)}_F\norm{\nabla_W L^1(W(k))}_F\\
    &\geqslant c^2 - 4\eta M\sqrt{L^1(W(k))} \geqslant c^2 - 4\eta M\sqrt{L^1(W(0))},
\end{aligned}    
\end{equation}
where we utilize (\ref{DW_equality}), (\ref{upper_Wi}) and (\ref{upper_W}), as well as non-increment of $L^1(W)$ throughout the training.
We denote $X_j =- \eta W_{j+1:N}(k)^\top \nabla_W L^1(W(k)) W_{1:j-1}(k)^{\top}$. 
We may choose 
\begin{equation*}
    \eta\leqslant \frac{c^2}{8M\sqrt{L^1(W(0))}},
\end{equation*}
so that for all $k\geqslant 0$,
\begin{equation}\label{sigma_A}
  \sigma_{\min}\left(A_{\xi, i}(k)A_{\xi, i}(k)^{\top}\right)\geqslant \frac{c^2}{2},\quad ~\text{and}~\quad  \sigma_{\min}\left(A_{\xi}(k)A_{\xi}(k)^{\top}\right)\geqslant \frac{c^{2N}}{2^N}.
\end{equation}
Then combining all estimates above, we have
\begin{equation*}
    \begin{aligned}
        \Big| \Big\langle (\overrightarrow{W}(k+1)&-\overrightarrow{W}(k))^{\top} \nabla^2 L^N (\overrightarrow{A_{\xi}}(k)), \overrightarrow{W}(k+1)-\overrightarrow{W}(k)\Big\rangle \Big| \\
        & \leqslant \sum_{j=1}^N \Big|\Big\langle X_j, \frac{\dif{}^2 L^N(\overrightarrow{A_{\xi}}(k))}{\dif W_j^2}X_j\Big\rangle \Big|+ \sum_{j=1}^N\sum_{i=1,i\neq j}^N \Big|\Big\langle X_j, \frac{\dif{}^2 L^N(\overrightarrow{A_{\xi}}(k))}{\dif W_i\dif W_j}X_i\Big\rangle \Big|\\
        &\leqslant\sum_{j=1}^N \frac{\lambda_{\max}^{1/2}(\Sigma_0^{1/2}
        A_{\xi}(k) A_{\xi}(k)^{\top}
    \Sigma_0^{1/2})}{2} \norm{ X_j \big( A_{\xi}(k) A_{\xi}(k)^{\top}\big)^{-1}}^2_F M^{2(N-1)/N}\\
    &~~~+\sum_{j=1}^N\sum_{i=1,i\neq j}^N M^{(N-2)/N}\norm{X_i}_F\norm{X_j}_F\norm{\nabla_W L^1(A_{\xi}(k))}_F \\
    &~~~+\sum_{j=1}^N\sum_{i=1,i\neq j}^N \Big(\frac{\lambda_{\max}^{1/2}(\Sigma_0^{1/2}
        A_{\xi}(k) A_{\xi}(k)^{\top}
    \Sigma_0^{1/2})}{2} \norm{ X_j \big( A_{\xi}(k) A_{\xi}(k)^{\top}\big)^{-1}}_F \\
    &~~~~~~~~~~~~~~~~~~~~~~~~~~~ \times\norm{ X_i \big( A_{\xi}(k) A_{\xi}(k)^{\top}\big)^{-1}}_F M^{2(N-1)/N}\Big),
    \end{aligned}
\end{equation*}
by using (\ref{bound_Ai}), (\ref{lem:bound-sod}) and applying the Cauchy-Schwarz inequality for the last term. Notice that $\norm{X_i}_F\leqslant 2\eta M^{(N-1)/N}\norm{\nabla_W L^1(W(k))}_F$. Now combining all the bounds we obtained previously, in addition to (\ref{sigma_A}), we get that
\begin{equation}
\begin{aligned}
 \Big| \Big\langle (\overrightarrow{W}(k+1)&-\overrightarrow{W}(k))^{\top} \nabla^2 L^N (\overrightarrow{A_{\xi}}(k)), \overrightarrow{W}(k+1)-\overrightarrow{W}(k)\Big\rangle \Big|\\
 &\leqslant  2\eta^2 N^2\norm{A_{\xi}(k)}_F \lambda^{1/2}_{\max}(\Sigma_0)\frac{M^{4(N-1)/N}}{\sigma_{\min}\left(A_{\xi}(k)A_{\xi}(k)^{\top}\right)}\norm{\nabla_W L^1(W(k))}_F^2\\
 &~~~+4\eta^2 N(N-1) M^{(3N-4)/N}\norm{\nabla_W L^1(A_{\xi}(k))}_F\norm{\nabla_W L^1(W(k))}_F^2.
 \end{aligned}
\end{equation}
Moreover, we can use (\ref{eq:variational-BW}), (\ref{DW_equality}) again to get
\begin{equation*}
\begin{aligned}
  \norm{\nabla_W L^1(A_{\xi}(k))}_F &= 2\sqrt{L^1(A_{\xi}(k))}  \leqslant 2\norm{\big(A_{\xi}(k) A_{\xi}(k)^{\top}\big)^{1/2}-\Sigma_0^{1/2}U}_F\\
  &\leqslant 2\norm{\big(A_{\xi}(k) A_{\xi}(k)^{\top}\big)^{1/2}}_F+2\norm{\Sigma_0^{1/2}}_F\leqslant 2M^{1/N}+2\norm{\Sigma_0^{1/2}}_F.
  \end{aligned}
\end{equation*}
Thus, we conclude the estimate for the second order term by
\begin{equation*}
\begin{aligned}
    \Big| \Big\langle (\overrightarrow{W}(k+1)&-\overrightarrow{W}(k))^{\top} \nabla^2 L^N (\overrightarrow{A_{\xi}}(k)), \overrightarrow{W}(k+1)-\overrightarrow{W}(k)\Big\rangle \Big|\\
    &\leqslant \eta^2\norm{\nabla_W L^1(W(k))}_F^2\Bigg(\frac{2^{N+1}}{c^{2N}}N^2M^{(4N-3)/N}\lambda^{1/2}_{\max}(\Sigma_0)\\
    &~~~~~~~~~~~~~~~~~~~+8N(N-1)M^{(3N-4)/N}\Big(M^{1/N}+\norm{\Sigma_0^{1/2}}_F\Big)\Bigg).
    \end{aligned}
\end{equation*}
Let us denote the constant 
\begin{equation*}
    \Delta : = \frac{2^{N+1}}{c^{2N}}N^2M^{(4N-3)/N}\lambda^{1/2}_{\max}(\Sigma_0)+8N(N-1)M^{(3N-4)/N}\Big(M^{1/N}+\norm{\Sigma_0^{1/2}}_F\Big),
\end{equation*}
then, we can write the iteration as 
\begin{equation*}
  L^N(\overrightarrow{W}(k+1)) =\left(1- 4N c^{\frac{2(N-1)}{N}}\eta+4\Delta \eta^2\right) L^N(\overrightarrow{W}(k)).
\end{equation*}
If we choose
\begin{equation*}
    \eta \leqslant \frac{N c^{\frac{2(N-1)}{N}}}{2\Delta},
\end{equation*}
then we have 
\begin{equation*}
    L^N(\overrightarrow{W}(k)) \leqslant \Big(1-2\eta Nc^{\frac{2(N-1)}{N}}\Big)^k L^N(\overrightarrow{W}(0)).
\end{equation*}
For $\eta$ being sufficiently small, we have $1-2\eta Nc^{\frac{2(N-1)}{N}}\leqslant \exp\left(-2\eta Nc^{\frac{2(N-1)}{N}}\right)$. Thus, to achieve $\epsilon$-error for the loss, 
\begin{equation*}
    k\geqslant \frac{1}{2\eta  Nc^{\frac{2(N-1)}{N}}}\log\left(\frac{
    L^1(W(0))}{\epsilon}\right).
\end{equation*}
\end{proof}

\section{Empirical evaluation of the Hessian} 
\label{app:hessian-losses} 

In order to compare the smooth Bures-Wasserstein loss and the Frobenius loss, here we conduct experiments evaluating the Hessian of both. We first discuss the Burer-Monteiro parametrization and relate the differential of a given loss in function and covariance space.  

\subsection{General computations for losses under the Burer-Monteiro
parametrization}
    Let $\pi\colon \bR^{n \times m} \to \cS_+(n),\, W \mapsto \pi(W) \defeq
    WW^\top$ be the so-called Burer-Monteiro parametrization of a positive semi-definite
    matrix. We will consider computing second-order derivatives of a
    differentiable function $f \colon \cS_+(n) \to \bR$ under the
    parametrization $ f \circ \pi$. 

\begin{proposition}[Second-order differential chain rule for the Burer-Monteiro parametrization]
    \label{prop:sod-BMparam}
    Let $\pi\colon \bR^{n \times m} \to \cS_+(n),\, W \mapsto \pi(W) \defeq WW^\top$ be the Burer-Monteiro parametrization of a positive semi-definite matrix. Then, for any twice-differentiable function $f \colon \cS_+(n) \to \bR$, the second-order differential of $f \circ \pi$ can be expressed as , with $W \in \bR^{n\times m}, Z \in \bR^{n \times m}$, and $\Sigma = \pi(W)$: 
    \begin{align}
        \label{eq:sod-BMparam}
        \dif{}^2 (f \circ \pi)(W)[Z] = \dif{}^2 f(\Sigma)[Z W^\top] +\dif{}^2 f(\Sigma)[W Z^\top] +  2 \dif{}^2 f(\Sigma)[WZ^\top,  ZW^\top] + 2 \dif{} f(\Sigma)[Z Z^\top].
    \end{align}
\end{proposition}

\begin{proof}
        The chain rule for the second order differential in \cref{lem:sod-chainrule} states that
    \begin{align*}
        \dif{}^2 (f \circ \pi)(W)[Z]  
        &= 
        \dif{}^2 f(\pi(W))[\dif \pi(W)[Z]] + \dif{} f (\pi(W))[\dif{}^2 \pi(W)[Z]] .\\
        \intertext{Since $\dif{} \pi(W)[Z] = Z W^\top + W Z^\top$ and $\dif{}^2 \pi(W)[Z] = 2  Z Z^\top$, and with $\Sigma = \pi(W)$, one further has}
        \dif{}^2 (f \circ \pi)(W)[Z]  
        &=
        \dif{}^2 f (\Sigma)[Z W^\top +W Z^\top] + \dif{} f(\Sigma)[2 Z Z ^\top]
        \\
        &= \dif{}^2 f (\Sigma)[Z W^\top] +\dif{}^2 f(\Sigma)[W Z^\top] +  2 \dif{}^2 f(\Sigma)[ZW^\top + WZ^\top] + 2 \dif{} f(\Sigma)[Z Z ^\top],
    \end{align*}
where the last inequality comes from the bilinearity of $\dif{}^2 f(\Sigma)$ and the linearity of $\dif{} f(\Sigma)$.
\end{proof}

Now, we turn to the expression of the Hessian matrix for the function $f \circ
\pi$ at some point $W \in \bR^{n\times m}$. Recall that, by definition, this is
the only symmetric matrix of size $nm \times nm$, denoted by $\nabla^2 (f \circ
\pi)(W)$, such that, for all $Z \in \bR^{n \times m}$, $\dif{}^2 (f \circ
\pi)(W)[Z] = {(\vc Z)} ^\top  [\nabla^2 (f \circ \pi) (W)] \vc Z$.

\begin{corollary}
    \label{cor:hess-BMparam}
    With the same notations as in \cref{prop:sod-BMparam}, the Hessian of the loss can then be identified as 
    \begin{equation}
        \label{eq:hess-BMparam}
        \begin{split}
        \nabla^2 (f\circ \pi)(W) &= ((W^\top \otimes I_n) K_{n} + W^\top \otimes
    I_n) \nabla^2 f(\Sigma) ( K_{n} (W \otimes I_n) + W \otimes I_n) + 2 I_m \otimes \nabla f(\Sigma) .    
    \end{split}
    \end{equation}
\end{corollary}

\begin{proof}
    In order to prove the relation~\eqref{eq:hess-BMparam}, we will rely on the
    identity $\vc ABC = (C^\top \otimes A) \vc(B)$ which holds for any matrices
    of compatible shapes. 
    For more details about the matrix computations, we refer to~\citet{Magnus2019Matrix}. Using the expression of the
    second-order differential~\eqref{eq:sod-BMparam}, it implies that, for $W, Z \in
    \bR^{n \times m}$, one has 
    \begin{align*}
        \dif{}^2 f(\Sigma) [Z W ^\top] &= {(\vc ZW^\top )}^\top \nabla ^2 f(\Sigma)
        \vc ZW^\top = {(W \otimes I_n \vc Z)}^\top \nabla^2 f (\Sigma) (W \otimes
        I_n) \vc Z \\
                                       &= {(\vc Z)}^\top (W^\top \otimes I_n) \nabla ^2 f(\Sigma) (W
        \otimes I_n) \vc Z, \\
        \dif{}^2 f(\Sigma) [W Z^\top] &= {(\vc {(ZW^\top)}^\top)}^\top \nabla
        ^2 f(\Sigma) \vc({(ZW^\top)}^\top)  = {(K_{n} \vc ZW^\top)}^\top \nabla
        ^2 f(\Sigma) (K_{n} \vc ZW^\top) \\
                                      &={( \vc Z)}^\top (W^\top \otimes I_n) K_n
                                     \nabla^2 f(\Sigma) K_n (W \otimes I_n) \vc
                                     Z,
                                     \intertext{where $K_n \in \bR^{n^2 \times
                                         n^2}$ is such that $K_n
                                     \vc X^\top = \vc X$ for $X \in \bR^{n
                                 \times n}$, and}
        \dif{}^2 f(\Sigma)[WZ^\top, ZW^\top] &={(\vc Z )}^\top (W^\top \otimes
        I_n) K_n \nabla^2 f(\Sigma) (W \otimes I_n) \vc Z \\
                                             & = {(\vc Z)}^\top
        (W^\top \otimes I_n) \nabla^2 f(\Sigma) K_n (W \otimes I_n) \vc Z.
    \end{align*}
    Moreover, 
    \begin{align*}
        \dif f(\Sigma)[Z Z^\top] = \ang{\nabla f(\Sigma), Z Z^\top} = \trc
            \nabla f(\Sigma) Z Z^\top = {(\vc Z)}^\top (I_m \otimes \nabla
            f(\Sigma)) \vc Z,
        \end{align*}
        where we have used the identity $\trc ABCD = {(\vc D^\top)}^\top (C^\top
        \otimes A) \vc B$ for the last equality.
        Adding the different terms according to~\eqref{eq:sod-BMparam}, and
        factorizing, we get the desired expression

        \begin{equation*}
            \nabla ^2 (f \circ \pi)(W)[Z] = ( W^\top \otimes I_n +  (W^\top
            \otimes I_n) K_n ) \nabla ^2 f(\Sigma) ( W \otimes I_n + K_n(W \otimes
            I_n) ) + 2 (I_m \otimes \nabla f(\Sigma)).
        \end{equation*}

\end{proof}
In order to compute $\nabla^2 (f\circ \pi)(\Sigma)$, one therefore only needs to
know $\nabla^2 f(\Sigma)$ and $\nabla f (\Sigma)$.

\subsection{Frobenius loss}
\label{app:hess-Fro}

Define
\begin{equation*}
    L_F(\Sigma) = \frac12 \norm{\Sigma - \Sigma_0}^2_F,
\end{equation*}
and denote $L^1_F$ the Frobenius loss defined on $W \in \bR^{n \times m}$, and
$L^N_F$ the Frobenius norm defined on the parameters $\theta \in \Theta$. 

Since $L^1_F$ admits a Burer-Monteiro factorization, we only need to recall the
gradient and Hessian matrix of $L_F$ in order to compute their counterparts on
the function space.

\begin{lemma}
    The first-order differential of $L_F$ at $\Sigma \in \cS(n)_+$ is $\dif{} L_F(\Sigma)[X] = \ang{\Sigma -
    \Sigma_0, X}$, and its gradient is $\nabla L_F(\Sigma) = \Sigma - \Sigma_0$. 
\end{lemma}

\begin{lemma}
    The second-order differential of $L_F$ at $\Sigma$ in the direction $X$ is 
    \begin{equation*}
        \dif{}^2 L_F(\Sigma)[X] = \trc{X X^\top},
    \end{equation*}
    and its Hessian matrix is $\nabla^2 L_F(\Sigma) = I_n \otimes I_n$. 
\end{lemma}

We can then give the Hessian matrix of the loss $L^1_F$. 
\begin{corollary}
    \label{cor:hessian-Fro-Func}
    The Hessian matrix  of $L^1_F$ at $W$ is given by
    \begin{equation*}
    \nabla^2 L^1_F(W) = 2 \Big( W^\top W \otimes I_n + K_{(n)} (W \otimes W^\top) +
I_m \otimes (\Sigma - \Sigma_0)\Big)
\end{equation*}
\end{corollary}

\begin{proof}
   This is a direct consequence of the previous lemma and \Cref{cor:hess-BMparam}. 
\end{proof}

\subsection{Bures-Wasserstein loss}
\label{app:hess-BW}

We now turn to the expression of the Hessian matrix for the Bures-Wasserstein
loss. 
Denote 
\begin{equation}
L_{\tau BW}(\Sigma) = L_\tau(\Sigma) = {\cB(\Sigma_\tau, \Sigma)}^2
\end{equation}
 the $\tau$-regularized Bures-Wasserstein loss, and $L^1_{\tau BW}(W) = L_{\tau
 BW}(WW^\top)$.
Again, we only need to derive the expressions of $\nabla L_{\tau BW}(\Sigma)$ and
$\nabla^2 L_{\tau BW}$.

Recall the expression of the second-order differential for the loss $L_{\tau
BW}$ given in \Cref{lem:sod-L}. Given the definition of the operator $\bG_{\tau}$
in~\eqref{eq:def-G}, and $\Delta$ in~\eqref{lem:sod-L}, we first need to  express the Hadamard product
of two matrices as a linear operation. This is done in the next lemma.

\begin{lemma}[Hadamard product as matrix multiplications]
    \label{lem:hadamard}
    For any $A$, $M$ of same order, with $U\Diag(\sigma_1, \ldots,
    \sigma_n)V^\top = A$ a SVD of $A$, and letting $E_k = \Diag(u_k)$ and
    $F_k = \Diag(v_k)$, one has 
    \begin{equation*}
        A \odot M = (\sum_k \sigma_k u_k \otimes v_k) \odot M = \sum_k \sigma_k
        E_k M F_k.
    \end{equation*}
\end{lemma}

Then, the Hessian of the loss $L_{\tau BW}$ can be computed as follows. 

\begin{proposition}
    \label{prop:sod-LW}
    Let $\Gamma \Diag(q_1, \ldots, q_n) \Gamma^\top = \Sigma_0^{1/2} \Sigma \Sigma_0^{1/2}$ be a
    spectral decomposition, and let $ U\Diag(\sigma_1, \ldots, \sigma_n)U^\top = \left({(\sqrt{q_i} +
    \sqrt{q_j})^{-1}}\right)_{i,j} \eqdef P$ be a spectral decomposition of the
    (symmetric) $P$. Furthermore, let $E_k = \Diag(u_k)$ and let $B_k =
    \Sigma_0^{1/2}\Gamma Q^{-1/2}E_k \Gamma^\top \Sigma_0^{1/2} = B_k^\top$.
    The Hessian matrix 
for the Bures-Wasserstein loss $L_{\tau BW}$ at $\Sigma \in
    \cS_+(n)$ is 
    \begin{equation*}
        \nabla^2 L_{\tau BW}(\Sigma) = \sum_{k = 1}^n \sigma_k B_k^\top \otimes
        B_k . 
    \end{equation*}
\end{proposition}

\begin{proof}
    This follows from the expression of the second-order differential given
    in~\Cref{lem:sod-L} and~\Cref{lem:hadamard}.
\end{proof}

The loss $L^1_{\tau BW}$ admits the parametrization $L^1_{\tau BW} = L_{\tau BW}
\circ \pi$, its Hessian matrix can therefore be computed
using~\Cref{cor:hess-BMparam} and~\Cref{lem:gradient-BW}.
The Hessian matrix of the loss $L^N_{\tau BW}$ can then be computed
using~\Cref{cor:hessian-loss}.

\subsection{Condition number of the Hessian} 

We evaluate the Hessian for both losses in the setting $n = m = d = 20$, and $N
= 3$. We evaluate the Hessians of $L^N_{BW}$ and $L^N_F$ (on the parameter
space) according
to the discussion in \Cref{app:hess-Fro,app:hess-BW}, together with
\cref{cor:hessian-loss}. 
Let $H$ be any symmetric matrix. 
We define the relative condition number for $H$ as 
\begin{equation}
\kappa_{\rel}(H) \defeq \frac{\lambda_{\max}(H)}{\lambda_{\min}(H)},
\end{equation}
and the absolute condition number for $H$ as 
\begin{equation}
    \kappa_{\absol}(H) \defeq \frac{\lambda_{\max}(H)}{\lambda_{\min}^{\absol}(H)},
\end{equation}
where $\lambda_{\min}^{\absol}(H)$ is the minimal eigenvalue in absolute value
of $H$ that is non-zero. Both $\kappa_{\rel}(H)$ and $\kappa_{\absol}(H)$ should
characterize the condition of the matrix $H$: 
if the negative eigenvalues of the Hessian are large in absolute value ($\kappa_{\rel}$ negative, small in absolute value), we expect the gradient descent iterations to escape the saddle points quicker, since the negative eigenvalues corresponds to those escaping, descent directions. Note that this $\kappa_{\rel}$ number is always negative for the Hessian matrices we will consider, but due to the log domain used for plotting, its absolute value is rather reported. 

\begin{figure}[h]
    \centering
    \begin{subfigure}[t]{0.32\textwidth}
        \centering
        \includegraphics[width=\linewidth]{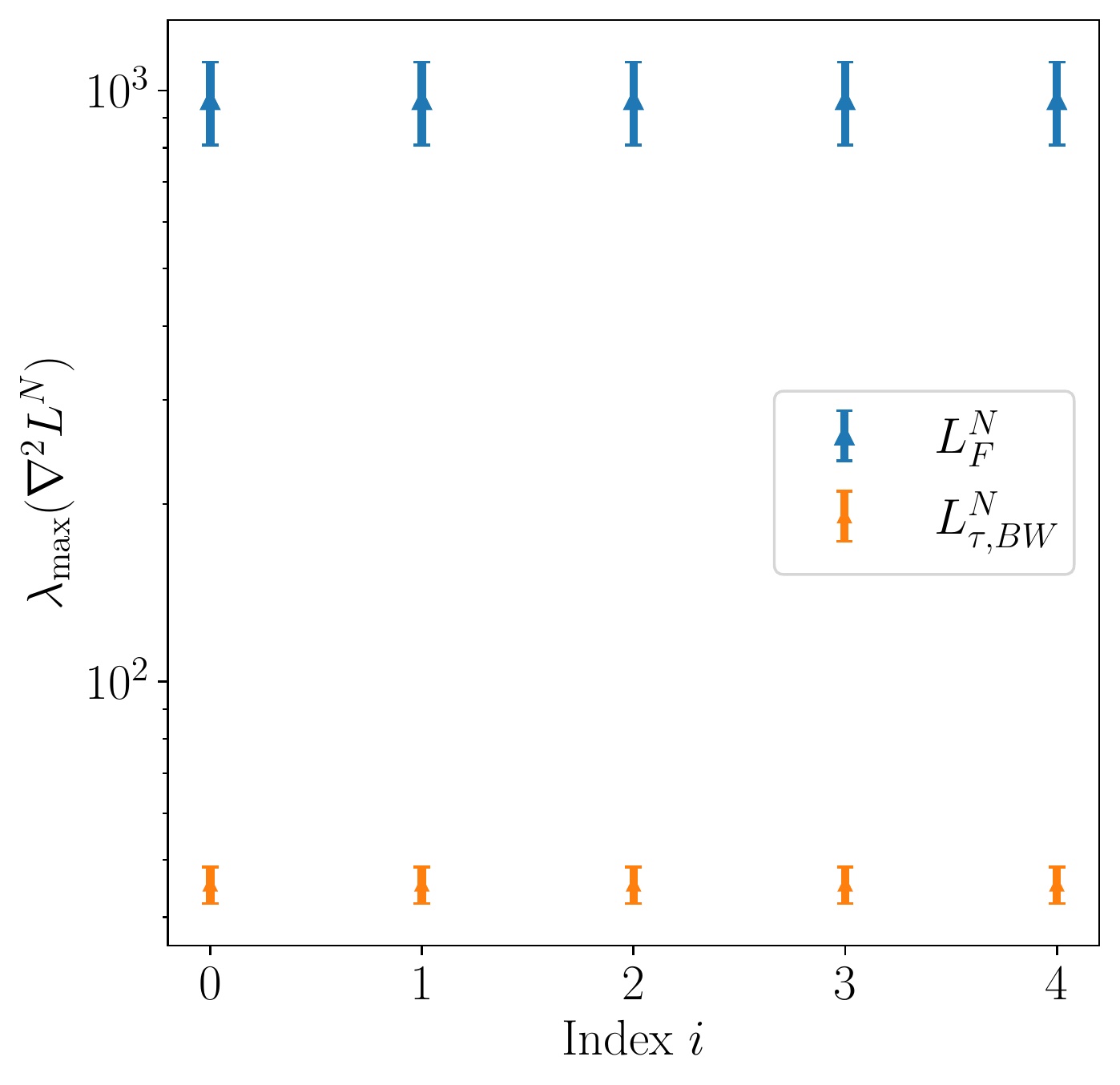}
    \caption{Upper-bounds of the spectrum of $\nabla^2 L^N$. There is an order of
magnitude between the two losses.}    \end{subfigure}
\hfill
    \begin{subfigure}[t]{0.32\textwidth}
        \centering
        \includegraphics[width=\linewidth]{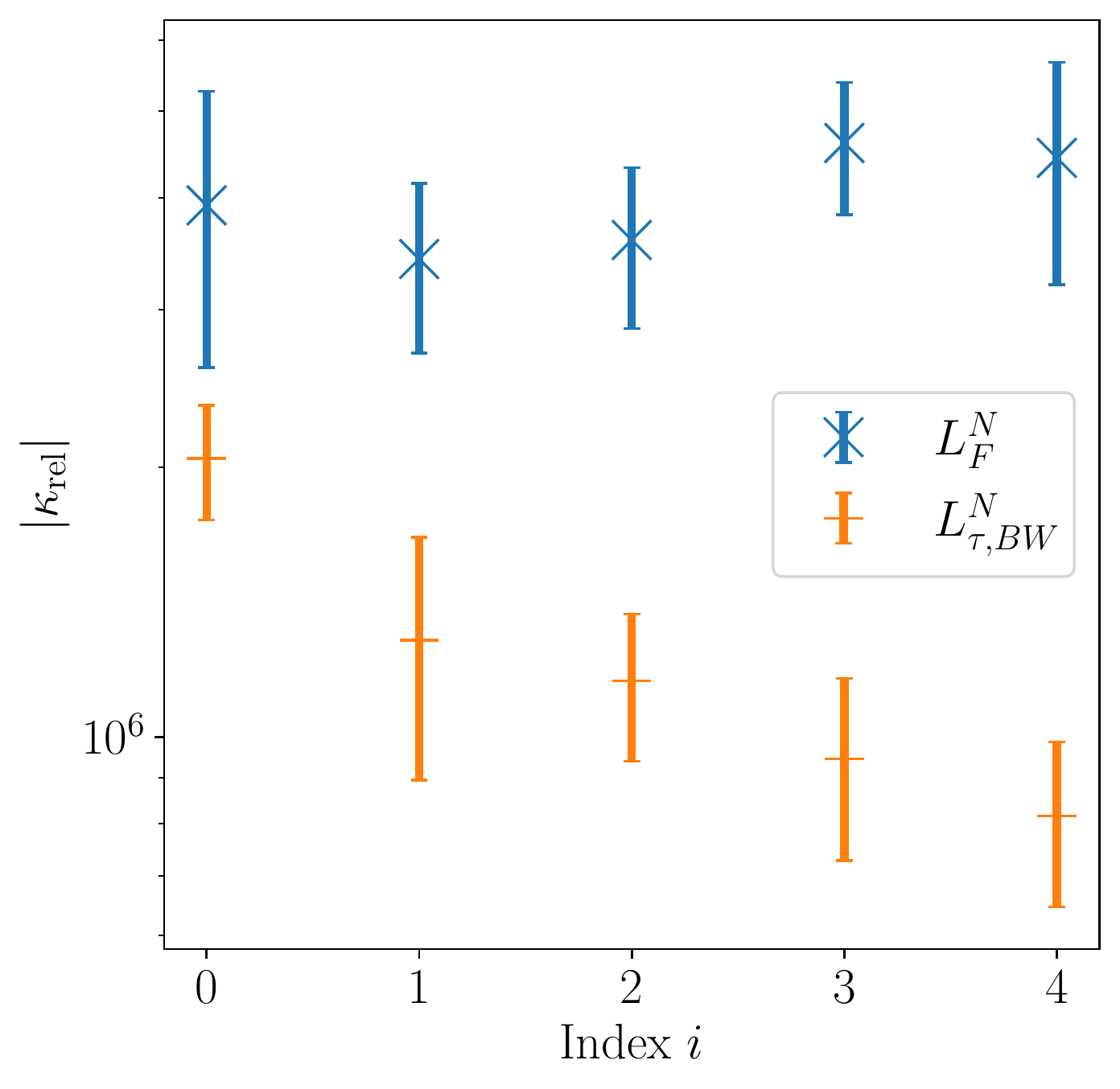}
        \caption{$\kappa_{\rel}(\nabla^2 L^N)$, in absolute value. Lower values
        are linked with better conditioned matrices.}
    \end{subfigure}
    \hfill
    \begin{subfigure}[t]{0.32\textwidth}
        \centering
        \includegraphics[width=\linewidth]{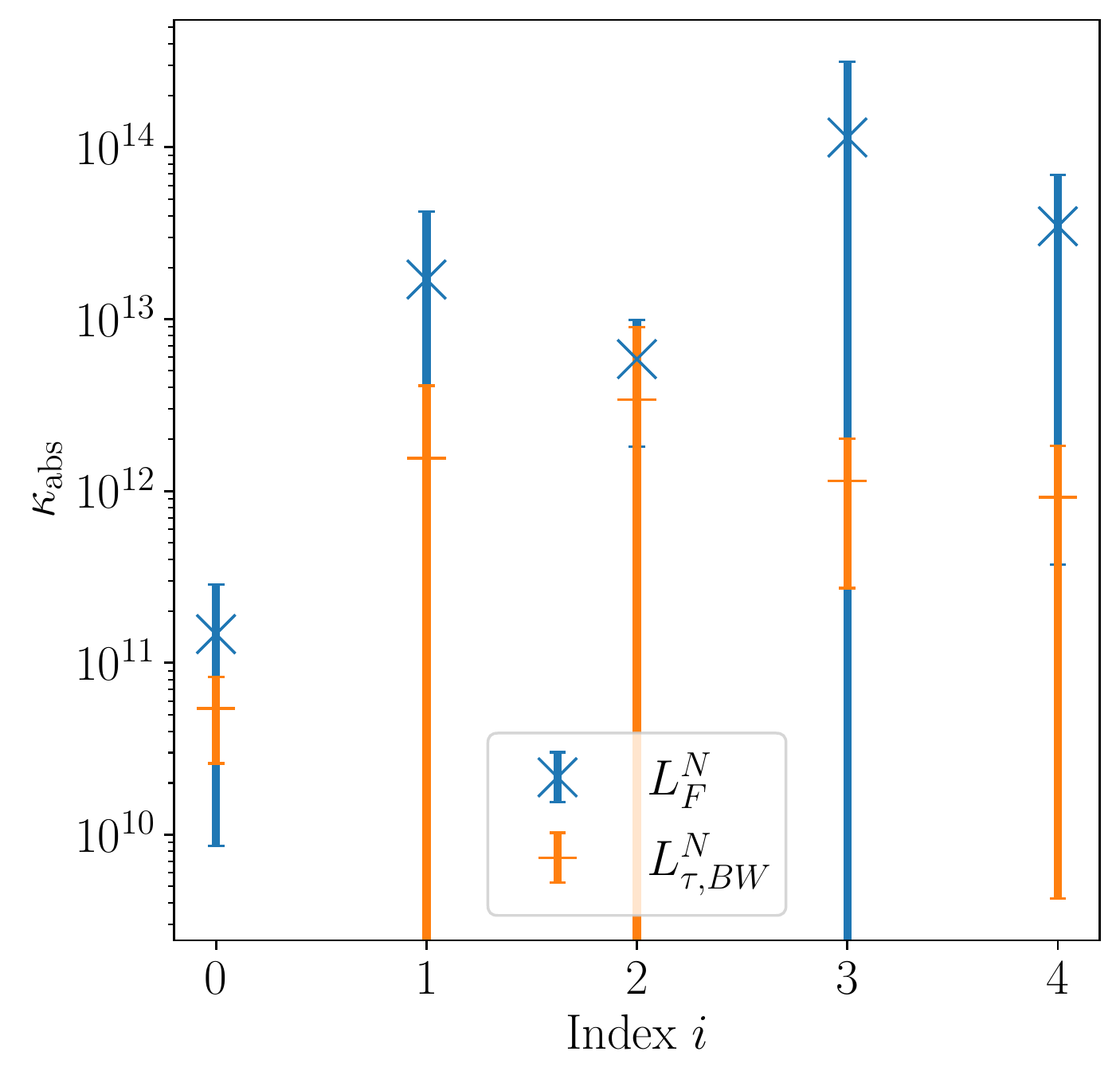}
        \caption{$\kappa_{\absol}(\nabla^2 L^N)$. Lower values are linked with
        better conditioned matrices.}
    \end{subfigure}
\caption{Different spectral values for the Hessian matrices in the case $\tau = 0.1$. The abscissa $i$ refers to the index of the critical point
    $W^*_{i}$. Mean and standard deviations are reported for seven
different targets $\Sigma_0$.}
    \label{fig:hessian-t0.1}       
    \centering
\vspace{.5cm}    
    \begin{subfigure}[t]{0.32\textwidth}
        \centering
        \includegraphics[width=\linewidth]{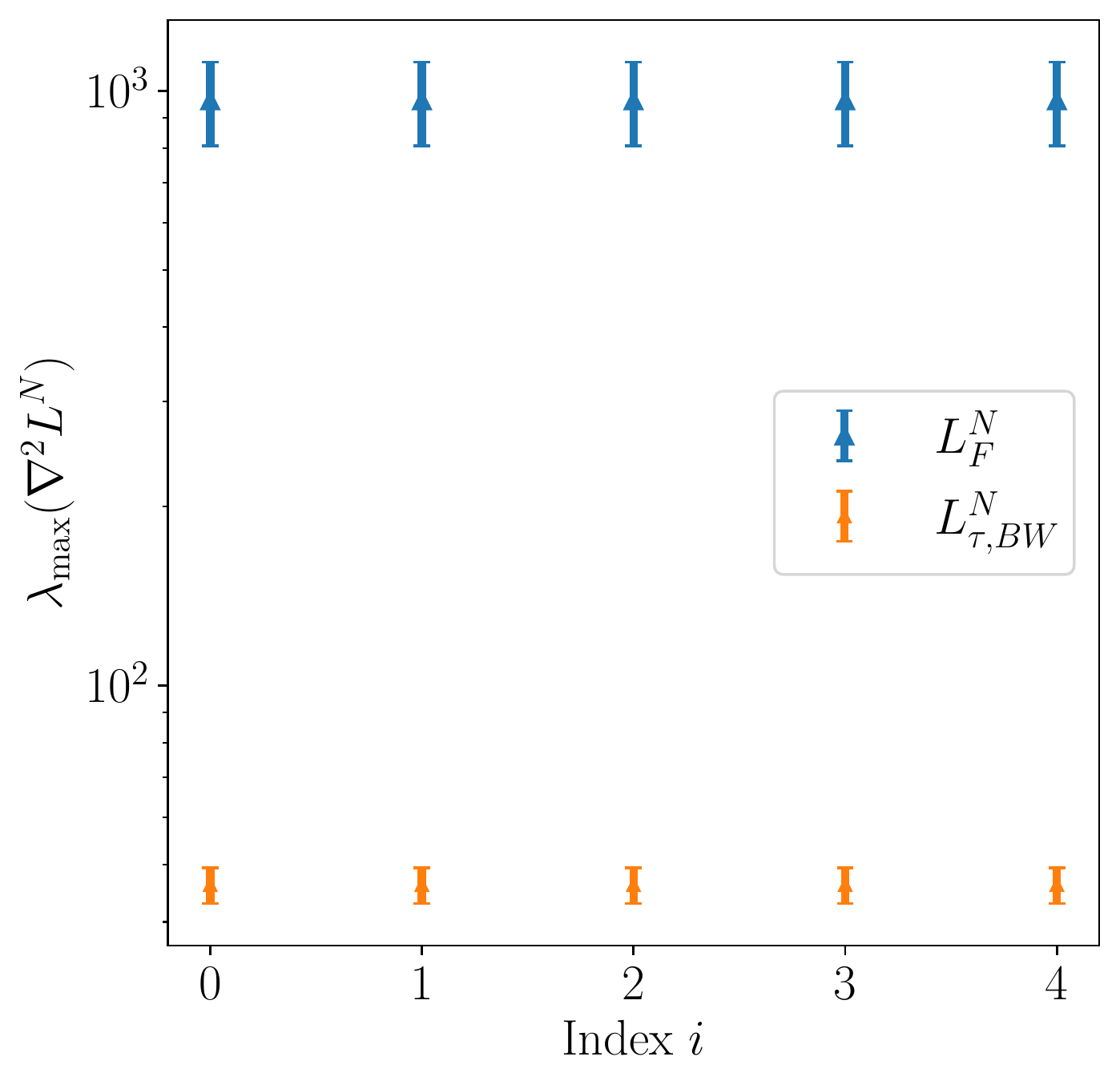}
    \caption{Upper-bounds of the spectrum of $\nabla^2 L^N$. There is an order of
magnitude between the two losses.}
\end{subfigure}
\hfill
    \begin{subfigure}[t]{0.32\textwidth}
        \centering
        \includegraphics[width=\linewidth]{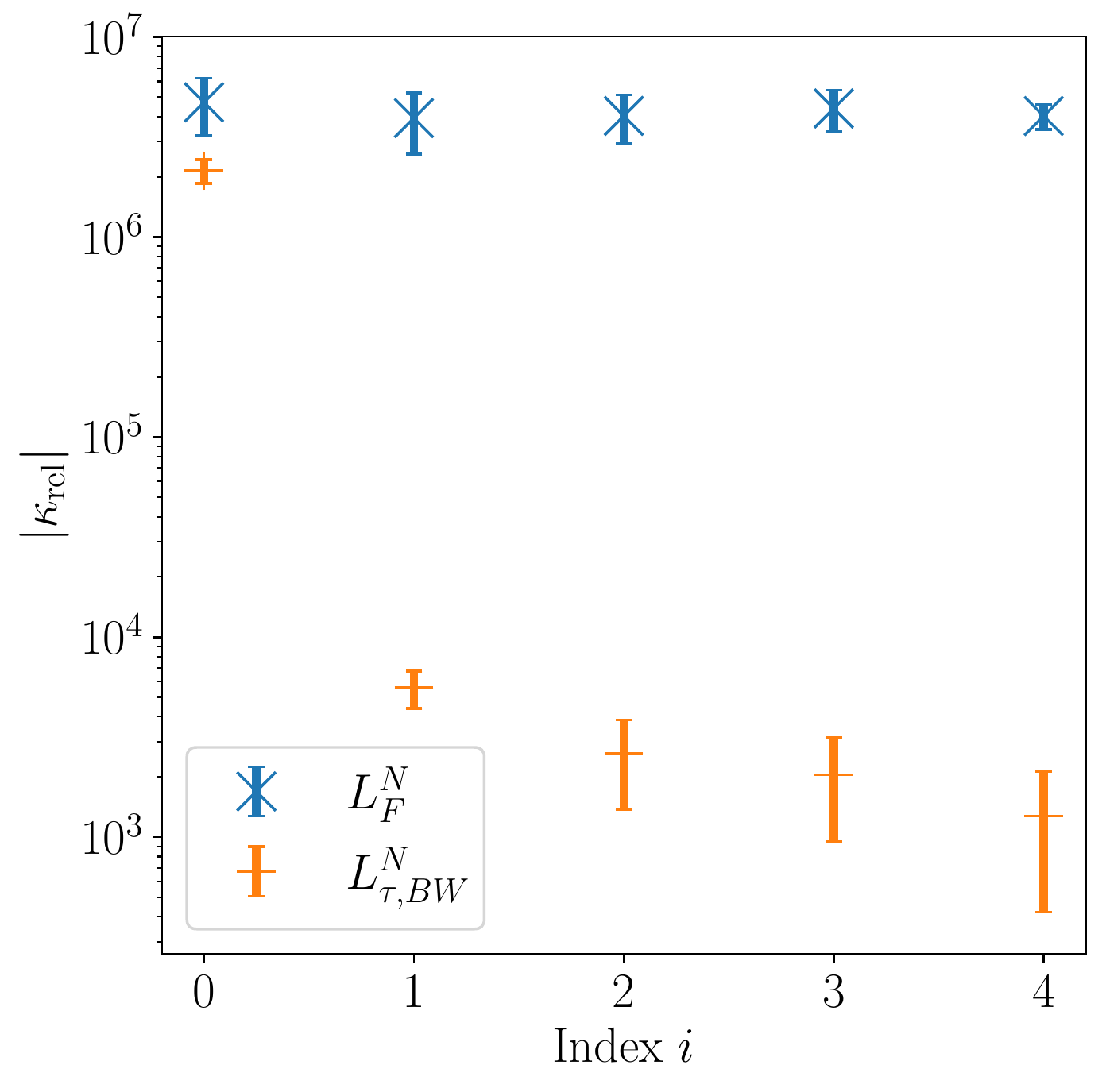}
        \caption{$\kappa_{\rel}(\nabla^2 L^N)$, in absolute value. Lower values
        are linked with better conditioned matrices.}
    \end{subfigure}
    \hfill
    \begin{subfigure}[t]{0.32\textwidth}
        \centering
        \includegraphics[width=\linewidth]{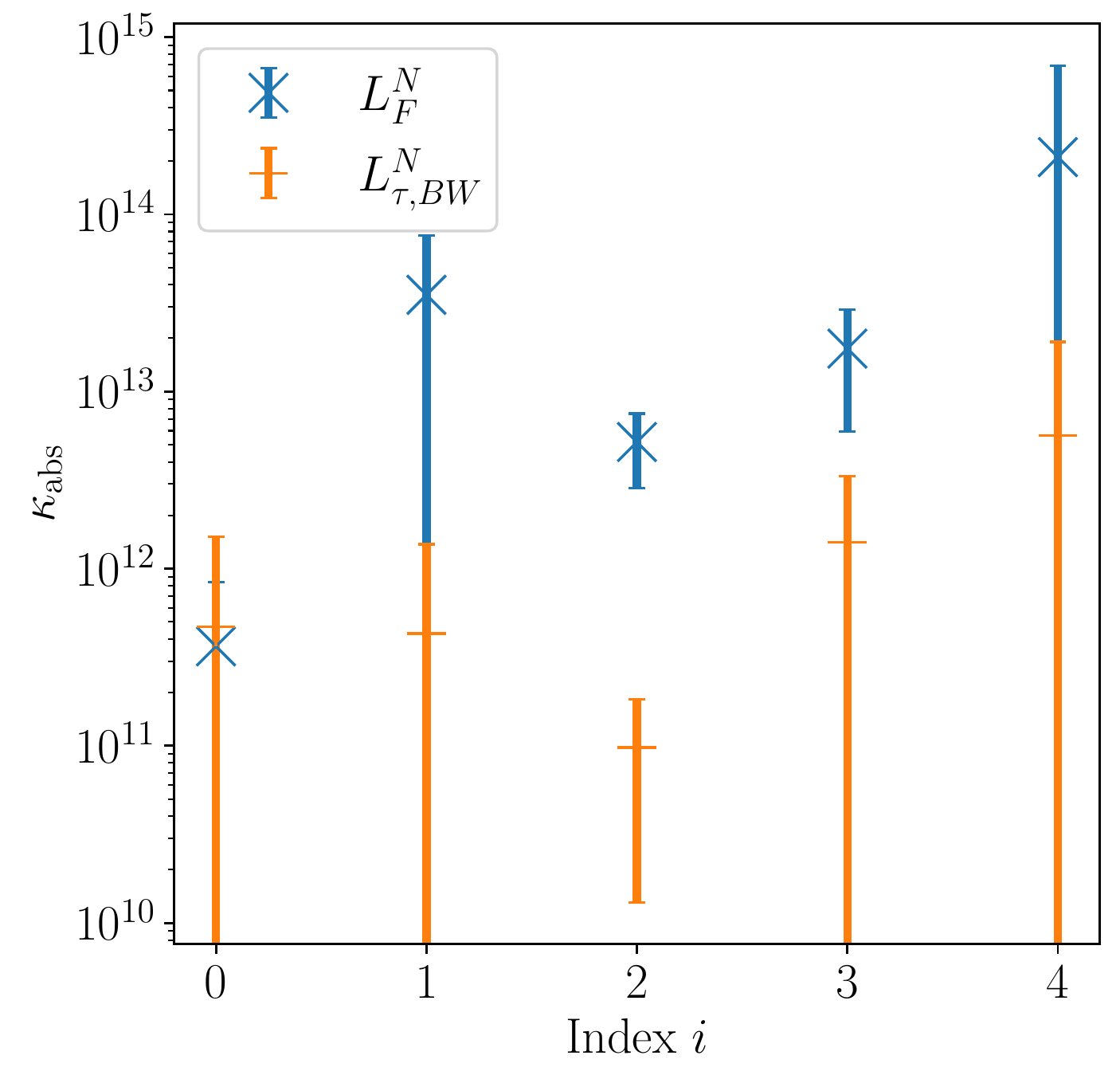}
        \caption{$\kappa_{\absol}(\nabla^2 L^N)$. Lower values are linked with
        better conditioned matrices.}
    \end{subfigure}
\caption{ Different spectral values for the Hessian matrices in the case $\tau =
    0.001$. The abscissa $i$ refers to the index of the critical point
    $W^*_{i}$. Mean and standard deviations are reported for seven
different targets $\Sigma_0$.}
    \label{fig:hessian-t0.001}
\end{figure}

If $\lambda_{\min}^{\absol}(H)$ is large, the Hessian is generally less degenerate
and the iterates can converge quicker to a critical point. We plot the different
quantities in \Cref{fig:hessian-t0.1} (for $\tau = 0.1$) and in
\Cref{fig:hessian-t0.001} (for $\tau = 0.001$), at the five first local
minimizers of the losses with decreasing rank (starting with the full-rank one).
Those are the saddle points for the optimization, and the behavior of the
Hessian at these locations is therefore relevant. Specifically, on the function
space, they are given by the spectral decomposition of the target:  
if $\Sigma_0 = \Omega \Lambda \Omega^\top$ and the eigenvalues are in decreasing
order, then $W^*_{i, F} = \Omega_{[n-i]} \bar\Lambda^{1/2}_{[n-i]} V_{[n-i]}^\top$ is the critical point of $L^1_F$ of index $i$ , and $W^*_{i, \tau BW} = \Omega_{[n-i]}{(\bar\Lambda_{[n-i]} - \tau I_{n-i})}^{1/2} V_{[n-i]}^\top$ is a a critical point of $L^1_{\tau BW}$ of index $i$ according to~\Cref{thm:critical-points-L1tau}. 
The corresponding points in the covariance spaces are simply the images through $\pi$ of $W^*_{i, F},\, W^*_{i, \tau BW}$. For the corresponding points on the parameter spaces,  there are infinitely many of them, but only one that satisfies the balancedness property. This is the one we choose. 

We observe that the Hessian of the Bures-Wasserstein loss $\nabla^2 L^N_{\tau BW}$ is better conditioned than the one of the Frobenius norm $\nabla^2 L^N_{F}$, both for the relative and absolute condition number.

\end{document}